\documentclass{article}
\usepackage{times}
\PassOptionsToPackage{numbers}{natbib}
\usepackage{safecolours}

\usepackage{amsmath}
\usepackage[utf8]{inputenc} %
\usepackage[T1]{fontenc}    %
\usepackage{url}            %
\usepackage{booktabs}       %
\usepackage{amsfonts}       %
\usepackage{microtype}      %
\usepackage{xcolor}         %
\usepackage{graphicx}
\usepackage{tikz,pgfplots}
\pgfplotsset{compat=1.18}
\usepgfplotslibrary{fillbetween}
\usepackage{cancel}
\usepackage[normalem]{ulem}

\usetikzlibrary{external,matrix,backgrounds,positioning}
  \usepackage{iclr2025_conference}
  \iclrfinalcopy

\usepackage{subcaption}
\usepackage{enumitem}
\usepackage{arydshln}
\usepackage{multirow}
\usepackage{array}
\usepackage{colortbl}
\makeatletter
\newcommand{\makesupplementtitle}{%
  \vbox{%
    \hsize\textwidth
    \linewidth\hsize
    \vskip 0.1in
    \toptitlebar
    \centering
    {\LARGE {\bf \@title}\par
    \vspace{0.5em}Supplementary Material\par}
    \bottomtitlebar
    \vskip 0.3in \@minus 0.1in
  }
}
\makeatother

\colorlet{scBlue}{tol-colour1}
\colorlet{scRed}{tol-colour2}
\colorlet{scGreen}{tol-colour3}
\colorlet{scYellow}{tol-colour4}
\colorlet{scCyan}{tol-colour5}
\colorlet{scPurple}{tol-colour6}

\usepackage{pgffor} 
\usepackage{wrapfig}

\newlength{\figurewidth}
\newlength{\figureheight}

\definecolor{cvprblue}{rgb}{0.21,0.49,0.74}
\definecolor{mydarkblue}{rgb}{0,0.08,0.45} %
\usepackage[pagebackref,breaklinks,colorlinks,allcolors=mydarkblue]{hyperref}

\colorlet{highlight}{scCyan!30} %

\newcolumntype{H}{>{\columncolor{highlight}}c}

\pgfplotsset{every axis/.append style={
		legend style={inner xsep=1pt, inner ysep=0.5pt, nodes={inner sep=1pt, text depth=0.1em},draw=none,fill=none}
}} 

\usepackage{shortex-light} %

\usetikzlibrary{fit}

\usepackage{xspace}
\newcommand{\ie}{\textit{i.e.}\xspace}
\newcommand{\cf}{\textit{cf.}\xspace} %
\newcommand{\eg}{\textit{e.g.}\xspace}
\newcommand{\wrt}{{w.r.t.}\xspace}

\newcommand{\Din}{\ensuremath{D_{\text{in}}}}
\newcommand{\Dout}{\ensuremath{D_{\text{out}}}}

\newcommand{\expect}[1]{\mathbb{E}\left \lbrack #1\right \rbrack}
\newcommand{\cov}[2]{\operatorname{\mathbb{C}ov}\sbra{#1,#2}}
\newcommand{\var}[1]{\operatorname{\mathbb{V}ar}\sbra{#1}}

\newcommand{\hessian}{\mathbf{Hess}}

\newcommand{\eigval}{\mathbf{\Lambda}}
\newcommand{\identity}{\mathbf{I}}
\newcommand{\BigO}{\mathcal{O}}
\newcommand{\softmax}[1]{\operatorname{softmax}\rbra{#1}}
\newcommand{\val}[3]{$#1#2$\scalebox{.75}{${\pm}#3$}}
\renewcommand{\mid}{\,|\,}

\newcommand{\RNum}[1]{\uppercase\expandafter{\romannumeral #1\relax}}

\renewcommand{\paragraph}[1]{\noindent{\bfseries#1}~~}

\crefname{appendix}{App.}{Apps.}
\crefname{section}{Sec.}{Secs.}

\tikzset{
  cross/.pic = {
    \draw[rotate = 45] (-#1,0) -- (#1,0);
    \draw[rotate = 45] (0,-#1) -- (0, #1);
  },
  pics/gauss/.style args={#1/#2}{
    code = {
      \begin{scope}
        \clip (-0.5,0) rectangle (1.5,1.5);
        \fill[black!10] (-0.5,0) rectangle (1.5,1.5);
        \fill[thick, domain=-10:10, samples=50] plot[smooth] (\x, {1/(#2*sqrt(2*pi))*exp(-(\x-#1)^2/(2*#2^2))});
        \draw[] (-0.5,0) rectangle (1.5,1.5);
      \end{scope}
    }
  }
}

\title{Streamlining Prediction in Bayesian \\ Deep Learning}

\author{Rui Li \qquad Marcus Klasson \qquad Arno Solin \qquad Martin Trapp \\
Department of Computer Science, Aalto University, Finland\\
\texttt{\{firstname.lastname\}@aalto.fi}
}

\pgfmathdeclarefunction{gauss}{2}{%
  \pgfmathparse{1/(#2*sqrt(2*pi))*exp(-((x-#1)^2)/(2*#2^2))}%
}

\pgfmathdeclarefunction{nnfun}{1}{%
  \pgfmathparse{2 + sin(deg(#1))*sqrt(#1+2) + 0.5*(#1-1)}%
}

\iclrfinalcopy
\begin{document}

\maketitle

\begin{abstract}
The rising interest in Bayesian deep learning (BDL) has led to a plethora of methods for estimating the posterior distribution. However, efficient computation of inferences, such as predictions, has been largely overlooked with Monte Carlo integration remaining the standard. In this work we examine streamlining prediction in BDL through a single forward pass without sampling. For this, we use local linearisation of activation functions and local Gaussian approximations at linear layers. Thus allowing us to analytically compute an approximation of the posterior predictive distribution. We showcase our approach for both MLP and transformers, such as ViT and GPT-2, and assess its performance on regression and classification tasks.\looseness-1 \\[1em]
Open-source library: \url{https://github.com/AaltoML/SUQ}.
\end{abstract}

\begin{figure}[b!]
  \centering
  \newlength{\imgexamplewidth}
  \setlength{\imgexamplewidth}{0.03\textwidth}
  \begin{tikzpicture}
    \begin{axis}[
      xmajorticks=false,
      ymajorticks=false,
      axis line style={draw=none},
      width=0.55\textwidth,
      height=0.35\textwidth,
      ymin=0, ymax=17,
      xmin=-0.2, xmax=2.3,
      ]
        \input{tikz/teaser_density.tex}
        \node (ID) at (axis cs: -0.01, 5) {};
        \node (OOD) at (axis cs: 2, 1.5) {};

      \end{axis}

      \node[font=\small] (IDlabel) at (2,2.5) {\textcolor{scBlue}{In} Domain};

      \node[draw, inner sep=0] at (2,2.1) {\includegraphics[width=\imgexamplewidth]{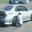}};
      \node[draw, inner sep=0] at ({-\imgexamplewidth + 1.95cm}, 2.1) {\includegraphics[width=\imgexamplewidth]{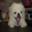}};
      \node[draw, inner sep=0] at ({\imgexamplewidth + 2.05cm},2.1) {\includegraphics[width=\imgexamplewidth]{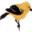}};
      \node[draw, inner sep=0] at ({2cm - 0.5\imgexamplewidth - 0.025cm} ,{2.05cm - \imgexamplewidth }) {\includegraphics[width=0.03\textwidth]{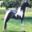}};
      \node[draw, inner sep=0] at ({2cm + 0.5\imgexamplewidth + 0.025cm},{2.05cm - \imgexamplewidth }) {\includegraphics[width=0.03\textwidth]{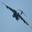}};
  
      \draw[-latex] ({2cm - \imgexamplewidth - 0.05cm},{2.05cm - \imgexamplewidth }) -- (ID);

      \begin{scope}[xshift=2.5cm]
        \node[font=\small] (OODlabel) at (2,2.5) {\textcolor{scRed}{Out} of Domain};

        \node[draw, inner sep=0] at (2,2.1) {\includegraphics[width=\imgexamplewidth]{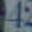}};
        \node[draw, inner sep=0] at ({-\imgexamplewidth + 1.95cm}, 2.1) {\includegraphics[width=\imgexamplewidth]{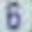}};
        \node[draw, inner sep=0] at ({\imgexamplewidth + 2.05cm},2.1) {\includegraphics[width=\imgexamplewidth]{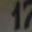}};
        \node[draw, inner sep=0] at ({2cm - 0.5\imgexamplewidth - 0.025cm} ,{2.05cm - \imgexamplewidth }) {\includegraphics[width=0.03\textwidth]{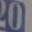}};
        \node[draw, inner sep=0] at ({2cm + 0.5\imgexamplewidth + 0.025cm},{2.05cm - \imgexamplewidth }) {\includegraphics[width=0.03\textwidth]{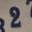}};
      \end{scope}
  
      \draw[-latex] ({4.5cm},{2.1cm - 2\imgexamplewidth }) -- (OOD);

      \node[font=\small] (MAPlabel) at (3,0.5) {\textcolor{gray}{MAP Model}};

      \node at (0.39\textwidth, -0.25) {\small Entropy $\rightarrow$};
      \node[align=center, anchor=center,inner sep=0] at (0.25\textwidth, 3.3) {\strut\textbf{Practical} Outlier Detection};
  
      \begin{scope}[xshift=0.5\textwidth]
        
        \begin{axis}[
            width=0.55\textwidth,
            height=0.34\textwidth,
            axis lines=middle,
            xmajorticks=false,
            ymajorticks=false,
            axis line style={->},
            xlabel={$x$},
            ylabel={$f(x)$},
            domain=0:10,
            samples=200,
            ytick distance=2
        ]

        \node[] (eight) at (axis cs: 2.5, 7) {\includegraphics[width=0.075\textwidth]{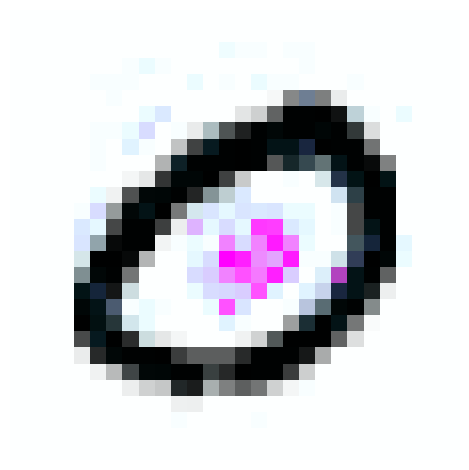}};
        \node[] (zero) at (axis cs: 5, 6) {\includegraphics[width=0.075\textwidth]{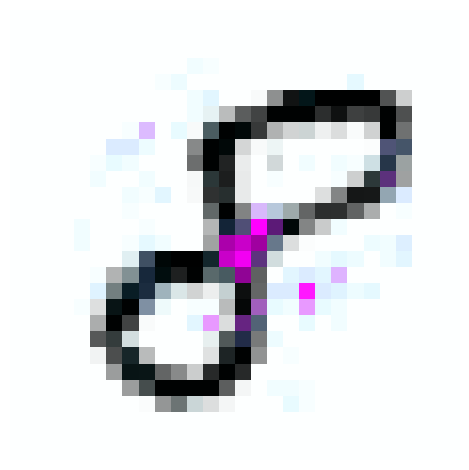}};
        \node[] (five) at (axis cs: 7.75, 3) {\includegraphics[width=0.075\textwidth]{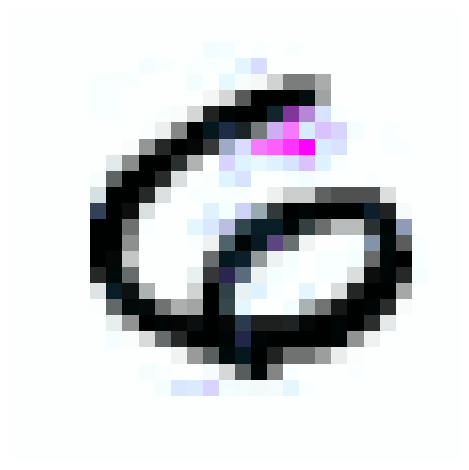}};

        \addplot[gray, very thick, smooth] { nnfun(x) };
        \addplot[gray, dashed, smooth] { nnfun(x) + (0.1 + 0.2*abs(x-5)) };
        \addplot[gray, dashed, smooth] { nnfun(x) - (0.1 + 0.2*abs(x-5)) };

        \def\myxa{1.5}

        \addplot[scPurple, very thick, smooth, domain=(\myxa-1):(\myxa+1), fill=scPurple] (x, { 0.5*gauss(\myxa,0.2) });
        \addplot[scPurple, very thick, smooth, domain=-(0.1 + 0.2*abs(\myxa-5)):(0.1 + 0.2*abs(\myxa-5)), fill=scPurple, fill opacity=0.5 ] ({ gauss(0,0.25)/3 + \myxa }, {x+nnfun(\myxa)});
        \draw[scPurple, very thick,-] (axis cs: \myxa,{ nnfun(\myxa) - (0.1 + 0.2*abs(\myxa-5)) }) --  (axis cs: \myxa,{ nnfun(\myxa) + (0.1 + 0.2*abs(\myxa-5)) });
  
        \def\myxb{3.25}

        \addplot[scCyan, very thick, smooth, domain=(\myxb-1):(\myxb+1), fill=scCyan] (x, { 0.5*gauss(\myxb,0.25) });
        \addplot[scCyan, very thick, smooth, domain=-(0.1 + 0.2*abs(\myxb-5)):(0.1 + 0.2*abs(\myxb-5)), fill=scCyan, fill opacity=0.5 ] ({ gauss(0,0.1)/10 + \myxb }, {x+nnfun(\myxb)});
        \draw[scCyan, very thick,-] (axis cs: \myxb,{ nnfun(\myxb) - (0.1 + 0.2*abs(\myxb-5)) }) --  (axis cs: \myxb,{ nnfun(\myxb) + (0.1 + 0.2*abs(\myxb-5)) });
  
        \def\myxc{8.5}

        \addplot[scYellow, very thick, smooth, domain=(\myxc-1):(\myxc+1), fill=scYellow] (x, { 0.5*gauss(\myxc,0.2) });
        \addplot[scYellow, very thick, smooth, domain=-(0.1 + 0.2*abs(\myxc-5)):(0.1 + 0.2*abs(\myxc-5)), fill=scYellow, fill opacity=0.5 ] ({ gauss(0,0.25)/3 + \myxc }, {x+nnfun(\myxc)});
        \draw[scYellow, very thick,-] (axis cs: \myxc,{ nnfun(\myxc) - (0.1 + 0.2*abs(\myxc-5)) }) --  (axis cs: \myxc,{ nnfun(\myxc) + (0.1 + 0.2*abs(\myxc-5)) });

        \node (pos0) at (axis cs: 3.25, 0.5) {};
        \node (pos8) at (axis cs: 1.5, 0.5) {};
        \node (pos5) at (axis cs: 8.5, 0.5) {};

      \end{axis}

      \node[draw=black, thick, minimum size=15pt, inner sep=0, circle] (circle0) at (pos0) {};
      \node[draw=black, thick, minimum size=30pt, inner sep=0, circle] (circle00) at (zero) {};
      \node[draw=black, thick, minimum size=15pt, inner sep=0, circle] (circle8) at (pos8) {};
      \node[draw=black, thick, minimum size=30pt, inner sep=0, circle] (circle88) at (eight) {};
            \node[draw=black, thick, minimum size=15pt, inner sep=0, circle] (circle5) at (pos5) {};
      \node[draw=black, thick, minimum size=30pt, inner sep=0, circle] (circle55) at (five) {};
  
      \draw (circle0) -- (circle00);
      \draw (circle8) -- (circle88);
      \draw (circle5) -- (circle55);
  
      \end{scope}
  
      \node[align=center, anchor=center,inner sep=0] at (0.75\textwidth, 3.3) {\strut\textbf{Input} Sensitivity Analysis};

      \path (0,0) -- (\textwidth,0);

  \end{tikzpicture}\\[-8pt]
  \caption{Our streamlined approach allows for \emph{practical} outlier detection and  sensitivity analysis. Locally linearising the network function with local Gaussian approximations enables many relevant inference tasks to be solved analytically, helping render BDL a practical tool for downstream tasks.}
  \label{fig:teaser}
\end{figure}

\section{Introduction}
Recent progress and adoption of deep learning models have led to a sharp increase in interest of improving their reliability and robustness.
In applications such as aided medical diagnosis \citep{begoli2019need}, autonomous driving \citep{michelmore2020uncertainty}, or supporting scientific discovery \citep{psaros2023uncertainty}, providing reliable and robust predictions as well as identifying failure modes is vital.
A principled approach to address these challenges is the use of Bayesian deep learning \citep[BDL,][]{wilson2020bayesian,papamarkou2024position} which promises a \emph{plug \& play} framework for uncertainty quantification.
However, while \emph{plugging} the Bayesian approach into deep learning is relatively straightforward \citep{blundell2015weight,gal2016dropout,wu2018deterministic}, the \emph{play} part is typically severely hampered by computational and practical challenges \citep{wenzel2020good,foong2020expressiveness,gelberg2024variational,coker2022wide,kristiadi2023promises}.

The key challenges associated with BDL can roughly be divided into three parts: \textit{(i)}~defining a meaningful prior, \textit{(ii)}~estimating the posterior distribution, and \textit{(iii)}~performing inferences of interest, \eg, making predictions for unseen data, detecting out-of-distribution settings, or analysing model sensitivities. 
While constructing a meaningful prior is an important research direction \citep{nalisnick2018priors,meronen2021periodic,fortuin2021bayesian,tran2022all}, it has been argued that the differentiating aspect of Bayesian deep learning is marginalisation \citep{wilson2020bayesian,wilson2020case} rather than the prior itself.
Hence, estimating the posterior distribution has seen significant progress in recent years \citep{blundell2015weight,maddox2019swag, daxberger2021laplaceredux} with a particular focus on post-hoc approximations \citep{kristiadi2020beingbayesian,daxberger2021subnetwork}.
However, while these approaches have shown promise in making BDL useful for real-world applications, they tackle only part of the computational and practical challenges associated with BDL.

In this work, we focus on streamlining prediction in BDL for downstream tasks by providing a straightforward and effective method to compute inferences of interest, \cf, \cref{fig:teaser}.
For this, we make the neural network locally linear with respect to the inputs.
Thus, inferences, such as computing predictions, admit a closed-form solution and can be estimated efficiently. 
In particular, we propose using local linearisation of non-linear activation functions at every layer of the network and local Gaussian approximations at linear layers.
Empirically, we find that local linearisation combined with Gaussian approximation of Bayesian neural networks provides accurate predictions, with useful predictive uncertainties, while being conceptually simple.
Moreover, complex inference tasks \wrt the inputs, such as analysing model sensitivities to input perturbations, can be computed efficiently.
Thus allowing us to truly account for all sources of uncertainties.

\textbf{Contributions:}~{\it\bfseries (i)}~We propose layer-wise local linearisation and local Gaussian approximations of neural networks to streamline BDL for downstream tasks (\cref{sec:method}).
{\it\bfseries (ii)}~We discuss how to handle different covariance structures and architecture choices (\cref{sec:architectures} \& \cref{sec:structures}).  
{\it\bfseries (iii)}~Finally, we present an empirical assessment of our approach on regression and classification tasks, and showcase its utility for uncertainty quantification, out-of-domain detection, and sensitivity analysis (\cref{sec:experiments}).

\section{Related Work} \label{sec:related}
To estimate the posterior in BDL, variational inference \citep[VI,][]{blei2017variational,zhang2018advances} utilises a variational approximation to the true posterior distribution and minimises a divergence measure between both distributions.
A typical choice for the variational family is a factorised Gaussian distribution, chosen for computational reasons.
Early works on mean-field VI (MFVI) and related approaches require modifications of the model structure \citep{blundell2015weight} to perform a reparametrisation of the variational distribution.
Recent work by \citet{shen2024ivon} developed an optimiser to ease the use of MFVI, and has shown good performance on large-scale models such as ResNets~\citep{he2016deep} and GPT-2~\citep{radford2019language}.
However, VI-based methods typically require Monte Carlo estimation to perform inferences, which can be problematic in practice due to additional computational overhead.

A recent trend in BDL are post-hoc methods, such as the Laplace approximation~\citep[LA,][]{mackay1992information}, which can be applied directly on the trained model without modification \citep{kristiadi2020beingbayesian,daxberger2021laplaceredux}.
\citet{daxberger2021subnetwork} extended the applicability of LAs by showing that treating a subset of parameters Bayesian can still give good predictive uncertainties.
Moreover, \citet{immer2021laplaceglm} proposed the linearised LA by performing a global linearisation, which is principled under the Generalised Gauss--Newton approximation to the Hessian, and has shown promise in providing useful predictive uncertainties. 
Recent works applied post-hoc methods in various applications, such as large language models \citep{yang2024bayeslora,kampen2024towards}, vision-language models \citep{baumann2024posthoc}, dynamic neural networks \citep{meronen2024fixing}, and sequential learning \citep{scannell2024function}.

In addition, various tailored ensemble-based methods for BDL have been proposed, such as Monte Carlo dropout~\citep{gal2016dropout}, deep ensembles~\citep{lakshminarayanan2017deepensemble}, and stochastic weight averaging-Gaussian \citep{maddox2019swag}.
While some works on deep ensembles enable estimating the predictive distribution in a single forward pass \citep{eschenhagen2021deepensemble1,havasi2021deepensemble2}, most methods typically require multiple forward passes to estimate the predictive distribution and do not explicate an approximation to the posterior distribution.

More recently, there has been a trend in exploring deterministic computations in BDL to avoid the need for sampling \citep{goulet2021tractable,giordano2024black,burroni2024sample}.
In particular, \citet{wu2018deterministic} derived an analytical training objective for VI by using moment-matching at each layer of the network.
However, the solutions to the moment-matching have to be derived manually for each type of activation function, making it impractical in practice.  
More recently, \citet{goulet2021tractable} proposed local linearisation of the network to perform message passing on the network under a mean-field assumption.
Moreover, \citet{peterse2024stable} used a local linearisation of the network to propagate aleatoric uncertainties over the input through a deterministic network.
In addition, \citet{dhawan2023efficient} investigated local linearisations of activation functions to estimate the function space distance of two neural networks, for example, relevant in continual learning settings.
In contrast, our work disentangles the approximation of the posterior distribution and the computation of inferences \wrt the posterior distribution. 
Hence, it provides a streamlined framework to propagate all forms of uncertainties through Bayesian neural networks.

\section{Method}\label{sec:method}
In Bayesian deep learning (BDL), predicting the output $y$ (\eg, class label, regression value) for an input $\bx \in \mcX$ is performed by \emph{marginalising} out the model parameters $\btheta$ of the neural network $f_{\btheta}(\cdot)$ instead of trusting a single point estimate, \ie, 
\begin{equation}
  p(y \mid \bx, \mcD) = \int_{\btheta} p(y \mid f_{\btheta}(\bx)) \, p(\btheta \mid \mcD) \, \dee \btheta, \label{eq:posterior_predictive}
\end{equation}
where $\mcD=\{(\bx_n, y_n)\}^N_{n=1}$ denotes the training data and the posterior distribution $p(\btheta \mid \mcD) = \frac{p(\btheta,\mcD)}{p(\mcD)}$ is given by Bayes' rule.
However, for most neural networks, integrating over the high-dimensional parameter space is intractable, necessitating the use of approximations to compute the posterior distribution $p(\btheta \mid \mcD)$ and the posterior predictive distribution $p(y \mid \bx, \mcD)$.

Recently, much progress has been made in efficiently approximating the posterior distribution for BDL, including scaling mean-field variational inference \citep{shen2024ivon} to large-scale models and performing post-hoc estimation using the Laplace approximations \citep{daxberger2021laplaceredux}. A common thread is using a tractable distribution $q$ to approximate the posterior distribution $q(\btheta) \approx p(\btheta \mid \mcD)$, commonly chosen as a Gaussian distribution.
Consequently, the posterior predictive distribution is typically approximated using Monte Carlo integration, \ie, by sampling from $q$, to estimate the integral in \cref{eq:posterior_predictive}, with the exception of the linearised Laplace approximation \citep{immer2021laplaceglm}.
However, while using a Gaussian approximation facilitates efficient computation of the approximate posterior distribution, sampling from the high-dimensional Gaussian approximation can be challenging \citep{vono2022high} and result in high computational overhead.

\begin{figure}[t!]
  \begin{subfigure}[b]{0.47\textwidth}
    \centering
    \resizebox{\linewidth}{!}{
    \begin{tikzpicture}

      \node (labelW) at (-1.5, 1) {$\bW^{(l)}$};
      \node (imgW) at (-1.5, 0) {\includegraphics[page=2]{tikz/figure2.pdf}};

      \node at (-0.5, 0) {$\times$};

      \node (labelh) at (0, 1) {$\ba^{(l-1)}$};
      \node (imgh) at (0,0) {\includegraphics[page=1]{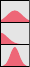}};

      \node (approx1) at (0.5, 0) {$\approx$};

      \node (labela) at (1, 1) {$\bh^{(l)}$};
      \node (imga) at (1, 0) {\includegraphics[page=3]{tikz/figure2.pdf}};

      \draw[-latex] (1.5, 0) -- node[above] {$g(\bh^{(l)})$} (2.75, 0);
      \node (approx2) at (3, 0) {$\approx$};
      \node (labelf) at (3.5, 1) {$\ba^{(l)}$};
      \node (imgf) at (3.5, 0) {\includegraphics[page=4]{tikz/figure2.pdf}};

      \node[draw, dashed, fit=(labelW) (imgW) (labelf) (imgf)] (fit) {};
      \draw[-latex, thick] ([xshift=-0.75cm, yshift=0]fit.west) -- node[above] {} ([xshift=0.25cm, yshift=0]fit.west);
      \draw[-latex,thick] ([xshift=-0.25cm, yshift=0]fit.east) -- ([xshift=0.75cm, yshift=0]fit.east);

      \node[font=\small, align=center] (text1) at ([xshift=0, yshift=2.cm]approx1) {Local Gaussian\\ Approximation};
      \draw[-latex] (text1) -- (approx1);
      
      \node[font=\small, align=center] (text2) at ([xshift=0, yshift=2.cm]approx2) {Local Linearisation\\ ~};
      \draw[-latex] (text2) -- (approx2);

    \end{tikzpicture}
    }
    \caption{Multi-layer Perceptron (MLP)}
  \end{subfigure}
  \hfill
  \begin{subfigure}[b]{0.52\textwidth}
    \centering
    \resizebox{\linewidth}{!}{
    \begin{tikzpicture}

      \node (labelh) at (0, 1) {$\bH^{(l-1)}$};
      \node (imgh) at (0, 0) {\includegraphics[page=5]{tikz/figure2.pdf}};
      
      \node[anchor=west] (query) at (2, 1) {$\expect{\bH^{(l-1)} \bW_Q^{(l)}}  $};
      \node[anchor=west] (key) at (2, 0) {$ \expect{\bH^{(l-1)}\bW_K^{(l)}} $};
      \node[anchor=west] (value) at (2, -1) {$ \bH^{(l-1)} \bW_V^{(l)}$};

      \draw[-latex] (imgh) -- (query.west);
      \draw[-latex] (imgh) -- (key.west);
      \draw[-latex] (imgh) -- (value.west);
      
      \node[anchor=west] (imgvalue) at (2, -2) {\includegraphics[page=6]{tikz/figure2.pdf}};

      \node (softmax) at (8.0, 1) {$\softmax{\frac{\bQ^{(l)} {\bK^{(l)}}^\top}{\sqrt{D}}}\bV^{(l)}$};
      \node (imgA) at (7.5, 0) {\includegraphics[page=7]{tikz/figure2.pdf}};

      \draw[-latex] (query.east) -- node[above] {$\bQ^{(l)}$} ([xshift=0, yshift=5pt]imgA.west);
      \draw[-latex] (key.east) -- node[above] {$\bK^{(l)}$} (imgA.west);
      \draw[-latex] (value.east) -- node[below] {$\bV^{(l)}$} ([xshift=0, yshift=-5pt]imgA.west);

      \node[draw, dashed, fit=(imgA) (imgh) (labelh) (softmax) (imgvalue)] (fit) {};

      \draw[thick, -latex] ([xshift=-0.5cm, yshift=0]fit.west) -- ([xshift=0.25cm, yshift=0]fit.west);
      \draw[thick, -latex] ([xshift=-0.25cm, yshift=0]fit.east) -- ([xshift=0.5cm, yshift=0]fit.east);
    \end{tikzpicture}
    }
    \caption{Attention Layer}
  \end{subfigure}
  \caption{Illustration of our approach for different network architectures. In MLPs, we can directly apply local Gaussian approximations and local linearisation of each layer. The distribution over activations is then propagated to the next layer.
  In attention layers, we treat the query $\bQ$ and key $\bK$ deterministically and only treat the value $\bV$ as a random quantity, resulting in a straightforward propagation path. The resulting distribution is then propagated to the subsequent MLP layer.
   }
  \label{fig:illustration}
\end{figure}

We will now shift our focus on estimating integrals of the form of \cref{eq:posterior_predictive} and assume that an approximation to the posterior distribution $q(\btheta)$ is given.
Further, we will assume that $q$ is in the family of stable distributions. Note that a linear combination of two independent random variables following a stable distribution has the same distribution as the distribution of the individual random variables.
The Gaussian distribution is a typical example of a stable distribution.
Marginalisation tasks such as in \cref{eq:posterior_predictive} appear in many scenarios, \eg, active learning \citep{mackay1992information,gal2017deep,smith2023prediction}, model selection \citep{immer2021scalable,mackay1996bayesian}, or outlier detection \citep{wilson2020bayesian}, and pose a reappearing challenge in downstream applications of BDL.

\subsection{Streamlining Computations with Local Approximations}
Let the weights and biases of the $m$\textsuperscript{th} linear layer of the network $f$ be denoted as $\bW^{(m)} \in \reals^{\Dout \times \Din}$ and $\bb^{(m)} \in \reals^{\Dout}$, respectively.
Then the pre-activation $\bh^{(m)}$ is given as $\bh^{(m)} = \bW^{(m)}\ba^{(m-1)} + \bb^{(m)}$, where $\ba^{(m-1)} \in \reals^{\Din}$ is the activation of the previous layer.
In case $m=1$, then $\ba^{(0)}$ corresponds to the input $\bx$.
We further denote the $k$\textsuperscript{th} element of $\bh^{(m)}$ as $h_k^{(m)} = \sum_{i=1}^{\Din} W_{ki}^{(m)} a_i^{(m-1)} + b_k^{(m)}$ and drop the superscript if it is clear from the context.

Given an approximate posterior distribution $q(\btheta)$ with $\btheta = \{\bW^{(m)}, \bb^{(m)}\}^M_{m=1}$, we aim to compute the probability distribution of the activation $\ba^{(m)}$ of each layer $m$.
For this, we need to estimate the distribution of the pre-activation $\bh^{(m)}$ and then compute an approximation to the activation $\ba^{(m)}$ after the application of a non-linear activation function $g(\cdot)$.

\paragraph{Approximating the pre-activation distribution}
In case the activation $\ba^{(m-1)}$ is deterministically given, \ie, for the input layer, we can compute the distribution over pre-activations analytically as a consequence of the stability of stable distributions under linear transformations \citep{peterse2024stable}.
However, for hidden layers, the distribution over pre-activations is generally not of the same family as the posterior distribution \citep{wolinski2022gaussian}.
Nevertheless, we will apply a local Gaussian approximation to the pre-activation at every hidden layer.
Specifically, we make the assumption:
\begin{assumption}
  Assume that the activations of the previous layer $a^{(m-1)}_i$ and parameters of the $m$\textsuperscript{th} layer are independent.
\end{assumption}

Then followed by a Gaussian approximation of $a^{(m-1)}_i\,W^{(m)}_{ki}$ for each $i$ and each $k$, the mean of the pre-activation $\bh^{(m)}$ is given as:
\begin{equation}
  \expect{\bh^{(m)}} = \expect{\bW^{(m)}}\expect{\ba^{(m-1)}} + \expect{\bb^{(m)}}, \label{eq:aw-mean}
\end{equation}
and the covariance between the $k$\textsuperscript{th} and the $j$\textsuperscript{th} hidden unit is computed as:
\begin{align}
  \cov{h^{(m)}_k}{h^{(m)}_l} &= \sum_{\mathclap{1\leq i,j \leq \Din}} \cov{a^{(m-1)}_iW^{(m)}_{ki}}{a^{(m-1)}_jW^{(m)}_{lj}} + \cov{b^{(m)}_k}{b^{(m)}_l} \nonumber \\ 
  &+ \sum_{\mathclap{1\leq i \leq \Din}} \expect{a^{(m-1)}_i}\rbra{\cov{W^{(m)}_{ki}}{b^{(m)}_l} + \cov{W^{(m)}_{li}}{b^{(m)}_k}}  , \label{eq:cov-whole}
\end{align}
where
\begin{align}
  \cov{a^{(m-1)}_iW^{(m)}_{ki}}{a^{(m-1)}_jW^{(m)}_{lj}} &= \expect{a^{(m-1)}_i}\expect{a^{(m-1)}_j}\cov{W^{(m)}_{ki}}{W^{(m)}_{lj}} \nonumber\\ 
  &+ \expect{W^{(m)}_{ki}} \expect{W^{(m)}_{lj}} \cov{a^{(m-1)}_i}{a^{(m-1)}_j}\nonumber\\
  &+ \cov{a^{(m-1)}_i}{a^{(m-1)}_j}\cov{W^{(m)}_{ki}}{W^{(m)}_{lj}}. \label{eq:cov-aw}
\end{align}
A detailed derivation alongside an empirical evaluation of the approximation quality can be found in \cref{app:derivation-full,app:aw-approx}. 
Depending on the structure of the covariance matrix, we can further simplify the computation of the covariance matrix, which we will discuss in \cref{sec:structures}.

\paragraph{Approximating the activation distribution}
Let $g(\cdot)$ denote a non-linear activation function computing $\ba = g(\bh)$ for a pre-activation $\bh$.
Inspired by the application of local linearisation in Bayesian filtering \citep[\eg,][]{sarkka2023bayesian}, we use a first-order Taylor expansion of $g(\cdot)$ at the mean of the pre-activation $\expect{\bh}$. %
Specifically, we approximate $g(\bh)$ using
\begin{equation}
    g(\bh) \approx g(\expect{\bh}) + \bJ_g|_{\bh=\expect{\bh}}(\bh - \expect{\bh}), \label{eq:first_order_taylor}
\end{equation}
where $\bJ_g|_{\bh=\expect{\bh}}$ is the Jacobian of $g(\cdot)$ at $\bh = \expect{\bh}$. 
Then, as stable distributions are closed under linear transformations, the distribution of $\ba$ can be computed analytically and is given as follows in case of a Gaussian distribution, \ie, 
\begin{equation}\label{eq: cov-activation}
    \ba \sim \mathcal{N}(g(\expect{\bh}), \bJ_g|_{\bh=\expect{\bh}}^{\top}\bSigma_{\bh}\bJ_g|_{\bh=\expect{\bh}}).
\end{equation}
Note that the quality of the local linearisation will depend on the scale of the distribution over the input $\bh$.
For ReLU activation functions, \citet{peterse2024stable} have shown that local linearisation provides the optimal Gaussian approximation of a univariate Gaussian distribution in total variation.
For classification tasks, we employ a probit approximation \cite{mackay1992prohit-1,kristiadi2020beingbayesian}. 

\paragraph{Intuition}
One way to understand the resulting approximation is as a piecewise linear function (or multilinear function). Globally, the function will still be non-linear, but locally it will behave linearly. In contrast to the original model, which composes piecewise linear functions in the case of a ReLU network, our approximation composes linear functions locally. We obtain a piecewise linear function due to the local composition, which allows us to capture the non-linear nature of the model.

\subsection{Architecture Choices} \label{sec:architectures}
By combining local Gaussian approximations for linear layers and local linearisation for non-linear activation functions, we can analytically compute the distribution over activations at each layer in a single forward pass.
In the case of a multi-layer perceptron (MLP) and common architecture choices, the described approach can be directly applied to each layer of the network.
However, further considerations are needed to streamline the computation path for more complex architectures such as attention.
\cref{fig:illustration} illustrates the computation path for MLPs and attention layers.

\paragraph{Attention layers}
Each block in a transformer \citep{vaswani2017attention} constitutes: 
multi-head attention, an MLP, layer normalisation, and a residual connection. 
For the MLP part, the propagation is the same as previously described.
Further, layer normalisation is a linear transformations, and the resulting distribution can be obtained analytically.
For residual connections by assuming independence, we could also obtain the resulting distribution analytically.
Treating the multi-head attention block is more involved as the softmax activation function `squashes' the distribution of the pre-activations.
We describe our method below, and further details are in \cref{app:derivation-transformer}.

Given an input $\bH \in \mathbb{R}^{T \times D}$, where $T$ is the number of tokens in the input sequence and $D$ is the dimension of each token, denote the query, key and value matrices as $\bW_Q \in \mathbb{R}^{D \times D}$, $\bW_K \in \mathbb{R}^{D \times D}$, $\bW_V \in \mathbb{R}^{D \times D}$, respectively.
Further, we denote the key, query and value in an attention block as $\bQ = \bH \bW_Q$, $\bK = \bH \bW_K$, and $\bV = \bH \bW_V$.
Then the output of attention layer is given as follows $\operatorname{Attention}(\bH) = \operatorname{Softmax}\left(\nicefrac{\bQ \bK^\top}{\sqrt{D}}\right) \bV.$
For computational reasons, we will assume the input distribution to the multi-head attention block has a diagonal covariance structure.
`Pushing' random vectors over a softmax activation may require further approximations and will not result in an output with a distribution close to a Gaussian distribution. Hence, we treat the query and key matrices as deterministically given. A possible remedy is to leverage an approximation to the softmax function such as \cite{lu2021soft}.
Consequently, the attention scores are given as:
\begin{equation}
 \operatorname{Attention}(\bH) = \operatorname{Softmax}\left(\frac{\expect{\bH}\expect{\bW_Q} (\expect{\bH} \expect{\bW_K})^\top}{\sqrt{D}}\right) \bV,
\end{equation}
where $\bV$ follows a stable distributions.
Due to linearity, the resulting distribution can again be obtained analytically.

\subsection{Covariance Structure} \label{sec:structures}
Computing the full covariance of the posterior is usually infeasible due to high computational and memory cost. %
We describe our methods for the two most common covariance approximations and will briefly discuss the computational cost in the case of a full and diagonal covariance structure.

\paragraph{Full covariance} When the posterior has full covariance, for the $m$\textsuperscript{th} linear layer the computational complexity for computing $\cov{h_k}{h_l}$ is $\BigO([\Din^{(m)}]^2)$. 
Consequently, computing the covariance of the activations for the $m$\textsuperscript{th} layer adds to $\BigO([\Dout^{(m)}]^2[\Din^{(m)}]^2)$.
Computing the local linearisation for element-wise activation functions results in a complexity of $\BigO([\Dout^{(l)}]^2)$.
Hence, we obtain a total cost of $\BigO\rbra{\sum_{m=1}^M [\Dout^{(m)}]^2[\Din^{(m)}]^2 + [\Dout^{(m)}]^2}$ for a network with $M$ layers. 
As the computational cost is directly linked to the number of parameters and their correlation structure, a natural way to reduce the computational cost is to either exploit the structure in the covariance matrix or consider only a subset of parameters, in the spirit of subnetwork Laplace \citep{daxberger2021subnetwork}. We will focus on exploiting the structure of the covariance as using a subset of parameters trivially extends from our discussion.\looseness-1

\paragraph{Diagonal approximation} In case the correlations between model parameters are dropped, as in mean-field variational inference, the computation of the pre-activation covariance reduces to:
\begin{equation}
	\cov{h^{(m)}_k}{h^{(m)}_l} =\sum_{\mathclap{1\leq i,j \leq \Din}}\expect{W^{(m)}_{ki}} \expect{W^{(m)}_{lj}} \cov{a^{(m-1)}_i}{a^{(m-1)}_j},  \label{eq:diag-cov}
\end{equation}
and variance of the $k$\textsuperscript{th} pre-activation is given as: $\var{h^{(m)}_k} =$
\begin{equation}
  \sum_{\mathclap{1\leq i \leq \Din}} \expect{a^{(m)}_i}^2\var{W^{(m)}_{ki}} + \var{b^{(m)}_k} + \var{a^{(m-1)}_i}\rbra{\expect{W^{(m)}_{ki}}^2 + \var{W^{(m)}_{ki}}} . \label{eq:diag-var}
\end{equation}
Hence, assuming a diagonal covariance structure can help in reducing the computational burden.
If we further drop the correlation between pre-activation, the computational cost can be reduced and adds to a total of $\BigO(\sum_{m=1}^M {\Dout^{(m)}}{\Din^{(m)}} + {\Dout^{(m)}})$. Further details are given in \cref{app:derivation-diag}.
An empirical run time analysis indicating little to no overhead is given in \cref{app:runtime}.

\paragraph{Kronecker-factorisation (KFAC)} Another common choice for approximating the posterior covariance is the use of a Kronecker-factorisation (KFAC) \citep{martens2015optimizing}, popularised in the context of Laplace approximations \citep{ritter2018scalable}.
In this case, the posterior covariance $\bSigma$ is given by a Kronecker product of two factors. 
Denote the Kronecker product as $\kron$ and the prior precision as $\lambda^2 \identity$, for column flattening convention, the posterior covariance is $\bSigma = (\bA \kron \bB + \lambda^2 \identity)^{-1}$.
For row flattening convention, the posterior covariance is $\bSigma = (\bB \kron \bA + \lambda^2 \identity)^{-1}$.
We refer the reader to \cite{dangel2025kroneckerfactored} for more details on the KFAC and flattening convention.
Row flattening convention allows easy access to the covariance between the $k\textsuperscript{th}$ and $l\textsuperscript{th}$ row in weight $\bW$, \ie, $\cov{\bW[k,:]}{\bW[l, :]}$.
Therefore, we use row flattening convention here and discuss how to retrieve $\cov{\bW[k,:]}{\bW[l, :]}$ when following column convention in \cref{app:derivation-kron}.

Note that in case of a non-zero prior precision, the covariance cannot be expressed in the form of a Kronecker matrix multiplication.
As our method requires direct access to the posterior covariance, to remain the benefit of memory storage of KFAC, we adopt a commonly used approximation on the covariance:
\begin{align}
	\bSigma = (\bB \kron \bA + \lambda^2 \identity)^{-1} &\approx \underbrace{(\bB + \lambda \identity)^{-1}}_{\widetilde{\bB}} \kron \underbrace{(\bA + \lambda \identity)^{-1}}_{\widetilde{\bA}} \label{eq:kfac-approx}
\end{align}
This way, the covariance between the $k\textsuperscript{th}$ and $l\textsuperscript{th}$ row in weight $\bW$ can be easily retrived by $\cov{\bW[k,:]}{\bW[l, :]} = \widetilde{\bB}[k, l]\widetilde{\bA}$ without explicating the full covariance matrix memory.

\section{Experiments}\label{sec:experiments}
We demonstrate \emph{practical applicability} of our approach on classification/regression tasks (\cref{sec:regression-classification}), large-scale classification results with ViT/GPT models (\cref{sec:vit-exp}), and sensitivity estimation (\cref{sec:sensitivity-exp}). 
Additional experiments and additional experimental results can be found in \cref{app:experiment}.

\paragraph{Data sets} 
We use a selection of data sets from the UCI repository~\citep{ucirepository} for the regression experiments. For classification, we experiment on 
MNIST~\citep{lecun1998gradient}, FMNIST~\citep{xiao2017fashion}, as well as the 11-class data sets OrganCMNIST and OrganSMNIST from MedMNIST~\citep{Yang2023medmnist}. 
To assess our method on higher-dimensional settings, we experiment with CIFAR-10 and CIFAR-100~\citep{krizhevsky2009learning}, DTD~\citep{cimpoi2014dtd}, RESISC~\citep{cheng2017resisc} and %
a subsampled version of ImageNet-R~\citep{hendrycks2021imagenetr} with 100 classes to reduce the memory overhead for the LA. 
For the GPT model, we used the BOOLQ, WIC, and MRPC tasks from GLUE \citep{wang2018glue} and SuperGLUE \citep{wang2019superglue} benchmarks.

\paragraph{Posterior approximations} 
We adopt the Laplace approximation (LA, \cite{mackay1996bayesian}) and mean-field variational inference \citep[MFVI,][]{blei2017variational} for approximating the posterior distribution of the network parameters. 
For the LA, we estimate the full covariance for the regression experiments, while we use diagonal or KFAC approximations for the covariance where applicable in the classification experiments. 
We compare our method using local Gaussian approximation and local linearisation against Monte Carlo (MC) sampling and a global linearised model \citep[GLM,][]{immer2021laplaceglm}. 
For MFVI, we adopt the IVON optimiser~\citep{shen2024ivon} to obtain the posterior approximation with a diagonal covariance structure by default, which has been shown to be effective and scalable to large-scale classification tasks. 
Here, we compare our method against MC sampling from the posterior to make predictions as done in \citet{shen2024ivon}. 
For the MFVI and LA sampling baselines, we used $1,000$ MC samples in the regression and classification experiments in \cref{sec:regression-classification}, and $50$ MC samples for the ViT and GPT-2 in \cref{sec:vit-exp}.
For our method, we fit an additional scaling factor on the predictive variance by minimising the NLPD on a validation set, similar to the pseudo-count used in \cite{ritter2018scalable}.

\paragraph{Network architectures} 
We experiment with one or two-layer multi-layer perceptron (MLP) on the UCI regression data sets with details given in \cref{app:experiment-regression}.
For MNIST, FMNIST, OrganCMNIST and OrganSMNIST, we use an MLP with layers containing $784-128-64-C$ neurons, where $C$ is the number of classes.
For CIFAR-10/100, DTD, RESISC and ImageNet-R, we fine-tune a Vision Transformer (ViT)~\citep{dosovitskiy2021vit} base model pre-trained on ImageNet-1k~\citep{deng2009imagenet}. 
For the GPT model, we use the pre-trained GPT-2 base model from Hugging Face Transformers~\citep{wolf2019huggingface} and fine-tune it on the respective tasks.

\paragraph{Evaluation metrics} 
For the regression experiments, we measure the negative log predictive density (NLPD) and root-mean-square error (RMSE) for each method. In the classification experiments, we use accuracy (ACC), NLPD, and expected calibration error (ECE) to compare the methods. 
We use a paired $t$-test with $p=0.05$ to bold results with significant statistical differences when reporting the results. 
For assessing out-of-distribution (OOD) robustness, we use a Gaussian kernel density estimator with a variance of 0.25 on the histogram of the predictive entropy evaluated on the test set.

\subsection{Does our Method Provide Useful Uncertainty Estimates?}
\label{sec:regression-classification}

\paragraph{Regression} 
We experiment on a selection of data sets from the UCI repository and run a $5$-fold cross validation to report results for each data set. 
We use either MFVI or LA to obtain the posterior approximation and separately compare our method against their corresponding prediction approaches. 
\cref{table:regression_nlpd} shows our method achieves better NLPD in general than the predictions with sampling for both MFVI and LA. Moreover, our method performs on par with the GLM, even though our method results in a locally linearised network \wrt the inputs. 
Similar conclusions are made inspecting \cref{table:regression_rmse}. \looseness-2

\begin{table*}[!h]
	\centering
	\caption{Negative log predictive density $\downarrow$ on UCI regression data sets. Ours results in better or matching performance compared with sampling and GLM, indicating the effectiveness of our method.}\vspace*{-6pt}
	\setlength{\tabcolsep}{9.5pt}
	\scriptsize
	\begin{tabular}{lc|cH|ccH}
\toprule
& {} & \multicolumn{2}{ c|}{\textit{MFVI (Diagonal Covariance)}} & \multicolumn{3}{c}{\textit{Laplace Approximation (Full Covariance)}} \\[0.2em]
{} & $(n,d)$ &  Sampling     &     Ours       &    Sampling      &       GLM        &      Ours        \\[0.2em]
\hline
    {\sc Servo}      &       (167, 4)       & \val{\bf}{1.287}{0.069} & \val{\bf}{1.136}{0.182} & \val{}{3.795}{0.110} & $\phantom{-}$\val{\bf}{1.047}{0.172} & $\phantom{-}$\val{}{1.443}{0.077} \\
    {\sc LD} &       (345, 5)       & \val{\bf}{1.346}{0.280} & \val{\bf}{1.369}{0.440} & \val{}{2.221}{0.110} & $\phantom{-}$\val{\bf}{1.495}{0.580} & $\phantom{-}$\val{\bf}{1.474}{0.648} \\
   {\sc AM}    &       (398, 7)       & \val{}{1.004}{0.052} & \val{\bf}{0.807}{0.087} & \val{}{1.812}{0.065} & $\phantom{-}$\val{\bf}{0.492}{0.279} & $\phantom{-}$\val{\bf}{0.478}{0.309} \\
   {\sc REV} &       (414, 6)       & \val{}{1.076}{0.059} & \val{\bf}{0.925}{0.091} & \val{}{1.932}{0.045} & $\phantom{-}$\val{}{0.859}{0.129} & $\phantom{-}$\val{\bf}{0.833}{0.156} \\
 {\sc FF}  &      (517, 12)       & \val{\bf}{2.160}{3.003} & \val{\bf}{2.333}{3.671} & \val{}{2.086}{0.292} & $\phantom{-}$\val{\bf}{1.584}{0.950} & $\phantom{-}$\val{\bf}{1.596}{1.217} \\
{\sc ITT} &      (1020, 33)      & \val{}{0.937}{0.047} & \val{\bf}{0.841}{0.065} & \val{}{1.681}{0.069} & $\phantom{-}$\val{}{0.825}{0.095} & $\phantom{-}$\val{\bf}{0.756}{0.164} \\
{\sc CCS} &      (1030, 8)       & \val{}{0.939}{0.068} & \val{\bf}{0.828}{0.108} & \val{}{1.612}{0.048} & $\phantom{-}$\val{}{0.319}{0.109} & $\phantom{-}$\val{\bf}{0.234}{0.161} \\
{\sc ASN} &      (1503, 5)       & \val{}{0.962}{0.054} & \val{\bf}{0.899}{0.065} & \val{}{1.788}{0.045} & $\phantom{-}$\val{}{0.422}{0.109} & $\phantom{-}$\val{\bf}{0.396}{0.133} \\
{\sc CAC} &     (1994, 127)      & \val{}{0.973}{0.092} & \val{\bf}{0.920}{0.118} & \val{}{1.848}{0.055} & $\phantom{-}$\val{\bf}{1.281}{0.069} & $\phantom{-}$\val{}{2.662}{1.096} \\
{\sc PT} &      (5875, 19)      & \val{}{0.976}{0.069} & \val{\bf}{0.940}{0.074} & \val{}{0.984}{0.101} & $\phantom{-}$\val{\bf}{0.576}{0.181} & $\phantom{-}$\val{}{0.651}{0.306} \\
{\sc CCPP} &      (9568, 4)       & \val{}{0.365}{0.040} & \val{\bf}{0.352}{0.042} & \val{}{1.345}{0.085} & \val{\bf}{-0.062}{0.182} & \val{\bf}{-0.062}{0.200} \\
\midrule
\multicolumn{2}{ c|}{Bold Count } & $3/11$ & $11/11$ & $0/11$ & $7/11$ & $8/11$ \\
\bottomrule
\end{tabular}

	\label{table:regression_nlpd}
\end{table*}

\paragraph{Classification} 
Here, we assess our method on MNIST-like classification tasks.
For the LA, we use KFAC approximation of the covariance to reduce the memory overhead. 
In \cref{table:classification_mlp}, we report the ACC and NLPD with their standard errors and the ECE for each method.  
Our method achieves similar ACC with the baselines, while outperforming them on the NLPD and ECE metrics. 
In \cref{app:experiment-classification}, we assess our method on robustness to OOD data. We evaluate an MLP trained on MNIST on rotated versions of the test set. 
Our method consistently reduces overconfidence on OOD data, \cf, \cref{fig:rmnist-ood}. 

\begin{table*}[!t]
	\centering
	\caption{Performance metrics on the MNIST-like data sets for each method with the standard error for ACC and NLPD. Our method achieves better or on par NLPD and ECE than the baselines.}\vspace*{-6pt}
    \setlength{\tabcolsep}{10pt}
	\scriptsize
	\begin{tabular}{l|l|cccc}
\toprule
Metrics & Methods & \sc MNIST & \sc FMNIST  & \sc OrganCMNIST  & \sc OrganSMNIST \\
\hline
   & LA Sampling & \val{\bf}{0.982}{0.133} & \val{\bf}{0.888}{0.316} & \val{\bf}{0.758}{0.428} & \val{\bf}{0.590}{0.492} \\
& LA GLM & \val{}{0.980}{0.139} & \val{\bf}{0.886}{0.317} & \val{}{0.750}{0.433} & \val{}{0.580}{0.494} \\
  \rowcolor{highlight}\cellcolor{white}ACC $\uparrow$
  & LA Ours & \val{}{0.980}{0.139} & \val{}{0.884}{0.321} & \val{}{0.752}{0.432} & \val{}{0.579}{0.494}\\ \cdashline{2-6} \rule{0pt}{2.25ex}
 & MFVI Sampling & \val{\bf}{0.981}{0.136} & \val{\bf}{0.890}{0.313} & \val{}{0.751}{0.432} & \val{\bf}{0.592}{0.491} \\
  \rowcolor{highlight}\cellcolor{white}
  & MFVI Ours & \val{}{0.980}{0.139} & \val{\bf}{0.891}{0.312} & \val{}{0.751}{0.432} & \val{}{0.574}{0.495} \\ 
\hline
 & LA Sampling & \val{\bf}{0.061}{0.413} & \val{}{0.369}{1.197} & \val{}{1.089}{2.581} & \val{}{1.403}{2.127} \\
 & LA GLM & \val{}{0.065}{0.453} & \val{}{0.367}{1.125} & \val{}{1.052}{2.217} & \val{}{1.498}{2.105} \\
  \rowcolor{highlight}\cellcolor{white}NLPD $\downarrow$
  & LA Ours & \val{}{0.063}{0.389} & \val{\bf}{0.331}{0.783} & \val{\bf}{0.793}{1.286} & \val{\bf}{1.277}{1.491}\\
\cdashline{2-6} \rule{0pt}{2.25ex}
 & MFVI Sampling & \val{\bf}{0.062}{0.411} & \val{\bf}{0.311}{0.780} & \val{\bf}{0.797}{1.339} & \val{\bf}{1.182}{1.410} \\
  \rowcolor{highlight}\cellcolor{white}
 & MFVI Ours & \val{\bf}{0.062}{0.392} & \val{\bf}{0.313}{0.795} & \val{\bf}{0.796}{1.343} & \val{}{1.226}{1.417} \\
\hline
\multirow{5}{*}{ece} & LA Sampling & $0.004$ & $0.038$ & $0.101$ & $0.100$ \\
  & LA GLM & $0.005$ & $0.043$ & $0.130$ & $0.189$ \\
 \rowcolor{highlight}\cellcolor{white}ECE $\downarrow$
 & LA Ours & $\bf0.003$ & $\bf0.006$ & $\bf0.024$ & $\bf0.095$\\ \cdashline{2-6} \rule{0pt}{2.25ex}
 & MFVI Sampling & $0.004$ & $\bf0.010$ & $0.023$ & $\bf0.052$ \\
  \rowcolor{highlight}\cellcolor{white}
 & MFVI Ours & $\bf0.003$ & $0.013$ & $\bf0.020$ & $0.075$ \\
\bottomrule
\end{tabular}
    
	\label{table:classification_mlp}
\end{table*}

\paragraph{Ours vs.\ moment-matching} 
To verify the viability of local linearisation, we compare our method against moment-matching (MM) used in \citet{wu2018deterministic}. We apply MM instead of local linearisation to our setting, assuming a diagonal posterior approximation from LA. 
In \cref{table:classification_moment_matching}, we show the results for our method against MM on MNIST-like classification tasks. 
Our approach outperforms MM across the data sets and the metrics, except for the ECE on OrganCMNIST.
Compared to MM, our method is applicable to any differentiable activation function and any type of stable distribution~\citep{peterse2024stable}, while MM requires tailored derivations for each case and, hence, is less \emph{plug-and-play}.

\begin{table*}[!t]
	\centering
	\caption{Performance comparison between our method and Moment-Matching (MM) on the MNIST-like classification tasks. We report the ACC and NLPD with standard errors and the ECE. Our method outperforms MM despite being simpler and more applicable to various distributions. }\vspace*{-6pt}
    \setlength{\tabcolsep}{8.5pt}
	\scriptsize
	\begin{tabular}{l|l|cccc}
\toprule
Metrics & Methods & \sc MNIST & \sc FMNIST  & \sc OrganCMNIST  & \sc OrganSMNIST \\
\hline
     & MM & \val{}{0.949}{0.219} & \val{\bf}{0.876}{0.330} & \val{}{0.726}{0.446} & \val{}{0.604}{0.489} \\
     \rowcolor{highlight}\cellcolor{white}ACC $\uparrow$
 & Ours  & \val{\bf}{0.976}{0.153} & \val{\bf}{0.878}{0.327} & \val{\bf}{0.858}{0.349} & \val{\bf}{0.712}{0.453} \\
 \hline
& MM & \val{}{0.867}{0.371} & \val{}{0.429}{0.615} & \val{}{0.920}{0.737} & \val{}{1.141}{0.767} \\
  \rowcolor{highlight}\cellcolor{white}NLPD $\downarrow$
 & Ours & \val{\bf}{0.095}{0.353} & \val{\bf}{0.361}{0.855} & \val{\bf}{0.515}{0.831} & \val{\bf}{0.808}{0.836} \\
 \hline
& MM & $0.498$ & $0.104$ & $0.190$ & $0.131$ \\
 \rowcolor{highlight}\cellcolor{white}ECE $\downarrow$
& Ours & $\bf0.027$ & $\bf0.011$ & $\bf0.099$ & $\bf0.076$ \\
\bottomrule
\end{tabular}
    
	\label{table:classification_moment_matching}
\end{table*}

\subsection{Is our Method Scalable?}
\label{sec:vit-exp}
We demonstrate that our method is applicable to large-scale networks by experimenting with pre-trained ViT and GPT-2 models. 
In particular, we experiment with applying our method on either the attention layer or the MLP after the attention layer in the last two (ViT) / four (GPT) transformer blocks. 
For each target data set, we fine-tune the layers we obtain the posterior approximation for. 
We assume a diagonal posterior to reduce the memory overhead.   
\cref{tab:vit_attention} shows the results when fine-tuning ViT models and obtaining the posterior approximation from the attention layers. 
We observe that our method achieves better or on par NLPD and ECE compared to the baselines for both LA and MFVI across all data sets while maintaining similar ACC as the baselines.

In \cref{tab:gpt} we show the results for a GPT-2 model with LA on the MLP layers. 
We observe that our method systematically outperforms sampling, while achieving similar performance to GLM in some cases.
Thus indicating that our method is applicable to different application domains.
We present additional results for ViT models in \cref{app:experiment}.

\begin{table*}[t]
	\centering
	\caption{Performance metrics using ViT with posterior approximation on the attention layers with the standard error for ACC and NLPD. Our method achieves better NLPD and ECE in general and achieves similar ACC compared to the baselines. 
  }\vspace*{-6pt}
  \setlength{\tabcolsep}{8.5pt}
	\scriptsize
	\begin{tabular}{l|l|ccccc}
\toprule
Metrics & Methods & \sc CIFAR-10 & \sc CIFAR-100  & \sc DTD  & \sc RESISC & \sc Imagenet-R \\
\hline
    \multirow{5}{*}{acc}  & LA Sampling & \val{\bf}{0.974}{0.002} & \val{\bf}{0.877}{0.003} & \val{\bf}{0.720}{0.010} & \val{\bf}{0.906}{0.004} & \val{\bf}{0.721}{0.012} \\
 & LA GLM & \val{\bf}{0.975}{0.002} & \val{\bf}{0.878}{0.003} & \val{\bf}{0.725}{0.010} & \val{\bf}{0.905}{0.004} & \val{\bf}{0.727}{0.012} \\
  \rowcolor{highlight}\cellcolor{white} ACC $\uparrow$
 & LA Ours & \val{\bf}{0.974}{0.002} & \val{\bf}{0.879}{0.003} & \val{\bf}{0.725}{0.010} & \val{\bf}{0.906}{0.004} & \val{\bf}{0.725}{0.012}\\ \cdashline{2-7} \rule{0pt}{2.25ex}
 & MFVI Sampling & \val{\bf}{0.975}{0.002} & \val{\bf}{0.885}{0.003} & \val{\bf}{0.749}{0.010} & \val{\bf}{0.916}{0.003} & \val{\bf}{0.741}{0.012} \\
 \rowcolor{highlight}\cellcolor{white}
  & MFVI Ours & \val{\bf}{0.975}{0.002} & \val{\bf}{0.885}{0.003} & \val{\bf}{0.740}{0.010} & \val{\bf}{0.917}{0.003} & \val{\bf}{0.736}{0.012} \\
\hline
 & LA Sampling & \val{\bf}{0.099}{0.008} & \val{}{0.520}{0.009} & \val{}{1.361}{0.024} & \val{}{0.390}{0.009} & \val{\bf}{1.262}{0.042} \\
 & LA GLM & \val{}{0.107}{0.008} & \val{}{0.499}{0.009} & \val{}{1.297}{0.025} & \val{\bf}{0.325}{0.009} & \val{\bf}{1.210}{0.041} \\
  \rowcolor{highlight}\cellcolor{white} NLPD $\downarrow$
 & LA Ours & \val{\bf}{0.087}{0.005} & \val{\bf}{0.426}{0.011} & \val{\bf}{0.981}{0.030} & \val{\bf}{0.297}{0.011} & \val{\bf}{1.192}{0.042}\\\cdashline{2-7} \rule{0pt}{2.25ex}
 & MFVI Sampling & \val{}{0.134}{0.011} & \val{}{0.568}{0.020} & \val{\bf}{0.967}{0.040} & \val{}{0.314}{0.016} & \val{\bf}{1.135}{0.051} \\
 \rowcolor{highlight}\cellcolor{white}
 & MFVI Ours & \val{\bf}{0.083}{0.005} & \val{\bf}{0.451}{0.012} & \val{\bf}{0.909}{0.032} & \val{\bf}{0.272}{0.011} & \val{\bf}{1.080}{0.046} \\
\hline
 & LA Sampling & $0.009$ & $0.114$ & $0.308$ & $0.113$ & $0.133$ \\
 & LA GLM & $0.015$ & $0.081$ & $0.275$ & $0.042$ & $0.103$ \\
  \rowcolor{highlight}\cellcolor{white} ECE $\downarrow$
 & LA Ours & $\bf0.007$ & $\bf0.017$ & $\bf0.041$ & $\bf0.013$ & $\bf0.089$\\  \cdashline{2-7} \rule{0pt}{2.25ex}
 & MFVI Sampling & $0.016$ & $0.047$ & $0.032$ & $0.023$ & $\bf0.037$ \\
 \rowcolor{highlight}\cellcolor{white}
 & MFVI Ours & $\bf0.005$ & $\bf0.023$ & $\bf0.031$ & $\bf0.008$ & $\bf0.037$ \\
\bottomrule
\end{tabular}
    
	\label{tab:vit_attention}
\end{table*}

\begin{table*}
	\centering
	\setlength{\tabcolsep}{5.5pt}
	\scriptsize
  \vspace{-3mm}
	\caption{Performance on language understanding tasks.}
  \vspace{-3mm}
  \begin{tabular}{l|l|ccc}
    \toprule
Metrics & Methods & \sc BOOL-Q & \sc WIC  & \sc MRPC  \\
    \hline
    & LA Sampling & \val{}{0.606}{0.009} & \val{}{0.511}{0.020} & \val{}{0.665}{0.011} \\
ACC $\uparrow$     
	 & LA GLM & \val{\bf}{0.678}{0.008} & \val{\bf}{0.630}{0.019} & \val{\bf}{0.720}{0.011} \\
	 	\rowcolor{highlight}\cellcolor{white} 
 	 & LA Ours & \val{}{0.622}{0.008} & \val{}{0.500}{0.020} & \val{}{0.416}{0.012}\\
 	      \hline 
  	& LA Sampling & \val{}{0.656}{0.006} & \val{}{0.716}{0.008} & \val{}{0.658}{0.012} \\
NLPD $\downarrow$   
   & LA GLM & \val{\bf}{0.633}{0.013} & \val{\bf}{0.695}{0.022} & \val{\bf}{0.609}{0.021} \\    
   \rowcolor{highlight}\cellcolor{white} 
 & LA Ours & \val{}{0.665}{0.005} & \val{\bf}{0.694}{0.002} & \val{}{0.706}{0.001}\\
     \hline 
 & LA Sampling & $0.030$ & $0.097$ & $\bf0.080$ \\
 ECE $\downarrow$  
 & LA GLM & $0.090$ & $0.117$ & $0.111$ \\
    \rowcolor{highlight}\cellcolor{white} 
 & LA Ours & $\bf0.027$ & $\bf0.024$ & $0.109$\\
\bottomrule
\end{tabular}
        
  \vspace*{-3mm}
	\label{tab:gpt}
\end{table*}

\begin{figure}[t]
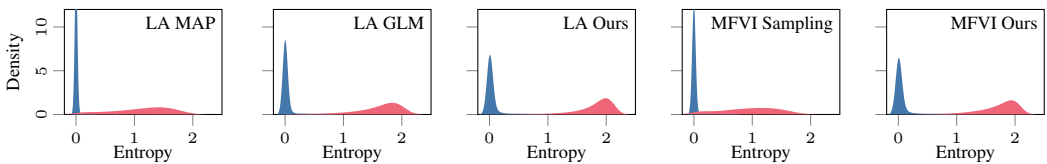

  \centering\scriptsize
  \setlength{\figurewidth}{.15\textwidth}%
  \setlength{\figureheight}{.10\textwidth}
  \pgfplotsset{
    x tick label style={font=\tiny}, 
    y tick label style={rotate=90, font=\tiny}, 
    y label style = {font=\scriptsize}, 
    x label style = {yshift=0.5em}, 
    xlabel={Entropy}, 
    ylabel={Density},
    scale only axis,
    tick align=outside,
    tick pos=left,
    xmin=-0.2, xmax=2.5,%
    ymin=0, ymax=12,%
    }
    \input{fig/ood_la_map.tex}
    \pgfplotsset{ylabel=\empty, yticklabel=\empty,
    }
    \input{fig/ood_la_glm.tex}
    \input{fig/ood_la_dbnn.tex}
    \input{fig/ood_ivon_sample.tex}
    \input{fig/ood_ivon_dbnn.tex}
  \caption{Kernel density plots over the predictive entropy from a ViT network finetuned on CIFAR-10 (blue, in-distribution) and data from SVHN (red, out-of-distribution). Our method results in a clear separation between the in- and out-of-distribution data. 
  }
  \label{fig:vit-ood}
  \end{figure}

We also assess the robustness to out-of-distribution (OOD) data for our method and the baselines. 
In particular, we take the ViT network fine-tuned on CIFAR-10 and evaluate its predictive entropy on the SVHN data set~\citep{netzer2011reading}. 
\cref{fig:vit-ood} shows the kernel density of the predictive entropy computed on the test sets of CIFAR-10 and SVHN, where the model should have high entropy for data different from the training data. 
Although our method is slightly underconfident on the in-distribution data, the entropy for in-distribution and OOD data is clearly separated, especially for MFVI.

\subsection{Can our Method Estimate Input Sensitivies?}
\label{sec:sensitivity-exp}
\begin{figure}
  \centering
  \newlength{\figsize}
  \setlength{\figsize}{0.11\textwidth}

  \def\images{0_2, 0_7, 6_6, 6_7, 8_6, 8_7}

  \begin{tikzpicture}
    \foreach \x [count=\xi] in {0, 0, 6, 6, 8, 8} {
      \draw[gray] (\xi*\figsize,0) -- (\xi*\figsize,1.7*\figsize);
      \node at (\xi*\figsize,1.8*\figsize) {$y=\x$};
    } 
    \foreach \img [count=\xi] in \images {
      \node[inner sep = 0pt] at (\xi*\figsize,\figsize) {\includegraphics[width=\figsize]{imgs/input_perturbation/individual/mnist_0-8-6_map_\img}};
    }
    \foreach \img [count=\xi] in \images {
      \node[inner sep = 0pt] at (\xi*\figsize,0) {\includegraphics[width=\figsize]{imgs/input_perturbation/individual/mnist_0-8-6_bayes_\img}};
    }

    \node[rotate=90] at (0.5, \figsize) {MAP};
    \node[rotate=90] at (0.5, 0) {Bayes};
  \end{tikzpicture}
  \caption{Pixel sensitivity maps of an MLP trained on a subset of MNIST digits (classes $0/6/8$). 
  The rows show sensitivities to pixel perturbations for the MLP (MAP) and MLP with last-layer Laplace approximation (Bayes) respectively. The sensitivities are visualised in the range (0.5~\protect\includegraphics[width=3em,height=.7em]{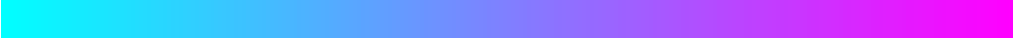}~1.0). 
  The Bayes MLP shows less spurious sensitivities across the pixels compared to MAP. 
  }
  \label{fig:input-sensitivity}
\end{figure}

We demonstrate that our method can estimate sensitivities \wrt the inputs to the network. 
For this, we use a $3$-class MLP trained on the digits $0/6/8$. Our goal is to estimate sensitivity maps by assuming that the input images $\bx \sim \mcN(\bx, \bSigma)$ are distributed according to a Gaussian centred at the pixel values with diagonal covariance $\bSigma$. 
We optimise the input covariance of each image by minimising the loss 
\begin{equation}\label{eq:sensitivity-loss}
  \ell = \textstyle\sum^N_{n=1} \operatorname{cross-entropy}(f(\bx_n), y_n) - \ent{\mcN(\bx_n, \Sigma_n)} .
\end{equation} 
In words, we jointly minimise the cross-entropy loss, after analytically propagating the input distribution through the network, while maximising the entropy $\ent{\mcN(\bx_n, \Sigma_n)}$ of the input distribution. 
The optimisation is stopped once the difference in NLPD between the current iteration and initial condition is more than $0.1$. 
\cref{fig:input-sensitivity} shows examples of the resulting sensitivity maps for a deterministic MLP (MAP) 
and the same MLP with last-layer LA (Bayes). 
We observe that the largest sensitivity for the digits $0$ and $8$ are generally in the middle, while for $6$ in the upper right corner. 
The Bayes model shows less spurious sensitivities across the pixels compared to the MAP model.
Thus, indicating that incorporating all sources of uncertainties can lead to a more interpretable sensitivity analysis.

\section{Discussion \& Conclusion}
In this work, we proposed to streamline prediction in Bayesian deep learning through local linearisation and local Gaussian approximations of the network.
For this, we discussed the propagation in different neural network architectures and covariance structures.
In particular, we discussed how to handle Kronecker-factorised posterior covariances and transformer architectures.
We showed through a series of experiments that our method obtains high predictive performance, provides useful predictive uncertainties, and can be used for sensitivity analysis.
Our method helps to make BDL more useful in practice and expands the use cases and sources of uncertainties that can be considered.

In future work, we aim to apply our approach to tasks with a larger number of output classes, explore additional use-case scenarios in which our streamlined approach can be beneficial, and scale to even larger networks.
Moreover, we aim to investigate further the computational benefits obtained by exploiting the posterior covariance structure and sparsity in the network.

\paragraph{Limitations} 
The local linearisation of activation functions induces an error that depends on both the activation function and the location and scale of the distribution over the input to the activation function. 
Moreover, we assume independence between the activations and model parameters for the local Gaussian approximation in linear layers and residual connections, which may incur a loss of information in the propagation. 
Especially, the independence assumption in the residual block is potentially harmful, and relaxing it would be a valuable future direction.
Further, it would be interesting to estimate the induced approximation error to identify potential failure modes.  
Finally, we assume access to a validation set for fitting the scaling factor of the predictive posterior distribution, which is currently done using a grid search. An interesting future step is the use of the marginal likelihood to optimise the scaling factor.

\section*{Acknowledgments}
AS and RL acknowledge funding from the Research Council of Finland (grant number 339730, 362408). MT acknowledges funding from the Research Council of Finland (grant number 347279). MK acknowledges funding from the Finnish Center for Artificial Intelligence (FCAI). We acknowledge CSC -- IT Center for Science, Finland, for awarding this project access to the LUMI supercomputer, owned by the EuroHPC Joint Undertaking, hosted by CSC (Finland) and the LUMI consortium through CSC. We acknowledge the computational resources provided by the Aalto Science-IT project. Lastly, we thank Jonas Vestergaard for finding a bug in our code and his feedback on the manuscript.

{

    \bibliographystyle{iclr2025_conference}
}

\clearpage
\appendix

\section*{Appendices}

The appendices are structured as follows: 
\cref{app:derivation} presents the derivations of our method in detail. \cref{app:experiment} describes the experimental setup and additional experimental results.

\section{Derivations} \label{app:derivation}

In this section, we derive how to propagate the distribution deterministically. See \cref{table:notation} for the list of notations that will be used throughout this section. 

We first derive the general result in \cref{app:derivation-full} where the posterior covariance has a full structure in the linear layer and evaluate the quality of local Gaussian approximation in \cref{app:aw-approx}.
Next, in \cref{app:derivation-diag} and \cref{app:derivation-kron} we give the derivation for diagonal and KFAC covariance, respectively.
Then, \cref{app:derivation-activation} shows the derivation for activation functions.
Finally, \cref{app:derivation-transformer} describes how we apply our method in a transformer network~\citep{vaswani2017attention}.

\begin{table}[!ht]
	\centering
	\renewcommand*{\arraystretch}{1.1}
	\setlength{\tabcolsep}{5.5pt}
	\caption{Notation.}
  \vspace{-3mm}
  \begin{tabular}{ll}
    \toprule
    $\bx$ & lowercase bolder letter, vector \\
    $\bW$ & uppercase bold letter, matrix \\
    $\mcD$ & set \\
    $x_i$ & $i$\textsuperscript{th} element of $\bx$ \\
    $W_{ki}$ & $k$\textsuperscript{th} row, $i$\textsuperscript{th} column of $\bW$ \\
    $\bW[k,:]$ & $k$\textsuperscript{th} row of a matrix \\
    $k$, $l$  & dimension of the output \\
    $i$, $j$  & dimension of the input \\
    $d$ & data feature dimension \\
    $n, N$ & number of data points \\
    $C$ & total number of classes \\
    $m$ & layer index \\
    \bottomrule
  \end{tabular}
  \label{table:notation}
\end{table}

\subsection{Derivation for Full Covariance Structure}\label{app:derivation-full}
Denote the weight and bias of the $m$\textsuperscript{th} linear layer as $\bW^{(m)} \in \reals^{\Dout \times \Din}$ and $\bb^{(m)} \in \reals^{\Dout}$ respectively, and its input as $\ba^{(m-1)} \in \reals^{\Din}$. 
The pre-activation is then given as $\bh^{(m)} = \bW^{(m)}\ba^{(m-1)} + \bb^{(m)}$ with its $k$\textsuperscript{th} element being $h^{(m)}_k= \sum_{i=1}^{\Din} W^{(m)}_{ki} a^{(m-1)}_i  + b^{(m)}_k$.

We make the following assumptions to obtain a tractable distribution on the pre-activation:
\begin{itemize}[noitemsep]
    \item Assumption 1: We assume each $a^{(m-1)}_iW^{(m)}_{ki}$ is a Gaussian distribution. 
    \item Assumption 2: We assume that the activations of the previous layer $a^{(m-1)}_i$ and parameters of the $m$\textsuperscript{th} layer are independent.
\end{itemize} 
From assumption 1, because now $a^{(m-1)}_iW^{(m)}_{ki}$ and $b^{(m)}_k$ are all Gaussian distributions, $h_k^{(m)}$ will follow Gaussian distribution as well.
We call this local Gaussian approximation as we approximate each local component $a^{(m-1)}_iW^{(m)}_{ki}$ with a Gaussian.
As now each $h_k^{(m)}$ is a Gaussian, $\bh^{(m)}$ will be jointly Gaussian.
We derive its mean and covariance and drop the layer index if it is clear from the context.

\paragraph{Derivation of mean} As $a_i$ is assumed to be uncorrected with $W_{ki}$, we have 
\[
	\expect{h_k} &= \expect{\sum_{i=1}^{\Din} W_{ki} a_i  + b_k} \\
				 &= \sum_{i=1}^{\Din} \expect{W_{ki} a_i  + b_k} \\
				 &=\sum_{i=1}^{\Din} \expect{W_{ki} a_i} +  \expect{b_k} \\
				 &\approx \sum_{i=1}^{\Din} \expect{W_{ki}}\expect{a_i} + \expect{b_k} \tag{Assumption 2}.
\]

\paragraph{Derivation of covariance} The covariance between the  $k$\textsuperscript{th} and $l$\textsuperscript{th} pre-activation can be written as
\[
  \cov{h_k}{h_l} &= \cov{\sum_{i=1}^{\Din} a_i W_{ki} + b_k}{\sum_{i=1}^{\Din} a_i W_{li} + b_l} \\
  				&= \cov{\sum_{i=1}^{\Din} a_i W_{ki}}{\sum_{i=1}^{\Din} a_i W_{li}} + \cov{\sum_{i=1}^{\Din} a_i W_{ki}}{b_l} + \cov{\sum_{i=1}^{\Din} a_i W_{li}}{b_k}  \nonumber\\
  				& \hspace{10em} + \cov{b_k}{b_l} \\
  				&= \sum_{1\leq i, j \leq D_{\text{in}}} \cov{a_iW_{ki}}{a_jW_{lj}} + \sum_{1\leq i \leq D_{\text{in}}} (\cov{a_i W_{ki}}{b_l} + \cov{a_i W_{li}}{b_k})  \nonumber\\
  				& \hspace{10em}  + \cov{b_k}{b_l}
\]
We first derive the form of $\cov{a_i W_{ki}}{a_i W_{li}}$:
\begingroup
\setlength{\jot}{10pt}
\[
    & \cov{a_iW_{ki}}{a_jW_{lj}} \nonumber \\
    & \quad= \expect{(a_iW_{ki} - \expect{a_iW_{ki}})(a_jW_{lj} - \expect{a_jW_{lj}})}  \\
    &\quad = \expect{a_iW_{ki}a_jW_{lj}  - a_iW_{ki}\expect{a_jW_{lj}} - \expect{a_iW_{ki}}a_jW_{lj} + \expect{a_iW_{ki}}\expect{a_jW_{lj}}} \\
    &\quad= \expect{a_ia_jW_{ki}W_{lj}}- \expect{a_iW_{ki}}\expect{a_jW_{lj}} - \expect{a_iW_{ki}}\expect{a_jW_{lj}} + \expect{a_iW_{ki}}\expect{a_jW_{lj}} \\
    &\quad \approx \expect{a_ia_j}\expect{W_{ki}W_{lj}}- \expect{a_i}\expect{W_{ki}}\expect{a_j}\expect{W_{lj}} \tag{Assumption 2} \\
    &\quad= ( \expect{a_i}\expect{a_j} + \cov{a_i}{a_j}) ( \expect{W_{ki}}\expect{W_{lj}} + \cov{W_{ki}}{W_{lj}}) \nonumber \\
    &\hspace{20em} - \expect{a_i}\expect{W_{ki}}\expect{a_j}\expect{W_{lj}}  \\
    &\quad= \expect{a_i}\expect{a_j}\cov{W_{ki}}{W_{lj}} + \expect{W_{ki}}\expect{W_{lj}} \cov{a_i}{a_j} + \cov{a_i}{a_j}\cov{W_{ki}}{W_{lj}}. \label{eq:cov-derivation}
\]
\endgroup

Then we derive the form of $\cov{a_i W_{ki}}{b_l}$:
\begingroup
\setlength{\jot}{10pt}
\[
    \cov{a_iW_{ki}}{b_l} &= \expect{(a_iW_{ki} - \expect{a_iW_{ki}})(b_l - \expect{b_l})} \\
                            &\approx \expect{(a_iW_{ki} - \expect{a_i} \expect{W_{ki}})(b_l - \expect{b_l})} \tag{Assumption 2}\\
                            &= \expect{a_iW_{ki}b_l - a_iW_{ki}\expect{b_l} - \expect{a_i} \expect{W_{ki}} b_l + \expect{a_i} \expect{W_{ki}} \expect{b_l}}\\
                            &=  \expect{a_iW_{ki}b_l} - \expect{a_i} \expect{W_{ki}} \expect{b_l} \\
                            &\approx \expect{a_i} \expect{W_{ki}b_l} - \expect{a_i} \expect{W_{ki}} \expect{b_l} \tag{Assumption 2}\\
                            &= \expect{a_i}( \expect{W_{ki}}\expect{b_l}+\cov{W_{ki}}{b_l}) -\expect{a_i} \expect{W_{ki}} \expect{b_l} \\
                            &=  \expect{a_i}\cov{W_{ki}}{b_l}.
\]
\endgroup

Putting it together, we have $\cov{h_k}{h_l} =$
\[
	 \sum_{\mathclap{1\leq i,j \leq \Din}} \cov{a_iW_{ki}}{a_jW_{lj}} + \sum_{i=1}^{\Din} \rbra{ \expect{a_i}\cov{W_{ki}}{b_l} + \expect{a_i}\cov{W_{li}}{b_k} } + \cov{b_k}{b_l}, 
\label{eq: cov-sum-aw}
\]

where $\cov{a_iW_{ki}}{a_jW_{lj}}=$
\[
   \expect{a_i}\expect{a_j}\cov{W_{ki}}{W_{lj}} +\expect{W_{ki}} \expect{W_{lj}} \cov{a_i}{a_j} + \cov{a_i}{a_j}\cov{W_{ki}}{W_{lj}}.
\]

\subsection{Error Induced Through Local Gaussian Approximation}\label{app:aw-approx}
In this section, we analyse the error induced by the local Gaussian approximation.
Recall that we made these two assumptions for the derivation:
\begin{itemize}
    \item Assumption 1: We assume $a^{(m-1)}_iW^{(m)}_{ki}$ is a Gaussian distribution. 
    \item Assumption 2: We assume that the activations of the previous layer $a^{(m-1)}_i$ and parameters of the $m$\textsuperscript{th} layer are independent.
\end{itemize} 

We first examine the error induced by A2 on the moments for $a_iW_{ki}$. 
Given two correlated univariate Gaussian $x_1$ and $x_2$, with the joint being
\[
\left[\begin{array}{l}
x_1 \\
x_2
\end{array}\right] \sim \mathcal{N}\left(\left[\begin{array}{l}
\expect{x_1} \\
\expect{x_2}
\end{array}\right],\left[\begin{array}{cc}
\sigma_{x_1}^2 & \cov{x_1}{x_2} \\
\cov{x_1}{x_2} & \sigma_{x_2}^2
\end{array}\right]\right),
\]
from \citet{nadarajah2016gaussian-product, kan2008moments}, although the distribution form of $x_1x_2$ is no longer Gaussian and intractable, its mean and variance can be computed analytically as
\[
\expect{x_1x_2} &= \expect{x_1}\expect{x_2} + \cov{x_1}{x_2},  \\
\var{x_1x_2} &= \sigma_1^2\sigma_2^2 + \sigma_1^2\expect{x_2}^2 + \sigma_2^2\expect{x_1}^2 + (\sigma_1^2\sigma_2^2+2\expect{x_1}\expect{x_2})\cov{x_1}{x_2}.
\]

Applying the above result in our case, we have %
\[
 &\expect{{a_i}W_{ki}}= \expect{a_i}\expect{W_{ki}} + \cov{a_i}{W_{ki}}, \\
 &\var{a_iW_{ki}} = \sigma_{a_i}^2\sigma_{W_{ki}}^2 + \sigma_{a_i}^2\expect{W_{ki}}^2 + \sigma_{W_{ki}}^2\expect{a_i}^2  \nonumber\\
 &\hspace{15em} + (2\expect{a_i}\expect{W_{ki}} + \sigma_{a_i}^2\sigma_{W_{ki}}^2) \cov{a_i}{W_{ki}}.
\]

As $\cov{a_i}{W_{ki}}$ is intractable, in A2 we ignore the correlation between $a_i$ and $W_{ki}$, which results in 
\[
 &\expect{{a_i}W_{ki}}\approx \expect{a_i}\expect{W_{ki}} \cancel{{\color{gray}+ \cov{a_i}{W_{ki}}}}, \\
 &\var{a_iW_{ki}} \approx \sigma_{a_i}^2\sigma_{W_{ki}}^2 + \sigma_{a_i}^2\expect{W_{ki}}^2 + \sigma_{W_{ki}}^2\expect{a_i}^2 \nonumber\\
 &\hspace{15em} \cancel{{\color{gray}+ (2\expect{a_i}\expect{W_{ki}} + \sigma_{a_i}^2\sigma_{W_{ki}}^2) \cov{a_i}{W_{ki}}}}. \label{eq:aw-var}
\]

Note that in the case of diagonal posterior covariance, as each parameter is independent of each other, A2 holds automatically. In this case, we recover the correct mean and variance for $a_iW_{ki}$.

Now, we examine the error induced by A1 and A2 through Monte Carlo estimation.
\cref{fig:aw_sample} provides a simulation result illustrating the error induced by the local Gaussian approximation on $a_iW_{ki}$.
We plot the results for weights with the largest absolute magnitude of an MLP trained on MNIST.
This approximation works well in practice but fails to capture the potential skewness of the
distributions. 

\begin{figure}[h!]
  \centering\scriptsize
  \setlength{\figurewidth}{.18\textwidth}
  \setlength{\figureheight}{.18\textwidth}
  \pgfplotsset{tick align=inside, x tick label style={font=\tiny}, y tick label style={rotate=90, font=\tiny}, y label style = {yshift=-2.5em, font=\scriptsize}, x label style = {yshift=0.5em}, scale only axis, legend style = {font = \tiny}, xlabel = {pre-activation value}}
  \begin{subfigure}[b]{\figurewidth}
     \centering
       \input{fig/local_gaussian_approx_1.tex}
     \end{subfigure}
  \begin{subfigure}[b]{\figurewidth}
     \centering
       \input{fig/local_gaussian_approx_2.tex}
     \end{subfigure}
  \begin{subfigure}[b]{\figurewidth}
     \centering
       \input{fig/local_gaussian_approx_3.tex}
  \end{subfigure}
  \begin{subfigure}[b]{\figurewidth}
     \centering
       \input{fig/local_gaussian_approx_4.tex}
  \end{subfigure}
  \begin{subfigure}[b]{\figurewidth}
     \centering
       \input{fig/local_gaussian_approx_5.tex}
  \end{subfigure}
  \caption{Comparison between Monte-Carlo estimates \protect\tikz[baseline=-0.5ex]{\protect\draw[thick,-](0,0)--(0.5,0);} of the distribution over $a_iW_{ki}$ and our analytic Gaussian approximation \protect\tikz[baseline=-0.5ex]{\protect\draw[thick,dashed,-, scRed](0,0)--(0.5,0);}.}
  \label{fig:aw_sample}
\end{figure}

\clearpage
\subsection{Derivation for Diagonal Covariance Structure}\label{app:derivation-diag}
When the posterior has diagonal covariance, the mean $\expect{h_k}$ will still be the same.

For covariance, when $k \neq l$ we have $\cov{h_k}{h_l} =$
\[
\cov{h_k}{h_l} &= \textcolor{gray!50}{\cov{b_k}{b_l}} +\sum_{i=1}^{\Din} \expect{a_i}\textcolor{gray!50}{\cov{W_{ki}}{b_l}} + \sum_{j=1}^{\Din} \expect{a_j}\textcolor{gray!50}{\cov{W_{lj}}{b_k}} \nonumber \\
			&\hspace{3em} +\sum_{i=1}^{\Din} \sum_{j=1}^{\Din}  \expect{a_i}\expect{a_j}\textcolor{gray!50}{\cov{W_{ki}}{W_{lj}}} \nonumber \\
			&\hspace{3em} +\sum_{i=1}^{\Din} \sum_{j=1}^{\Din} \expect{W_{ki}}\expect{W_{lj}} \cov{a_i}{a_j} \nonumber \\
			&\hspace{3em} +\sum_{i=1}^{\Din} \sum_{j=1}^{\Din} \cov{a_i}{a_j}\textcolor{gray!50}{\cov{W_{ki}}{W_{lj}}} \nonumber \\
			&= \sum_{i=1}^{\Din} \sum_{j=1}^{\Din} \expect{W_{ki}}\expect{W_{lj}} \cov{a_i}{a_j} 
\]

For $k=l$, we have $\var{h_k} =$
\[
\var{h_k} &= \cov{\sum_{i=1}^{\Din} a_i W_{ki}+b_k}{\sum_{j=1}^{\Din} a_j W_{kj}+b_k}\\
&=\var{b_k} +\sum_{i=1}^{\Din} \expect{a_i}\textcolor{gray!50}{\cov{W_{ki}}{b_k}} + \sum_{j=1}^{\Din} \expect{a_j}\textcolor{gray!50}{\cov{W_{kj}}{b_k}} \nonumber \\
			&\hspace{3em} +\sum_{i=1}^{\Din} \sum_{j=1}^{\Din}  \expect{a_i}\expect{a_j}\cov{W_{ki}}{W_{kj}} \nonumber \\
			&\hspace{3em} +\sum_{i=1}^{\Din} \sum_{j=1}^{\Din} \expect{W_{ki}}\expect{W_{kj}} \cov{a_i}{a_j} \nonumber \\
			&\hspace{3em} +\sum_{i=1}^{\Din} \sum_{j=1}^{\Din} \cov{a_i}{a_j}\cov{W_{ki}}{W_{kj}} \nonumber \\
&=\var{b_k} \nonumber \\
			&\hspace{3em} +\sum_{i=1}^{\Din}  \expect{a_i}^2 \var{W_{ki}} \nonumber \\
			&\hspace{3em} +\sum_{i=1}^{\Din} \sum_{j=1}^{\Din} \expect{W_{ki}}\expect{W_{kj}} \cov{a_i}{a_j} \nonumber \\
			&\hspace{3em} +\sum_{i=1}^{\Din} \var{a_i}\var{W_{ki}}  \nonumber \\
&\approx \var{b_k} +\sum_{i=1}^{\Din}  \expect{a_i}^2 \var{W_{ki}} +\sum_{i=1}^{\Din} \expect{W_{ki}}^2 \var{a_i} +\sum_{i=1}^{\Din} \var{a_i}\var{W_{ki}} \tag{drop correlation for $\ba$}
\]

We drop the correlation between pre-activation for faster compute.

\subsection{Derivation for Kronecker Covariance Structure}\label{app:derivation-kron}

In KFAC when using column convention, the Hessian is represented in Kronecker product form $\hessian = \bA \kron \bB$.
Denote the prior precision as $\lambda^2$, then the posterior covariance is 
\begin{equation}
	\bSigma = (\hessian + \lambda^2 \identity)^{-1} = (\bA \kron \bB + \lambda^2 \identity)^{-1}
\end{equation}

As there is no closed form for the inverse, to express the covariance in the form of the Kronecker product as well, we approximate the covariance as
\begin{align}
	\bSigma &= (\bA \kron \bB + \lambda^2 \identity)^{-1} \\
	 &= \left[(\bU_A\eigval_A\bU_A^\top) \kron  (\bU_B\eigval_B\bU_B^\top) + \lambda^2 \identity \right]^{-1} \tag{Eigen Decomposition}\\
	 &\approx \left[(\bU_A(\eigval_A + \lambda\identity_A)\bU_A^\top)\kron (\bU_B(\eigval_B + \lambda\identity_B)\bU_B^\top) \right]^{-1} \\
	 &= \underbracket{\left(\bU_A(\eigval_A + \lambda\identity_A)\bU_A^\top\right)^{-1}}_{\bC} \kron  \underbracket{\left(\bU_B(\eigval_B + \lambda\identity_B)\bU_B^\top\right)^{-1}}_{\bD}. \tag{$(\bA \kron \bB)^{-1} = \bA^{-1} \kron \bB^{-1}$}
\end{align} 
Our method requires retrieve the covariance between the $k\textsuperscript{th}$ row of weight and $l\textsuperscript{th}$ row of weight, which is a $\Din \times \Din$ matrix:
\[
\cov{\bW[k,:]}{\bW[l, :]} = \left[\begin{array}{ccc}
\cov{W_{k1}}{W_{l1}} & \ldots & \cov{W_{k1}}{W_{l\Din}} \\
\vdots & \ddots & \vdots  \\
\cov{W_{k\Din}}{W_{l1}} & \ldots & \cov{W_{k\Din}}{W_{l\Din}}
\end{array}\right].
\]

As $\bSigma \approx \bC \kron \bD$ where $\bC \in \mathbb{\Din \times \Din}$ and $\bD \in \mathbb{\Dout \times \Dout}$, the posterior covariance is represented by a total number of $\Din \times \Din$ matrix with size $\Dout \times \Dout$. 
Retrieving a $\Din \times \Din$ matrix from it is not trivial.
In the toy example as shown in \cref{fig: kron-retrieve-app}, for a $\Din=3$ and $\Dout=2$ matrix $\bW$, its covariance is represented by a total number of $9$~$(\Din \times \Din)$ matrix \RNum{1}, \RNum{2}, $\ldots$, \RNum{9} with shape $2 \times 2$~$(\Dout \times \Dout)$.
To retrieve $\cov{\bW[1,:]}{\bW[2, :]}$,  we need to first decide which Kronecker blocks contain it (in this case block \RNum{2}, \RNum{3}, \RNum{5} and \RNum{6}) and reconstruct these Kronecker blocks.
Then, we retrieve $\cov{\bW[1,:]}{\bW[2, :]}$ from the reconstructed blocks.

In general, the retrieval process consists of two steps: (1) identifying the block indices within the Kronecker product matrix that correspond to the required covariance block and (2) extracting the covariance of interests from the constructed block.

\paragraph{Identifying Block Indices} We first identify the Kronecker blocks that contain the covariance of interest. This is achieved by calculating the block indexed for $\bC$, which is later used to construct Kronecker blocks. Specifically, the start and end positions of the covariance block corresponding to rows $k$ and $l$ can be computed as:
    \[
    \text{row\_start} = \left\lfloor \frac{k \cdot D_{\text{in}}}{D_{\text{out}}} \right\rfloor,
    \]
    \[
    \text{row\_end} = \left\lceil \frac{(k + 1) \cdot D_{\text{in}}}{D_{\text{out}}} \right\rceil,
    \]
    \[
    \text{col\_start} = \left\lfloor \frac{l \cdot D_{\text{in}}}{D_{\text{out}}} \right\rfloor,
    \]
    \[
    \text{col\_end} = \left\lceil \frac{(l + 1) \cdot D_{\text{in}}}{D_{\text{out}}} \right\rceil.
    \]
Then, we can construct the Kronecker blocks that contain the covariance of interest by $\bC[\text{row\_start}:\text{row\_end}, \text{col\_start}:\text{col\_end}]\bD$.

\paragraph{Extract the Covariance}
Once we have $\bC[\text{row\_start}:\text{row\_end}, \text{col\_start}:\text{col\_end}]$, and as we know the covariance we need to retrieve has shape $\Din \times \Din$, we only need to compute the start row and column index, which can be computed as 
    \[
    \text{select\_row\_start} = (k \cdot D_{\text{in}}) \bmod D_{\text{out}},
    \]
    \[
    \text{select\_col\_start} = (l \cdot D_{\text{in}}) \bmod D_{\text{out}}.
    \]

\begin{figure}
  \centering
  \begin{tikzpicture}
    \scriptsize
    \matrix[matrix of math nodes, left delimiter=(,right delimiter=), nodes={font=\small, color=gray}] at (0,0) (M) {
      W_{11}, W_{21} & W_{11}, W_{23} &  W_{11}, W_{33} & W_{11}, W_{31} & W_{11}, W_{32} & W_{11}, W_{33} \\
      W_{12}, W_{11} & W_{12}, W_{12} & W_{12}, W_{13} & W_{12}, W_{31} & W_{12}, W_{32} & W_{12}, W_{33} \\
      W_{13}, W_{11} & W_{13}, W_{13} & W_{13}, W_{13} & W_{13}, W_{31} & W_{13}, W_{32} & W_{13}, W_{33} \\
      W_{21}, W_{11} & W_{21}, W_{12} & W_{21}, W_{13} & W_{21}, W_{31} & W_{21}, W_{32} & W_{21}, W_{33} \\
      W_{22}, W_{11} & W_{22}, W_{12} & W_{22}, W_{13} & W_{22}, W_{31} & W_{22}, W_{32} & W_{22}, W_{33} \\
      W_{23}, W_{11} & W_{23}, W_{12} & W_{23}, W_{13} & W_{23}, W_{31} & W_{23}, W_{32} & W_{23}, W_{33} \\ 
    };

    \begin{scope}[on background layer]
      \node[fill=highlight, rounded corners=2pt, fit=(M-1-4)(M-3-6)] (subM) {};
    \end{scope}

    \node[fit=(M-1-1)(M-2-2), draw=scPurple!50, inner sep=0pt, fill=white, fill opacity=0.5] {\textcolor{scPurple!37.5}{ \large \RNum{1} }};
    \node[fit=(M-1-3)(M-2-4), draw=scPurple, inner sep=0pt] {\textcolor{scPurple!75}{\large \RNum{2} }};
    \node[fit=(M-1-5)(M-2-6), draw=scPurple, inner sep=0pt] {\textcolor{scPurple!75}{\large \RNum{3} }}; 
    \node[fit=(M-3-1)(M-4-2), draw=scPurple!50, inner sep=0pt, fill=white, fill opacity=0.5] {\textcolor{scPurple!37.5}{\large \RNum{4} }};
    \node[fit=(M-3-3)(M-4-4), draw=scPurple, inner sep=0pt] {\textcolor{scPurple!75}{\large \RNum{5} }}; 
    \node[fit=(M-3-5)(M-4-6), draw=scPurple, inner sep=0pt] {\textcolor{scPurple!75}{\large \RNum{6} }};
    \node[fit=(M-5-1)(M-6-2), draw=scPurple!50, inner sep=0pt, fill=white, fill opacity=0.5] {\textcolor{scPurple!37.5}{\large \RNum{7} }};
    \node[fit=(M-5-3)(M-6-4), draw=scPurple!50, inner sep=0pt, fill=white, fill opacity=0.5] {\textcolor{scPurple!37.5}{\large \RNum{8} }};
    \node[fit=(M-5-5)(M-6-6), draw=scPurple!50, inner sep=0pt, fill=white, fill opacity=0.5] {\textcolor{scPurple!37.5}{\large \RNum{9} }};
    
    \normalsize
    \node[left of=M-3-1, anchor=base, xshift=-1.2cm, yshift=-0.25cm] (labelM) {$\operatorname{\mathbb{C}ov}\sbra{\bW}=$};

  \end{tikzpicture}
  \caption{To retrieve the highlighted submatrix $\cov{\bW[1,:]}{\bW[2, :]}$ of the covariance for $\bW \in \mathbb{R}^{2 \times 3}$, we identify the Kronecker blocks that contain the covariance of interest (\RNum{2}, \RNum{3}, \RNum{5}, and \RNum{6}), explicate those blocks in memory, and then retrieve the relevant submatrix.}
  \label{fig: kron-retrieve-app}
\end{figure}

\subsection{Derivation for Activation Layers}
\label{app:derivation-activation}

For $\ba = g(\bh)$ where $\bh \sim \mcN(\bh; \expect{\bh},\bSigma_h)$ and $g(\cdot)$ is the activation function, we use local linearisation to approximate the distribution of $\ba$. 
Specifically, we do a first-order Taylor expansion on $g(\cdot)$ at $\expect{\bh}$:
\begin{align}
	\ba &= g(\bh) \\
			& \approx g(\expect{\bh}) + \bJ_g|_{\bh=\expect{\bh}}(\bh - \expect{\bh}).
\end{align}
Given that Gaussian distribution is closed under linear transformation, we have 
\[      \bh \sim &\mcN(\expect{\bh},\bSigma_h) \\
		\bh - \expect{\bh} \sim & \mcN(\mathbf{0}, \bSigma_h) \\
		 \bJ_g|_{\bh=\expect{\bh}} (\bh - \expect{\bh}) \sim & \mcN(\mathbf{0},  {\bJ_g|_{\bh=\expect{\bh}}}^{\top}\bSigma_h \bJ_g|_{\bh=\expect{\bh}} )	 \\
		g(\expect{\bh}) + \bJ_g|_{\bh=\expect{\bh}}(\bh - \expect{\bh}) \sim & \mcN(g(\expect{\bh}),  {\bJ_g|_{\bh=\expect{\bh}}}^{\top}\bSigma_h \bJ_g|_{\bh=\expect{\bh}}) \\
		 \ba \underset{\text{approx}}{\sim} & \mcN(\ba; g(\expect{\bh}), {\bJ_g|_{\bh=\expect{\bh}}}^{\top}\bSigma_h \bJ_g|_{\bh=\expect{\bh}}).
\]

\subsection{Transformer Block}\label{app:derivation-transformer}

There are four components in each transformer block~\citep{vaswani2017attention}: (1) multi-head attention; (2) MLP; (3) layer normalisation; and (4) residual connection. 
For MLP blocks, the propagation is the same as described above.
For layer normalisation and residual connection, as Gaussian distributions are closed under linear transformations, `pushing' distributions through them is straightforward.
We describe how to push distributions through attention layers below.
Note that for computational reasons, we always assume the input has diagonal covariance.

Given an input $\bH \in \mathbb{R}^{T \times D}$ where $T$ is the number of tokens in the input sequence and $D$ is the dimension of each token, denote the query, key and value matrices as $\bW_Q \in \mathbb{R}^{D \times D}$, $\bW_K \in \mathbb{R}^{D \times D}$, $\bW_V \in \mathbb{R}^{D \times D}$ respectively, the key, query and value in an attention blocks are
\[
  \bQ = \bH \bW_Q, \quad \bK = \bH \bW_K, \quad  \bV = \bH \bW_V,
\]
and the output of attention block is 
\[
 \text{Attention}(\bH) = \text{Softmax}(\frac{\bQ \bK^\top}{\sqrt{D}}) \bV.
\]

When the input $\bH$ is a distribution, $\bQ$, $\bK$ and $\bV$ will all be distributions as well.
As pushing a distribution over a softmax activation requires further approximation, we ignore the distribution over $\bQ$ and $\bK$ for computational reasons and compute their value by using the mean of the input:
\[
  \bQ = \expect{\bH} \expect{\bW_Q}, \quad \bK = \expect{\bH} \expect{\bW_K}.
\]

For $\bV$, for simplicity we describe our approximation for a single token $\bh$ whose value is $\bv=\bW_V \bh$ with $k$\textsuperscript{th} element being $v_k=\sum_{i=1}^D W_{V_{ki}}h_i$. 
Assuming $\bh$ is a Gaussian, the covariance between the $k$\textsuperscript{th} and the $l$\textsuperscript{th} value is
\[
\cov{v_k}{v_l} &= \cov{\sum_{i=1}^D W_{V_{ki}}h_i}{\sum_{j=1}^D W_{V_{lj}}h_j} \\
			   &= \sum_{i=1}^D\sum_{j=1}^D\cov{ W_{V_{ki}}h_i}{ W_{V_{lj}}h_j}.
\]

In the case of deterministic $\bW_V$, we have for $k\neq l$,
\[
\cov{v_k}{v_l} &\approx \sum_{i=1}^{\Din}\sum_{j=1}^{\Din} \expect{h_i}\expect{h_j}\textcolor{gray!50}{\cov{W_V[k,i]}{W_V[l,j]}}\nonumber\\
&\hspace{3em} +\sum_{i=1}^{\Din}\sum_{j=1}^{\Din}  \expect{W_V[k,i]}\expect{W_V[l,j]} \cov{h_i}{h_j}\nonumber\\ 
				&\hspace{3em} +\sum_{i=1}^{\Din}\sum_{j=1}^{\Din}  \cov{h_i}{h_j}\textcolor{gray!50}{\cov{W_V[k,i]}{W_V[l,j]}} \\
              	&= \sum_{i=1}^{\Din}\sum_{j=1}^{\Din}  W_V[k,i]W_V[l,j]\cov{h_i}{h_j} \tag{$\bW_V$ deterministic} \\
              	&\approx \sum_{i=1}^{\Din} W_V[k,i]W_V[l,i] \var{h_i}. \tag{we assume $\bh$ has diagonal covariance}
\]

\[
\var{v_k} &\approx \sum_{i=1}^{\Din}\sum_{j=1}^{\Din} \expect{h_i}\expect{h_j}\textcolor{gray!50}{\cov{W_V[k,i]}{W_V[k,j]}} \nonumber\\
              	&\hspace{3em} +\sum_{i=1}^{\Din}\sum_{j=1}^{\Din} \expect{W_V[k,i]} \expect{W_V[k,j]} \cov{h_i}{h_j} \nonumber\\
              	&\hspace{3em} +\sum_{i=1}^{\Din}\sum_{j=1}^{\Din} \cov{h_i}{h_j}\textcolor{gray!50}{\cov{W_V[k,i]}{W_V[k,j]}}\\
              	&=\sum_{1\leq i \leq D} W_V[k,i]^2\var{h_i} \tag{we assume $\bh$ has diagonal covariance} \\
\]

When $\bW_V$ is an isotropic Gaussian, we have for $k\neq l$,
\[
\cov{v_k}{v_l} &\approx \sum_{i=1}^{\Din}\sum_{j=1}^{\Din} \expect{h_i}\expect{h_j}\textcolor{gray!50}{\cov{W_V[k,i]}{W_V[l,j]}}\nonumber\\
&\hspace{3em} +\sum_{i=1}^{\Din}\sum_{j=1}^{\Din}  \expect{W_V[k,i]}\expect{W_V[l,j]} \cov{h_i}{h_j}\nonumber\\ 
				&\hspace{3em} +\sum_{i=1}^{\Din}\sum_{j=1}^{\Din}  \cov{h_i}{h_j}\textcolor{gray!50}{\cov{W_V[k,i]}{W_V[l,j]}} \label{eq:covariance_transformer}\\
              	&= \sum_{i=1}^{\Din}\sum_{j=1}^{\Din} \expect{W_V[k,i]} \expect{W_V[l,j]} \cov{h_i}{h_j} \tag{$\bW_V$ is isotropic Gaussian} \\
              	&\approx \sum_{1\leq i \leq D} \expect{W_V[k,i]} \expect{W_V[l,i]} \var{h_i}. \tag{ignore correlation between $\bh$ for computational reason} 
\]

\[
\var{v_k} &\approx \sum_{i=1}^{\Din}\sum_{j=1}^{\Din} \expect{h_i}\expect{h_j}\cov{W_V[k,i]}{W_V[k,j]} \nonumber\\
              	&\hspace{3em} +\sum_{i=1}^{\Din}\sum_{j=1}^{\Din} \expect{W_V[k,i]} \expect{W_V[k,j]} \cov{h_i}{h_j} \nonumber \\
              	&\hspace{3em} +\sum_{i=1}^{\Din}\sum_{j=1}^{\Din} \cov{h_i}{h_j}\cov{W_V[k,i]}{W_V[k,j]} \\
              	&= \sum_{i=1}^{\Din}  \expect{h_i}^2\var{W_V[k,i]} \nonumber \\
              	&\hspace{3em} +\sum_{i=1}^{\Din}\sum_{j=1}^{\Din} \expect{W_V[k,i]} \expect{W_V[k,j]} \cov{h_i}{h_j} \nonumber \\
              	&\hspace{3em} +\sum_{i=1}^{\Din} \var{h_i} \var{W_V[k,i]}  \nonumber \tag{$\bW_V$ is isotropic Gaussian} \\
                &\approx \sum_{i=1}^{\Din}  \expect{h_i}^2\var{W_V[k,i]} \nonumber \\
              	&\hspace{3em} +\sum_{i=1}^{\Din} \expect{W_V[k,i]}^2 \var{h_i} \nonumber \\
              	&\hspace{3em} +\sum_{i=1}^{\Din} \var{h_i} \var{W_V[k,i]}  \nonumber \tag{we assume $\bh$ has diagonal covariance}
\]

Once we have the distribution over $\bV$, the distribution over $\text{Attention}(\bH)$ becomes a distribution of linear combination of Gaussian, which is tractable.

Then for multi-head attention, we assume each attention head's output is independent, which allows us to compute the distribution over the final output in tractable form. 
As we assume the input is isotropic, we only need to compute the variance for each dimension.

\subsection{Convolutional Neural Network}\label{app:derivation-cnn}
\newcommand{\cin}{c_{\text{in}}}
\newcommand{\Cin}{C_{\text{in}}}
\newcommand{\cout}{c_{\text{out}}}

The derivation for convolutional layers is similar to fully connected layers as convolution layers can be considered as a shared weight fully connected layer. We first give the derivation for convolutional layers, then discuss pooling layers in convolutional neural networks.

Denote the pixel value at $(i,j)$ of $\cin$\textsuperscript{th} channel as $a_{\cin}[i,j]$, the $\cin$\textsuperscript{th} channel of convolutional kernel corresponding to $\cout$\textsuperscript{th} output channel as $W_{\cout, \cin}[i,j]$ and the pixel value at $(k,l)$ of the $\cout$\textsuperscript{th} output channel as $h_{\cout}[k,l]$. Then, suppose there are $\Cin$ channels in total and the kernel size is $K_h \times K_w$, we can write the convolutional layer as
\[
h_{\cout}[k,l] = \sum_{\cin=1}^{\Cin} \sum_{i=1}^{K_h}\sum_{j=1}^{K_w} a_{\cin}[k+i-1, l+j-1]W_{\cout, \cin}[i,j].
\]

\paragraph{Derivation of mean} Following our assumption that $a_{\cin}[k+i-1, l+j-1]$ is uncorrelated with $W_{\cout, \cin}[i,j]$, we have 
\[
\expect{h_{\cout}[k,l]} = \sum_{\cin=1}^{\Cin} \sum_{i=1}^{K_h}\sum_{j=1}^{K_w} \expect{a_{\cin}[k+i-1, l+j-1]}\expect{W_{\cout, \cin}[i,j]}
\]

\paragraph{Derivation of covariance} The covariance between pixels of the $\cout$\textsuperscript{th} output channel are given as:
\begin{align}
&\cov{h_{\cout}[k_1,l_1]}{h_{\cout}[k_2,l_2]} \\ 
&=\mathbb{C}\text{ov} \left\lbrack  \sum_{c_{\text{in},1}=1}^{\Cin}\sum_{i_1=1}^{K_h}\sum_{j_1=1}^{K_w} a_{c_{\text{in},1}}[k_1+i_1-1, l_1+j_1-1]W_{\cout, \cin}[i_1,j_1] \right., \\
&\left. \hspace{4em} \sum_{c_{\text{in},2}=1}^{\Cin} \sum_{i_2=1}^{K_h}\sum_{j_2=1}^{K_w} a_{c_{\text{in},2}}[k_2+i_2-1, l_2+j_2-1]W_{\cout, \cin}[i_2,j_2]  \right\rbrack \\
&=\sum_{c_{\text{in},1}=1}^{\Cin}\sum_{i_1=1}^{K_h}\sum_{j_1=1}^{K_w} \sum_{c_{\text{in},2}=1}^{\Cin} \sum_{i_2=1}^{K_h}\sum_{j_2=1}^{K_w} \mathbb{C}\text{ov} \left\lbrack a_{c_{\text{in},1}}[k_1+i_1-1, l_1+j_1-1]W_{\cout, \cin}[i_1,j_1] \right. \\
&\left. \hspace{18em}a_{c_{\text{in},2}}[k_2+i_2-1, l_2+j_2-1]W_{\cout, \cin}[i_2,j_2]\right \rbrack.
\end{align}

Using earlier results from \cref{eq:cov-derivation} and the shorthand $W = W{\cout, \cin}$, we have:
\[
&\cov{a_{\text{in},1}[k_1\!+\!i_1\!-\!1, l_1\!+\!j_1\!-\!1]W[i_1,j_1]}{a_{\text{in},2}[k_2\!+\!i_2\!-\!1, l_2\!+\!j_2\!-\!1]W[i_2,j_2]} \\
&\approx \expect{a_{\text{in},1}[k_1\!+\!i_1\!-\!1, l_1\!+\!j_1\!-\!1]a_{\text{in},2}[k_2\!+\!i_2\!-\!1, l_2\!+\!j_2\!-\!1]}\expect{W[i_1,j_1]W[i_2,j_2]}  \\
& - \expect{a_{\text{in},1}[k_1\!+\!i_1\!-\!1, l_1\!+\!j_1\!-\!1]}\expect{a_{\text{in},2}[k_2\!+\!i_2\!-\!1, l_2\!+\!j_2\!-\!1]}\expect{W[i_1,j_1]}\expect{W[i_2,j_2]} \\
&= \expect{a_{\text{in},1}[k_1\!+\!i_1\!-\!1, l_1\!+\!j_1\!-\!1]}\expect{a_{\text{in},2}[k_2\!+\!i_2\!-\!1, l_2\!+\!j_2\!-\!1]}\cov{W[i_1,j_1]}{W[i_2,j_2]} \\
& + \expect{W[i_1,j_1]}\expect{W[i_2,j_2]} \cov{a_{\text{in},1}[k_1\!+\!i_1\!-\!1, l_1\!+\!j_1\!-\!1]}{a_{\text{in},2}[k_2\!+\!i_2\!-\!1, l_2\!+\!j_2\!-\!1]}\\
& + \cov{a_{\text{in},1}[k_1\!+\!i_1\!-\!1, l_1\!+\!j_1\!-\!1]}{a_{\text{in},2}[k_2\!+\!i_2\!-\!1, l_2\!+\!j_2-1]}\cov{W[i_1,j_1]}{W[i_2,j_2]} .
\]

\section{Additional Experiments} \label{app:experiment}
In the Appendix, we provide additional details on (i) the regression experiments \cref{app:experiment-regression}, (ii) the classification experiments \cref{app:experiment-classification}, and (iii) the image sensitivity experiment \cref{app:experiment-image-sensitivity}.
In addition, we also present additional experiments on (i) measuring the performance by varying the number of MC samples for the sampling baseline \cref{app:experiment-mcsamples}, (ii) estimating the degree of local linearity in our method \cref{app:experiment-linearity}, and (iii) comparing the runtime of our method against the baselines \cref{app:runtime} to further demonstrate the benefits of our streamlined prediction with local linearisation and local Gaussian approximation.

\subsection{Regression} \label{app:experiment-regression}
\cref{table:uci-dataset} gives the UCI regression data set information and the neural network structure we used. 
For all neural networks, we use the ReLU activation function.
In \cref{table:regression_rmse}, we report the Root Mean Square Error (RMSE). Our method results in matching or better performance compared with sampling and GLM, indicating the effectiveness of our method.
Note that as the mean of the posterior prediction of our method is the same as the prediction made by setting the weights of the neural network to be the mean of the posterior, we result in the same prediction as GLM of LA, and hence the same performance.

\begin{table}[h]
	\centering
	\renewcommand*{\arraystretch}{1.1}
	\setlength{\tabcolsep}{5.5pt}
	\scriptsize
	\caption{UCI regression experiment setup.}
\begin{tabular}{llcc}
\toprule
{Data Set Name} & {Shorthand} & $(n,d)$ &  Network Structure    \\[0.2em]
\hline
{\sc Servo}  & {\sc Servo}    &       (167, 4)     & $d$-50-1    \\
{\sc Liver Disorders} &   {\sc LD} &    (345, 5)   & $d$-50-1   \\
{\sc Auto MPG} &  {\sc AM}    &  (398, 7)    & $d$-50-1    \\
{\sc Real Estate Valuation} &   {\sc REV}  & (414,6) & $d$-50-1 \\ 
{\sc Forest Fires} & {\sc FF}  &   (517, 12)  & $d$-50-1    \\
{\sc Infrared Thermography Temperature} & {\sc ITT} & (1020, 33)   & $d$-100-1   \\
{\sc Concrete Compressive Strength} & {\sc CCS} &      (1030, 8)   & $d$-100-1     \\
{\sc Airfoil Self-Noise} & {\sc ASN} &   (1503, 5)    & $d$-100-1   \\
{\sc Communities and Crime} & {\sc CAC} & (1994, 127)   & $d$-100-1   \\
{\sc Parkinsons Telemonitoring} & {\sc PT} &  (5875, 19)   & $d$-50-50-1   \\
{\sc Combined Cycle Power Plant} & {\sc CCPP} &   (9568, 4)  & $d$-50-50-1  \\
\bottomrule\end{tabular}
\label{table:uci-dataset}
\end{table}

\begin{table*}[!ht]
	\centering
	\renewcommand*{\arraystretch}{1.1}
	\setlength{\tabcolsep}{5.5pt}
	\scriptsize
	\caption{Root Mean Square Error $\downarrow$ on UCI regression data sets. Our method results in better or matching performance compared with sampling and GLM, indicating its effectiveness.}
	\begin{tabular}{lc|cH|ccH}
\toprule
& {} & \multicolumn{2}{ c|}{\textit{MFVI (Diag. Cov.)}} & \multicolumn{3}{c}{\textit{Laplace Approximation (Full Cov.)}} \\[0.2em]
{} & $(n,d)$ &  Sampling     &     Ours       &    Sampling      &       GLM        &      Ours        \\[0.2em]
\hline
    {\sc Servo}      &       (167, 4)       & \val{}{0.749}{0.147} & \val{\bf}{0.740}{0.143} & \val{}{1.632}{0.233} & \val{\bf}{0.658}{0.141} & \val{\bf}{0.658}{0.141} \\
    {\sc LD} &       (345, 5)       & \val{\bf}{0.884}{0.273} & \val{\bf}{0.881}{0.272} & \val{\bf}{0.989}{0.441} & \val{\bf}{0.977}{0.418} & \val{\bf}{0.977}{0.418} \\
   {\sc AM}    &       (398, 7)       & \val{\bf}{0.415}{0.115} & \val{\bf}{0.417}{0.113} & \val{}{0.505}{0.105} & \val{\bf}{0.371}{0.103} & \val{\bf}{0.371}{0.103} \\
{\sc REV} &       (414, 6)       & \val{\bf}{0.563}{0.096} & \val{\bf}{0.562}{0.095} & \val{}{0.789}{0.130} & \val{\bf}{0.532}{0.104} & \val{\bf}{0.532}{0.104} \\
 {\sc FF}  &      (517, 12)       & \val{\bf}{0.874}{1.123} & \val{\bf}{0.874}{1.124} & \val{\bf}{0.910}{0.824} & \val{\bf}{0.852}{0.792} & \val{\bf}{0.852}{0.792} \\
 {\sc ITT} &      (1020, 33)      & \val{\bf}{0.481}{0.057} & \val{\bf}{0.497}{0.066} & \val{}{0.560}{0.075} & \val{\bf}{0.507}{0.072} & \val{\bf}{0.507}{0.072} \\
{\sc CCS} &      (1030, 8)       & \val{\bf}{0.472}{0.102} & \val{\bf}{0.476}{0.106} & \val{}{0.494}{0.102} & \val{\bf}{0.301}{0.057} & \val{\bf}{0.301}{0.057} \\
{\sc ASN} &      (1503, 5)       & \val{}{0.568}{0.062} & \val{\bf}{0.560}{0.062} & \val{}{0.550}{0.069} & \val{\bf}{0.352}{0.055} & \val{\bf}{0.352}{0.055} \\
{\sc CAC} &     (1994, 127)      & \val{\bf}{0.571}{0.105} & \val{}{0.585}{0.092} & \val{}{1.481}{0.167} & \val{\bf}{0.703}{0.101} & \val{\bf}{0.703}{0.101} \\
{\sc PT} &      (5875, 19)      & \val{}{0.601}{0.067} & \val{\bf}{0.590}{0.068} & \val{}{0.479}{0.081} & \val{\bf}{0.410}{0.076} & \val{\bf}{0.410}{0.076} \\
{\sc CCPP} &      (9568, 4)       & \val{\bf}{0.241}{0.038} & \val{\bf}{0.241}{0.038} & \val{}{0.358}{0.041} & \val{\bf}{0.224}{0.037} & \val{\bf}{0.224}{0.037} \\
\midrule
\multicolumn{2}{ c|}{Bold Count } & $8/11$ & $10/11$ & $2/11$ & $11/11$ & $11/11$ \\
\bottomrule
\end{tabular}
	\label{table:regression_rmse}
\end{table*}

\subsection{Classification} \label{app:experiment-classification}
\cref{table:classification-dataset} gives the classification data sets information and the neural network structure we used for the MLP experiment.
We use ReLU activation for MLP.

\begin{table}[!ht]
	\centering
	\renewcommand*{\arraystretch}{1.1}
	\setlength{\tabcolsep}{5.5pt}
	\scriptsize
	\caption{Classification experiment setup.}
  \vspace{-3mm}
  \begin{tabular}{lcc}
    \toprule
    {Data Set Name} & $(n,d)$ &  Network Structure    \\[0.2em]
    \hline
    {\sc MNIST}  &   (50000, 784)     & $d$-128-64-10    \\
    {\sc FMNIST}  &   (50000, 784)     & $d$-128-64-10    \\
    {\sc OrganCMNIST}  &   (12975, 784)      & $d$-128-64-11    \\
    {\sc OrganSMNIST}  &   (13932, 784)      & $d$-128-64-11    \\
    \bottomrule
  \end{tabular}
  \label{table:classification-dataset}
\end{table}

\begin{figure}[h!]
  \centering\scriptsize
  \setlength{\figurewidth}{.44\textwidth}
  \setlength{\figureheight}{.2\textwidth}
  \pgfplotsset{
    height=\figureheight,
    width=\figurewidth,
    tick align=inside, 
    x tick label style={font=\tiny}, 
    y tick label style={rotate=90, font=\tiny}, 
    y label style = {yshift=0em, font=\scriptsize}, 
    x label style = {yshift=0.5em}, scale only axis, 
    xmin=-1, xmax=181,
    legend cell align={left},
  }
  \begin{subfigure}[t]{0.49\textwidth}
    \centering
    \pgfplotsset{xtick = {0,60,120,180}, 
      xticklabels = {0,60,120,180}, 
      xlabel={Rotation Degree}, 
      ylabel = {$\leftarrow$ NLPD},
      ymin=-0.601873469258552, ymax=14.3034773407351,
      legend style={at={(0.03,0.97)},anchor=north west,},
    }
    % This file was created with tikzplotlib v0.10.1.post12.
\begin{tikzpicture}

\definecolor{darkgray176}{RGB}{176,176,176}
\definecolor{lightgray204}{RGB}{204,204,204}
\definecolor{limegreen}{RGB}{50,205,50}
\definecolor{royalblue}{RGB}{65,105,225}

\begin{axis}[
height=\figureheight,
legend cell align={left},
legend style={
  fill opacity=0.8,
  draw opacity=1,
  text opacity=1,
  at={(0.03,0.97)},
  anchor=north west,
  draw=lightgray204
},
tick align=outside,
tick pos=left,
width=\figurewidth,
x grid style={darkgray176},
xlabel={Rotation Degree},
xmin=-9, xmax=189,
xtick style={color=black},
y grid style={darkgray176},
ymin=-0.568320439538692, ymax=13.2814761944032,
ytick style={color=black}
]
\addplot [thick, limegreen, dash pattern=on 1pt off 3pt on 3pt off 3pt]
table {%
0 0.0649787512318457
10 0.128372598665505
20 0.348216755643499
30 0.934335276687886
40 1.89229236721201
50 3.08732579917433
60 4.34791820090307
70 5.48073479927781
80 6.56547780383884
90 7.24814609111413
100 7.61854698418934
110 7.4371598594306
120 7.04363282561448
130 6.45907146948089
140 6.08526700494756
150 5.90720431608412
160 6.04525579930442
170 6.22572094469143
180 6.15974562404057
};
\addlegendentry{LA GLM}
\addplot [thick, limegreen, dashed]
table {%
0 0.0612157710950313
10 0.123068536698732
20 0.344664643283615
30 0.967851418114792
40 2.11458186026987
50 3.67695545186147
60 5.26398034421
70 6.75725024008489
80 8.23316486673074
90 8.99218659234008
100 9.40019241992389
110 9.12850242606859
120 8.64961384553815
130 7.94090623038001
140 7.42721455225542
150 7.10459300882833
160 7.19507949126131
170 7.37680426789163
180 7.25953404926747
};
\addlegendentry{LA Sampling}
\addplot [very thick, limegreen]
table {%
0 0.0631684304028892
10 0.122173513017443
20 0.319752188936123
30 0.818771674582339
40 1.60245487814889
50 2.56285953347988
60 3.5576351045283
70 4.45745335192779
80 5.32736495258371
90 5.93378612011098
100 6.25605611999713
110 6.11119094126884
120 5.78489048738656
130 5.31001981805441
140 5.02762742566004
150 4.96326233341655
160 5.18117778958999
170 5.41654882489725
180 5.39667900693582
};
\addlegendentry{LA Ours}
\addplot [thick, royalblue, dashed]
table {%
0 0.0623742010259243
10 0.122177717474539
20 0.31884338165884
30 0.916122533815191
40 1.9251483663225
50 3.22175101619184
60 4.52715092059748
70 5.74364192059521
80 6.78722218019513
90 7.64090689953234
100 8.23759669512953
110 8.2461460462984
120 8.01373963517431
130 7.5905386575175
140 7.37276862209182
150 7.26545436486921
160 7.19889148472131
170 7.22684158312975
180 7.01344717537454
};
\addlegendentry{IVON Sampling}
\addplot [very thick, royalblue]
table {%
0 0.0623276403006729
10 0.119098408379803
20 0.303884423305164
30 0.841953423348979
40 1.70479115069785
50 2.74022372043569
60 3.74094657641186
70 4.65545818293818
80 5.41191315913619
90 6.02667732554248
100 6.56605624627223
110 6.63499738073352
120 6.56118533015508
130 6.28595006753144
140 6.14814996604496
150 6.10269778828599
160 6.09028323846818
170 6.13020757352693
180 5.99380448571225
};
\addlegendentry{IVON Ours}
\addplot [thick, black]
table {%
0 0.125647824254404
10 0.262872742211815
20 0.770653404007922
30 2.22865489122189
40 4.45719312609655
50 6.88577683894908
60 9.04838986412739
70 10.7627023309828
80 11.9955261116683
90 12.4278298545465
100 12.6519399837695
110 12.4280764147347
120 11.8181589784704
130 11.0610513358148
140 10.5597767881389
150 9.92261294873313
160 9.58840413900573
170 9.38770828169662
180 9.07973910406923
};
\addlegendentry{MAP}
\end{axis}

\end{tikzpicture}
    \vspace{-1.1em}
  \end{subfigure}
  \hfill
  \begin{subfigure}[t]{0.49\textwidth}
    \centering
    \pgfplotsset{xtick = {0, 60, 120, 180},
      xticklabels = {0,60,120,180}, 
      xlabel={Rotation Degree}, 
      ylabel = {ACC $\rightarrow$},
      ymin=0.0710638933121019, ymax=1.02119327229299,
      legend style={at={(0.6,0.97)},anchor=north west,},
    }
  % This file was created with tikzplotlib v0.10.1.post12.
\begin{tikzpicture}

\definecolor{darkgray176}{RGB}{176,176,176}
\definecolor{lightgray204}{RGB}{204,204,204}
\definecolor{limegreen}{RGB}{50,205,50}
\definecolor{royalblue}{RGB}{65,105,225}

\begin{axis}[
height=\figureheight,
legend cell align={left},
legend style={
  fill opacity=0.8,
  draw opacity=1,
  text opacity=1,
  at={(0.03,0.03)},
  anchor=south west,
  draw=lightgray204
},
tick align=outside,
tick pos=left,
width=\figurewidth,
x grid style={darkgray176},
xlabel={Rotation Degree},
xmin=-9, xmax=189,
xtick style={color=black},
y grid style={darkgray176},
ymin=0.072065, ymax=1.025435,
ytick style={color=black}
]
\addplot [thick, limegreen, dash pattern=on 1pt off 3pt on 3pt off 3pt]
table {%
0 0.9803
10 0.9607
20 0.8981
30 0.7552
40 0.5746
50 0.4015
60 0.2719
70 0.1849
80 0.1345
90 0.1154
100 0.1165
110 0.1171
120 0.151
130 0.1875
140 0.2137
150 0.2464
160 0.2876
170 0.3176
180 0.3371
};
\addlegendentry{LA GLM}
\addplot [thick, limegreen, dashed]
table {%
0 0.9821
10 0.9618
20 0.9022
30 0.7552
40 0.5748
50 0.4009
60 0.2719
70 0.1876
80 0.1379
90 0.1193
100 0.119
110 0.1201
120 0.1543
130 0.1903
140 0.219
150 0.2523
160 0.2928
170 0.3216
180 0.3397
};
\addlegendentry{LA Sampling}
\addplot [very thick, limegreen]
table {%
0 0.9803
10 0.9603
20 0.8978
30 0.755
40 0.5739
50 0.4012
60 0.2716
70 0.1851
80 0.1345
90 0.1157
100 0.1168
110 0.1169
120 0.1506
130 0.187
140 0.2135
150 0.2466
160 0.2875
170 0.3187
180 0.3379
};
\addlegendentry{LA Ours}
\addplot [thick, royalblue, dashed]
table {%
0 0.981
10 0.9606
20 0.9047
30 0.7523
40 0.5654
50 0.3959
60 0.286
70 0.1992
80 0.1556
90 0.1318
100 0.1205
110 0.1201
120 0.151
130 0.1839
140 0.2126
150 0.242
160 0.2784
170 0.3009
180 0.3181
};
\addlegendentry{IVON Sampling}
\addplot [very thick, royalblue]
table {%
0 0.9803
10 0.9609
20 0.903
30 0.7539
40 0.5663
50 0.4018
60 0.2876
70 0.1982
80 0.1536
90 0.1323
100 0.1203
110 0.1192
120 0.1566
130 0.1862
140 0.2141
150 0.244
160 0.2807
170 0.3029
180 0.3221
};
\addlegendentry{IVON Ours}
\addplot [thick, black]
table {%
0 0.9818
10 0.9623
20 0.9002
30 0.7555
40 0.5759
50 0.4058
60 0.2751
70 0.189
80 0.138
90 0.1187
100 0.1212
110 0.1217
120 0.1547
130 0.191
140 0.2183
150 0.2487
160 0.2894
170 0.3199
180 0.3397
};
\addlegendentry{MAP}
\end{axis}

\end{tikzpicture}
  \vspace{-1.1em}
  \end{subfigure}
  \hfill
  \caption{NLPD and ACC for MNIST-trained MLP on rotated versions of the MNIST test set. The rotation degree interval is $10^{\circ}$ from $0-180^{\circ}$. Our method achieves similar NLPD for both LA and MFVI posterior approximations. 
  }
  \label{fig:rmnist-ood}
\end{figure}
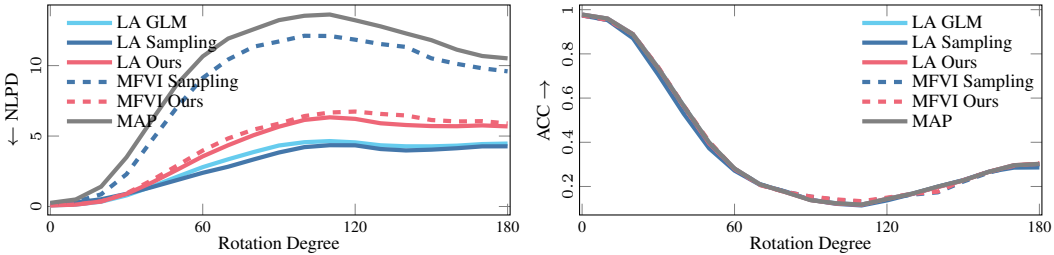

\begin{figure}[h!]
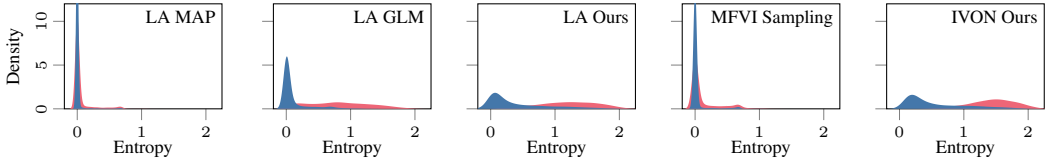

  \centering\scriptsize
  \setlength{\figurewidth}{.15\textwidth}
  \setlength{\figureheight}{.1\textwidth}
  \pgfplotsset{
    x tick label style={font=\tiny}, 
    y tick label style={rotate=90, font=\tiny}, 
    y label style = {font=\scriptsize}, 
    x label style = {yshift=0.5em}, 
    xlabel = {Entropy}, ylabel = {Density},
    scale only axis,
    tick align=outside,
    tick pos=left,
    xmin=-0.2, xmax=2.25,
    ymin=0, ymax=7.5,%
  }
  \input{fig/ood_mlp_la_map.tex}
  \pgfplotsset{ylabel=\empty, yticklabel=\empty,}
  \input{fig/ood_mlp_la_glm.tex}
  \input{fig/ood_mlp_la_dbnn.tex}
  \input{fig/ood_mlp_ivon_sample.tex}
  \input{fig/ood_mlp_ivon_dbnn.tex}
  \caption{Kernel density plots over the predictive entropy from an MLP trained on FMNIST (blue, in-distribution) and data from MNIST (red, out-of-distribution). Our method results in a clear separation between the in- and out-of-distribution data. 
  }
  \label{fig:mlp-ood}
\end{figure}

\paragraph{OOD Experiments with MLP} 
To test our method on out-of-distribution (OOD) data, we first evaluate the MNIST-trained MLP on rotated versions of the test set as shown in \cref{fig:rmnist-ood}. The rotation degree interval is $10^{\circ}$ from $0-180^{\circ}$. 
We observe that with increasing rotation degree, our method achieves a lower NLPD compared to LA MAP and MFVI Sampling while being close compared with LA Sampling and GLM. Also, our method achieves similar NLPD for both LA and MFVI posterior approximations across the rotation degrees. All methods perform on par regarding their ACC. 
In \cref{fig:mlp-ood}, we show kernel density plots over the predictive entropy of an FMNIST-trained MLP evaluated on MNIST. Our method can distinguish between in-distribution and OOD data better than the LA MAP and MFVI Sampling. Although our method under fits the in-distribution data, the separation between them is clear for the OOD data. 

\paragraph{Our method applied to MLP in ViT} In \cref{tab:vit_mlp} we report the results for fine-tuning the MLPs after the attention layers in the last two transformer blocks in ViT and later treating them Bayesian. 
We observe that our method achieves better or on par NLPD and ECE compared to the baselines for both LA and MFVI across all data sets while maintaining similar ACC as the baselines. 

\begin{table*}[!ht]
	\centering
	\caption{Performance metrics using ViT with posterior approximation on MLPs after the attention layers with the standard error for ACC and NLPD. Our method achieves better NLPD and ECE in general and achieves similar ACC compared to the baselines. 
  }
  \vspace*{-6pt}
	\setlength{\tabcolsep}{8.5pt}
	\scriptsize
	\begin{tabular}{l|l|ccccc}
\toprule
Metrics & Methods & \sc CIFAR-10 & \sc CIFAR-100  & \sc DTD  & \sc RESISC & \sc Imagenet-R \\
\hline
& LA Sampling & \val{}{0.964}{0.002} & \val{}{0.847}{0.004} & \val{}{0.671}{0.011} & \val{}{0.819}{0.005} & \val{}{0.642}{0.013} \\
 & LA GLM & \val{\bf}{0.975}{0.002} & \val{}{0.874}{0.003} & \val{\bf}{0.715}{0.010} & \val{}{0.896}{0.004} & \val{\bf}{0.714}{0.012} \\
 \rowcolor{highlight}\cellcolor{white} ACC $\uparrow$
 & LA Ours & \val{\bf}{0.976}{0.002} & \val{\bf}{0.888}{0.003} & \val{\bf}{0.718}{0.010} & \val{\bf}{0.911}{0.004} & \val{\bf}{0.727}{0.012}\\\cdashline{2-7} \rule{0pt}{2.25ex}
 & MFVI Sampling & \val{\bf}{0.979}{0.001} & \val{\bf}{0.891}{0.003} & \val{\bf}{0.739}{0.010} & \val{\bf}{0.927}{0.003} & \val{\bf}{0.762}{0.011} \\
 \rowcolor{highlight}\cellcolor{white}
 & MFVI Ours & \val{\bf}{0.979}{0.001} & \val{\bf}{0.891}{0.003} & \val{\bf}{0.729}{0.010} & \val{\bf}{0.926}{0.003} & \val{\bf}{0.751}{0.011} \\
\hline
 & LA Sampling & \val{}{0.194}{0.004} & \val{}{1.036}{0.011} & \val{}{2.085}{0.020} & \val{}{1.301}{0.011} & \val{}{2.226}{0.037} \\
 & LA GLM & \val{\bf}{0.092}{0.006} & \val{}{0.661}{0.012} & \val{}{1.319}{0.027} & \val{}{0.527}{0.011} & \val{}{1.455}{0.043} \\
  \rowcolor{highlight}\cellcolor{white} NLPD $\downarrow$
 & LA Ours & \val{\bf}{0.087}{0.005} & \val{\bf}{0.483}{0.013} & \val{\bf}{1.016}{0.033} & \val{\bf}{0.299}{0.012} & \val{\bf}{1.232}{0.045}\\ \cdashline{2-7} \rule{0pt}{2.25ex}
 & MFVI Sampling & \val{}{0.096}{0.008} & \val{}{0.524}{0.019} & \val{\bf}{0.947}{0.041} & \val{\bf}{0.282}{0.016} & \val{\bf}{1.137}{0.065} \\
  \rowcolor{highlight}\cellcolor{white} 
 & MFVI Ours & \val{\bf}{0.073}{0.005} & \val{\bf}{0.425}{0.011} & \val{\bf}{0.985}{0.034} & \val{\bf}{0.250}{0.011} & \val{\bf}{1.039}{0.048} \\
\hline
 & LA Sampling & $0.086$ & $0.330$ & $0.480$ & $0.432$ & $0.401$ \\
 & LA GLM & $0.007$ & $0.104$ & $0.246$ & $0.139$ & $0.164$ \\
  \rowcolor{highlight}\cellcolor{white} ECE $\downarrow$
 & LA Ours & $\bf0.005$ & $\bf0.036$ & $\bf0.031$ & $\bf0.021$ & $\bf0.070$\\  \cdashline{2-7} \rule{0pt}{2.25ex}
 & MFVI Sampling & $0.011$ & $0.055$ & $0.044$ & $0.030$ & $0.051$ \\
  \rowcolor{highlight}\cellcolor{white} 
 & MFVI Ours & $\bf0.004$ & $\bf0.023$ & $\bf0.039$ & $\bf0.012$ & $\bf0.036$ \\
\bottomrule
\end{tabular}
    
	\label{tab:vit_mlp}
\end{table*}

We present results for GPT-2 on tasks from GLUE \citep{wang2018glue} and SuperGLUE \citep{wang2019superglue} benchmarks.
These natural language understanding tasks could be turned into classification tasks with the prompt shown in \cref{table:gpt_prompt}.
We add a classification layer on top of the encoder and use the embedding of the last token in each input to do classification.

\begin{table}[h]
\centering
\caption{Prompt templates for fine-tuning GPT-2 on natural language understanding tasks.}
\begin{tabular}{c|c}
\toprule
\textbf{Task} & \textbf{Prompt} \\
\hline
MRPC & Answer whether sentence 2 is equivalent to sentence 1. \\
& Sentence 1: \{\texttt{sentence1}\}. Sentence 2: \{\texttt{sentence2}\}. Answer: \\
\hline
WiC & Select whether word \{\texttt{word}\} has the same meaning in these two sentences. \\
& Sentence 1: \{\texttt{sentence1}\}. Sentence 2: \{\texttt{sentence2}\}. Answer: \\
\hline
BoolQ & Answer the question with only True or False. \\
& Passage: \{\texttt{passage}\}. Question: \{\texttt{question}\}. Answer: \\
\bottomrule
\end{tabular}
\label{table:gpt_prompt}
\end{table}

\paragraph{Lasy Layer Laplace Approximation on ViT} In \cref{table:vit-last-layer} we report the results for fine-tuning only the last classification layer in ViT base and later treating it Bayesian. We observe that our method (LL-LA Ours) achieves better or on par NLPD and ECE compared to last layer Laplace approximation (LL-LA GLM/Sampling) across all data sets while maintaining similar ACC. Compared to the case where more layers are treated Bayesian (LA Ours) (results are taken from \cref{tab:vit_attention}), last layer approximations in general have lower accuracies and higher NLPD and ECE, \emph{indicating the benefits gained by treating more layers Bayesian.}
In \cref{table:wall-clock-last-layer} we report the wall-clock run times for last layer Laplace approximation on CIFAR-10 in milliseconds (see \cref{app:runtime} for the run time setting) Our method has matching speed with MAP and slight speed improvements over GLM.

\begin{table}
\caption{Performance metrics using ViT with posterior approximation on last layer with the standard error for ACC and NLPD. In the last layer Laplace approximation (LL-LA), our method achieves better NLPD and ECE in general and achieves similar ACC compared to the baselines. Compared with the case where more intermediate layers are treated Bayesian (LA Ours), last layer Laplace approximation in general has lower accuracies and higher NLPD and ECE.
  }
  \vspace*{-6pt}
	\setlength{\tabcolsep}{8.5pt}
	\scriptsize
\centering
\begin{tabular}{l|l|ccccc}
\toprule
Metrics & Methods & \sc CIFAR-10 & \sc CIFAR-100  & \sc DTD  & \sc RESISC & \sc Imagenet-R \\
\hline
    \multirow{3}{*}{ACC $\uparrow$}  & LL-LA GLM & \val{\bf}{0.966}{0.002} & \val{}{0.850}{0.004} & \val{}{0.795}{0.005} & \val{\bf}{0.583}{0.011} & \val{}{0.632}{0.013}\\
 & LL-LA Sampling & \val{\bf}{0.967}{0.002} & \val{}{0.850}{0.004} & \val{}{0.795}{0.005} & \val{}{0.526}{0.012} & \val{\bf}{0.633}{0.013}\\
 \rowcolor{highlight}\cellcolor{white}
 & LL-LA Ours & \val{\bf}{0.967}{0.002} & \val{\bf}{0.853}{0.004} & \val{\bf}{0.803}{0.005} & \val{\bf}{0.585}{0.011} & \val{\bf}{0.639}{0.013}\\ \cdashline{2-7} \rule{0pt}{2.25ex}
 & LA Ours & \val{}{0.974}{0.002} & \val{}{0.879}{0.003} & \val{}{0.725}{0.010} & \val{}{0.906}{0.004} & \val{}{0.725}{0.012}\\ 
 \hline
\multirow{3}{*}{NLPD $\downarrow$} & LL-LA GLM & \val{}{0.127}{0.004} & \val{}{0.791}{0.010} & \val{}{0.951}{0.011} & \val{}{2.622}{0.018} & \val{}{2.134}{0.038}\\
 & LL-LA Sampling  & \val{}{0.131}{0.004} & \val{}{0.708}{0.010} & \val{}{0.907}{0.011} & \val{}{2.485}{0.022} & \val{}{1.976}{0.041}\\
  \rowcolor{highlight}\cellcolor{white}
 & LL-LA Ours & \val{\bf}{0.106}{0.005} & \val{\bf}{0.541}{0.011} & \val{\bf}{0.731}{0.012} & \val{\bf}{2.273}{0.023} & \val{\bf}{1.832}{0.041}\\ \cdashline{2-7} \rule{0pt}{2.25ex}
 & LA Ours & \val{}{0.087}{0.005} & \val{}{0.426}{0.011} & \val{}{0.981}{0.030} & \val{}{0.297}{0.011} & \val{}{1.192}{0.042}\\  
   \hline
\multirow{3}{*}{ECE $\downarrow$} & LL-LA GLM & $0.033$ & $0.217$ & $0.263$ & $0.471$ & $0.355$\\
 & LL-LA Sampling & $0.039$ & $0.177$ & $0.245$ & $\bf0.377$ & $0.304$\\
 \rowcolor{highlight}\cellcolor{white}
& LL-LA Ours & $\bf0.007$ & $\bf0.041$ & $\bf0.125$ & $0.398$ & $\bf0.245$\\ \cdashline{2-7} \rule{0pt}{2.25ex}
& LA Ours & $0.007$ & $0.017$ & $0.041$ & $0.013$ & $0.089$\\ 
\bottomrule
\end{tabular}
\label{table:vit-last-layer}
\end{table}

\begin{table}
\centering
\setlength{\tabcolsep}{9.5pt}
\scriptsize
\caption{Wallclock times for Last layer ViT base on CIFAR-10 in milliseconds.}
\begin{tabular}{l|l|r}
  \toprule
  Model & Methods & \multicolumn{1}{c}{\sc Avg. Runtime ($\pm$ std) $\downarrow$} \\
  \midrule
   & MAP & \val{}{3.732}{0.091} \\ 
   & LA Sampling & \val{}{188.732}{0.051} \\
   \cellcolor{white} Last Layer ViT   & LA GLM & \val{}{5.517}{0.033}  \\
  \rowcolor{highlight} \cellcolor{white}
    & Ours & \val{}{3.782}{0.088} \\ 
  \bottomrule
\end{tabular}
\label{table:wall-clock-last-layer}
\end{table}

\subsection{Image Pixel Sensitivity} \label{app:experiment-image-sensitivity}

\begin{figure}[t]
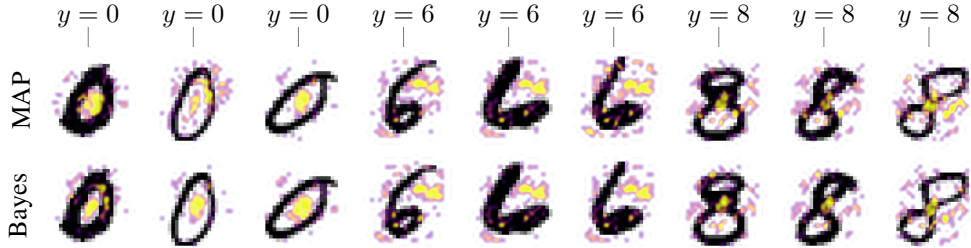

  \centering
  \setlength{\figsize}{0.10\textwidth}
  \def\images{0_2, 0_4, 0_7, 6_1, 6_6, 6_7, 8_0, 8_7, 8_6}
  \begin{tikzpicture}
    \foreach \x [count=\xi] in {0, 0, 0, 6, 6, 6, 8, 8, 8} {
      \draw[gray] (\xi*\figsize,0) -- (\xi*\figsize,1.7*\figsize);
      \node at (\xi*\figsize,1.8*\figsize) {$y=\x$};
    }
    \foreach \img [count=\xi] in \images {
      \node[inner sep = 0pt] at (\xi*\figsize,\figsize) {\includegraphics[width=\figsize]{imgs/input_perturbation/individual/mnist_0-8-6_map_\img}};
    }
    \foreach \img [count=\xi] in \images {
      \node[inner sep = 0pt] at (\xi*\figsize,0) {\includegraphics[width=\figsize]{imgs/input_perturbation/individual/mnist_0-8-6_bayes_\img}};
    }
    \node[rotate=90] at (0.5, \figsize) {MAP};
    \node[rotate=90] at (0.5, 0) {Bayes};
  \end{tikzpicture}
  \caption{Pixel sensitivity of MLP classifiers trained on binary classification tasks ($0/6/8$) for MNIST digits. The rows show the sensitivity of the MAP predictor to pixel perturbations, and the pixel sensitivity for a last-layer Laplace approximation. The predictive distribution is approximated analytically in both cases. We observe that the Bayesian model using a Laplace approximation has less spurious sensitivities to pixel perturbations indicating that it is more robust to input perturbations. The sensitivities are visualised in the range (0.5~\protect\includegraphics[width=3em,height=.7em]{fig/cool.png}~1.0)  
  } 
  \label{fig:mnist-sensitivity}
\end{figure}

We trained a $4$ layer MLP classifier on MNIST digits zero and eight using a batch size of $64$, learning rate of $1e-3$, weight decay set to $1e-5$, and for $50$ epochs.
We used a subset of $0.1\%$ of the training data as held-out validation set and assumed a full covariance Gaussian distribution for each input centred at the pixel values of the datum and with a fixed covariance of $1e-5$.
Furthermore, we then computed the pixel sensitivities for the trained model by learning the pixel-wise input covariance matrices by minimizing the negative log-likelihood of the held-out validation set and jointly maximizing the entropy of the input distributions.
The optimisation was performed for each image independently and using Adam with a learning rate of $5e-3$ until the validation loss dropped below a divergence to the initial loss of $1e-2$. Doing so typically took around $700$ iterations.
\cref{fig:mnist-sensitivity} shows some additional examples with the input-dependent sensitivities. 

\tikzexternaldisable

\subsection{Effect of the Number of MC Samples on Performance} \label{app:experiment-mcsamples}
We investigate the influence of number of samples on performance. 

\begin{figure}[h!]
  \centering\scriptsize
  \setlength{\figurewidth}{.4\textwidth}
  \setlength{\figureheight}{.18\textwidth}
  \pgfplotsset{
    height=\figureheight,
    width=\figurewidth,
    tick align=inside, 
    x tick label style={font=\tiny}, 
    y tick label style={rotate=45, font=\tiny, /pgf/number format/.cd, zerofill, precision=3}, 
    y label style = {yshift=0em, font=\scriptsize}, 
    x label style = {yshift=0.5em}, scale only axis, 
    legend cell align={left},
  }
  \begin{subfigure}[t]{0.49\textwidth}
    \centering
    \pgfplotsset{ 
      ylabel = {$\leftarrow$ NLPD},
      title = {\small \bfseries MFVI on {\sc CIFAR-10} dataset}
    }
    % This file was created with tikzplotlib v0.10.1.post12.
\begin{tikzpicture}

\definecolor{darkgray176}{RGB}{176,176,176}

\begin{axis}[
height=\figureheight,
tick align=outside,
tick pos=left,
width=\figurewidth,
x grid style={darkgray176},
xmin=5.5, xmax=104.5,
xtick style={color=black},
y grid style={darkgray176},
ymin=0.126977540958761, ymax=0.152721717107923,
ytick style={color=black}
]
\path [draw=blue, semithick]
(axis cs:10,0.144582586865973)
--(axis cs:10,0.151551527282961);

\path [draw=blue, semithick]
(axis cs:20,0.139275586880671)
--(axis cs:20,0.141552987282682);

\path [draw=blue, semithick]
(axis cs:30,0.136209554123393)
--(axis cs:30,0.140446422512312);

\path [draw=blue, semithick]
(axis cs:40,0.133885833579779)
--(axis cs:40,0.138577306437829);

\path [draw=blue, semithick]
(axis cs:50,0.132689295472451)
--(axis cs:50,0.136391270569031);

\path [draw=blue, semithick]
(axis cs:60,0.133043477099775)
--(axis cs:60,0.136776179091555);

\path [draw=blue, semithick]
(axis cs:70,0.132932354545388)
--(axis cs:70,0.135847858409671);

\path [draw=blue, semithick]
(axis cs:80,0.131379014125506)
--(axis cs:80,0.1344541338327);

\path [draw=blue, semithick]
(axis cs:90,0.128147730783723)
--(axis cs:90,0.133734753122127);

\path [draw=blue, semithick]
(axis cs:100,0.131033562348917)
--(axis cs:100,0.135072739889337);

\addplot [semithick, blue, mark=-, mark size=5, mark options={solid}, only marks]
table {%
10 0.144582586865973
20 0.139275586880671
30 0.136209554123393
40 0.133885833579779
50 0.132689295472451
60 0.133043477099775
70 0.132932354545388
80 0.131379014125506
90 0.128147730783723
100 0.131033562348917
};
\addplot [semithick, blue, mark=-, mark size=5, mark options={solid}, only marks]
table {%
10 0.151551527282961
20 0.141552987282682
30 0.140446422512312
40 0.138577306437829
50 0.136391270569031
60 0.136776179091555
70 0.135847858409671
80 0.1344541338327
90 0.133734753122127
100 0.135072739889337
};
\addplot [semithick, black, dashed]
table {%
50 0.126977540958761
50 0.152721717107923
};
\addplot [semithick, scBlue, mark=*, mark size=2, mark options={solid}]
table {%
10 0.148067057074467
20 0.140414287081676
30 0.138327988317852
40 0.136231570008804
50 0.134540283020741
60 0.134909828095665
70 0.134390106477529
80 0.132916573979103
90 0.130941241952925
100 0.133053151119127
};
\end{axis}

\end{tikzpicture}
  \end{subfigure}
  \hfill
  \begin{subfigure}[t]{0.49\textwidth}
    \centering
    \pgfplotsset{ 
      title = {\small \bfseries LA on {\sc CIFAR-10} dataset}
    }
    % This file was created with tikzplotlib v0.10.1.post12.
\begin{tikzpicture}

\definecolor{darkgray176}{RGB}{176,176,176}

\begin{axis}[
height=\figureheight,
tick align=outside,
tick pos=left,
width=\figurewidth,
x grid style={darkgray176},
xmin=5.5, xmax=104.5,
xtick style={color=black},
y grid style={darkgray176},
ymin=0.167721159815787, ymax=0.18408965673465,
ytick style={color=black}
]
\path [draw=blue, semithick]
(axis cs:10,0.176531258499628)
--(axis cs:10,0.183345634147429);

\path [draw=blue, semithick]
(axis cs:20,0.172702795268064)
--(axis cs:20,0.175334121902388);

\path [draw=blue, semithick]
(axis cs:30,0.170587525198266)
--(axis cs:30,0.172190985142685);

\path [draw=blue, semithick]
(axis cs:40,0.16895794525313)
--(axis cs:40,0.171796042835434);

\path [draw=blue, semithick]
(axis cs:50,0.169373028957593)
--(axis cs:50,0.170472655814792);

\path [draw=blue, semithick]
(axis cs:60,0.168465182403008)
--(axis cs:60,0.171612976959104);

\path [draw=blue, semithick]
(axis cs:70,0.168731011195111)
--(axis cs:70,0.170395376651904);

\path [draw=blue, semithick]
(axis cs:80,0.168502043597211)
--(axis cs:80,0.171044005822196);

\path [draw=blue, semithick]
(axis cs:90,0.168548478241366)
--(axis cs:90,0.170329939871004);

\path [draw=blue, semithick]
(axis cs:100,0.168849375765959)
--(axis cs:100,0.170184037797194);

\addplot [semithick, blue, mark=-, mark size=5, mark options={solid}, only marks]
table {%
10 0.176531258499628
20 0.172702795268064
30 0.170587525198266
40 0.16895794525313
50 0.169373028957593
60 0.168465182403008
70 0.168731011195111
80 0.168502043597211
90 0.168548478241366
100 0.168849375765959
};
\addplot [semithick, blue, mark=-, mark size=5, mark options={solid}, only marks]
table {%
10 0.183345634147429
20 0.175334121902388
30 0.172190985142685
40 0.171796042835434
50 0.170472655814792
60 0.171612976959104
70 0.170395376651904
80 0.171044005822196
90 0.170329939871004
100 0.170184037797194
};
\addplot [semithick, black, dashed]
table {%
50 0.167721159815787
50 0.18408965673465
};
\addplot [semithick, scBlue, mark=*, mark size=2, mark options={solid}]
table {%
10 0.179938446323529
20 0.174018458585226
30 0.171389255170476
40 0.170376994044282
50 0.169922842386192
60 0.170039079681056
70 0.169563193923508
80 0.169773024709703
90 0.169439209056185
100 0.169516706781576
};
\end{axis}

\end{tikzpicture}
  \end{subfigure}
  \hfill
  \begin{subfigure}[t]{0.49\textwidth}
    \centering
    \pgfplotsset{ 
      ylabel = {ACC $\rightarrow$}
    }
  % This file was created with tikzplotlib v0.10.1.post12.
\begin{tikzpicture}

\definecolor{darkgray176}{RGB}{176,176,176}

\begin{axis}[
height=\figureheight,
tick align=outside,
tick pos=left,
width=\figurewidth,
x grid style={darkgray176},
xmin=5.5, xmax=104.5,
xtick style={color=black},
y grid style={darkgray176},
ymin=0.973444051905411, ymax=0.975715948094589,
ytick style={color=black}
]
\path [draw=blue, semithick]
(axis cs:10,0.973752948309953)
--(axis cs:10,0.975207051690047);

\path [draw=blue, semithick]
(axis cs:20,0.973725374230251)
--(axis cs:20,0.975114625769749);

\path [draw=blue, semithick]
(axis cs:30,0.973967676425409)
--(axis cs:30,0.975192323574591);

\path [draw=blue, semithick]
(axis cs:40,0.973654280057166)
--(axis cs:40,0.975025719942834);

\path [draw=blue, semithick]
(axis cs:50,0.974221537694538)
--(axis cs:50,0.974858462305462);

\path [draw=blue, semithick]
(axis cs:60,0.974026227193411)
--(axis cs:60,0.974933772806589);

\path [draw=blue, semithick]
(axis cs:70,0.97354731991401)
--(axis cs:70,0.97561268008599);

\path [draw=blue, semithick]
(axis cs:80,0.974146654060877)
--(axis cs:80,0.974733345939123);

\path [draw=blue, semithick]
(axis cs:90,0.973870392156925)
--(axis cs:90,0.974849607843074);

\path [draw=blue, semithick]
(axis cs:100,0.974191940006249)
--(axis cs:100,0.974768059993751);

\addplot [semithick, blue, mark=-, mark size=5, mark options={solid}, only marks]
table {%
10 0.973752948309953
20 0.973725374230251
30 0.973967676425409
40 0.973654280057166
50 0.974221537694538
60 0.974026227193411
70 0.97354731991401
80 0.974146654060877
90 0.973870392156925
100 0.974191940006249
};
\addplot [semithick, blue, mark=-, mark size=5, mark options={solid}, only marks]
table {%
10 0.975207051690047
20 0.975114625769749
30 0.975192323574591
40 0.975025719942834
50 0.974858462305462
60 0.974933772806589
70 0.97561268008599
80 0.974733345939123
90 0.974849607843074
100 0.974768059993751
};
\addplot [semithick, black, dashed]
table {%
50 0.973444051905411
50 0.975715948094589
};
\addplot [semithick, scBlue, mark=*, mark size=2, mark options={solid}]
table {%
10 0.97448
20 0.97442
30 0.97458
40 0.97434
50 0.97454
60 0.97448
70 0.97458
80 0.97444
90 0.97436
100 0.97448
};
\end{axis}

\end{tikzpicture}
  \end{subfigure}
  \hfill
  \begin{subfigure}[t]{0.49\textwidth}
    \centering
    \pgfplotsset{ 
    }
  % This file was created with tikzplotlib v0.10.1.post12.
\begin{tikzpicture}

\definecolor{darkgray176}{RGB}{176,176,176}

\begin{axis}[
height=\figureheight,
tick align=outside,
tick pos=left,
width=\figurewidth,
x grid style={darkgray176},
xmin=5.5, xmax=104.5,
xtick style={color=black},
y grid style={darkgray176},
ymin=0.964634175143264, ymax=0.972566784868615,
ytick style={color=black}
]
\path [draw=blue, semithick]
(axis cs:10,0.964994748312599)
--(axis cs:10,0.967245251687402);

\path [draw=blue, semithick]
(axis cs:20,0.966560146347993)
--(axis cs:20,0.969839853652007);

\path [draw=blue, semithick]
(axis cs:30,0.968434447487018)
--(axis cs:30,0.970725552512982);

\path [draw=blue, semithick]
(axis cs:40,0.968754722005666)
--(axis cs:40,0.971805277994334);

\path [draw=blue, semithick]
(axis cs:50,0.970033540793077)
--(axis cs:50,0.971006459206923);

\path [draw=blue, semithick]
(axis cs:60,0.968854173891009)
--(axis cs:60,0.97158582610899);

\path [draw=blue, semithick]
(axis cs:70,0.968124617618235)
--(axis cs:70,0.972195382381765);

\path [draw=blue, semithick]
(axis cs:80,0.970010837850681)
--(axis cs:80,0.971469162149319);

\path [draw=blue, semithick]
(axis cs:90,0.970513788300719)
--(axis cs:90,0.972206211699281);

\path [draw=blue, semithick]
(axis cs:100,0.970121908079452)
--(axis cs:100,0.971638091920548);

\addplot [semithick, blue, mark=-, mark size=5, mark options={solid}, only marks]
table {%
10 0.964994748312599
20 0.966560146347993
30 0.968434447487018
40 0.968754722005666
50 0.970033540793077
60 0.968854173891009
70 0.968124617618235
80 0.970010837850681
90 0.970513788300719
100 0.970121908079452
};
\addplot [semithick, blue, mark=-, mark size=5, mark options={solid}, only marks]
table {%
10 0.967245251687402
20 0.969839853652007
30 0.970725552512982
40 0.971805277994334
50 0.971006459206923
60 0.97158582610899
70 0.972195382381765
80 0.971469162149319
90 0.972206211699281
100 0.971638091920548
};
\addplot [semithick, black, dashed]
table {%
50 0.964634175143264
50 0.972566784868615
};
\addplot [semithick, scBlue, mark=*, mark size=2, mark options={solid}]
table {%
10 0.96612
20 0.9682
30 0.96958
40 0.97028
50 0.97052
60 0.97022
70 0.97016
80 0.97074
90 0.97136
100 0.97088
};
\end{axis}

\end{tikzpicture}
  \end{subfigure}
  \hfill
   \begin{subfigure}[t]{0.49\textwidth}
    \centering
    \pgfplotsset{ 
      xlabel={Number of MC samples}, 
      ylabel = {Time (min)},
      y tick label style={precision=0}, 
    }
  % This file was created with tikzplotlib v0.10.1.post12.
\begin{tikzpicture}

\definecolor{darkgray176}{RGB}{176,176,176}

\begin{axis}[
height=\figureheight,
tick align=outside,
tick pos=left,
width=\figurewidth,
x grid style={darkgray176},
xmin=5.5, xmax=104.5,
xtick style={color=black},
y grid style={darkgray176},
ymin=3.464165, ymax=63.652535,
ytick style={color=black}
]
\addplot [semithick, scBlue, mark=*, mark size=2, mark options={solid}]
table {%
10 6.2
20 12.25
30 18.3167
40 24.4
50 30.5667
60 36.5667
70 42.7667
80 48.7667
90 54.8667
100 60.9167
};
\addplot [semithick, black, dashed]
table {%
50 3.464165
50 63.652535
};
\end{axis}

\end{tikzpicture}
  \end{subfigure}
  \hfill
  \begin{subfigure}[t]{0.49\textwidth}
    \centering
    \pgfplotsset{ 
      xlabel={Number of MC samples}, 
      y tick label style={precision=0}, 
    }
  % This file was created with tikzplotlib v0.10.1.post12.
\begin{tikzpicture}

\definecolor{darkgray176}{RGB}{176,176,176}

\begin{axis}[
height=\figureheight,
tick align=outside,
tick pos=left,
width=\figurewidth,
x grid style={darkgray176},
xmin=5.5, xmax=104.5,
xtick style={color=black},
y grid style={darkgray176},
ymin=1.31413, ymax=22.63587,
ytick style={color=black}
]
\addplot [semithick, scBlue, mark=*, mark size=2, mark options={solid}]
table {%
10 2.2833
20 4.4333
30 6.5833
40 8.75
50 10.9
60 13.05
70 15.2
80 17.4333
90 19.5167
100 21.6667
};
\addplot [semithick, black, dashed]
table {%
50 1.31413
50 22.63587
};
\end{axis}

\end{tikzpicture}
  \end{subfigure}
  \hfill
  \caption{Effects of the number of MC samples on performance for LA and MFVI on CIFAR-10 with ViT-Base model. In the results reported in main paper, we set the number of MC samples to $50$ (dashed line).}
  \label{fig:mc_vit}
\end{figure}
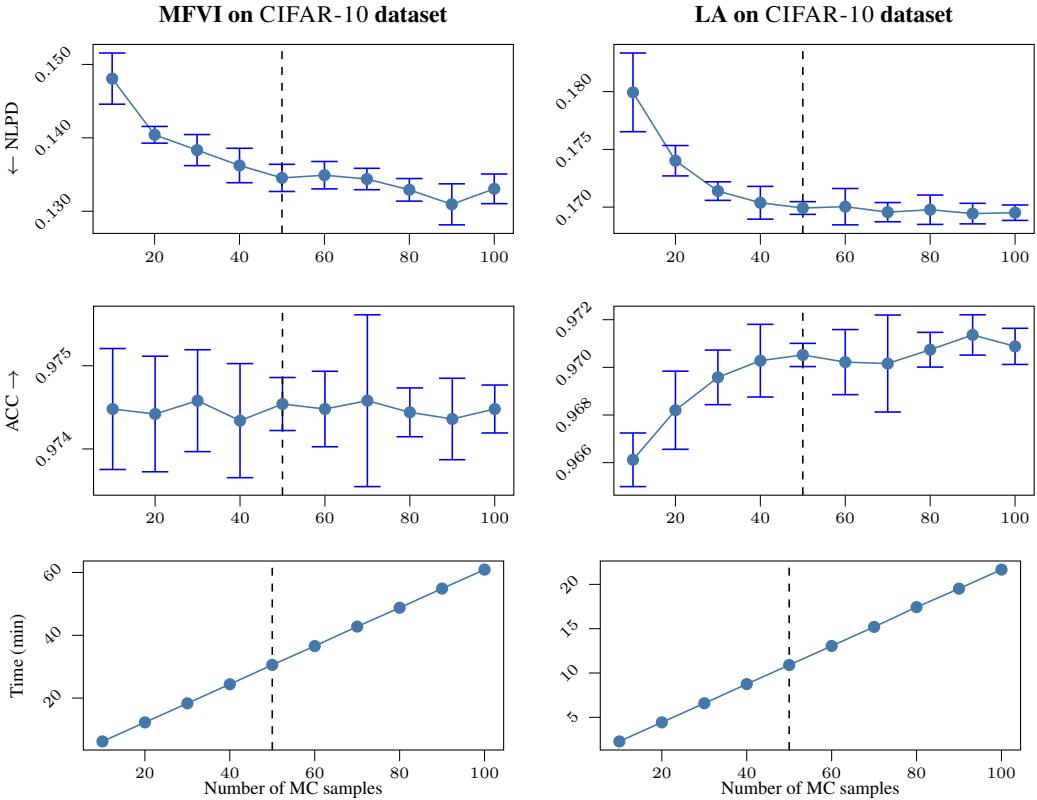

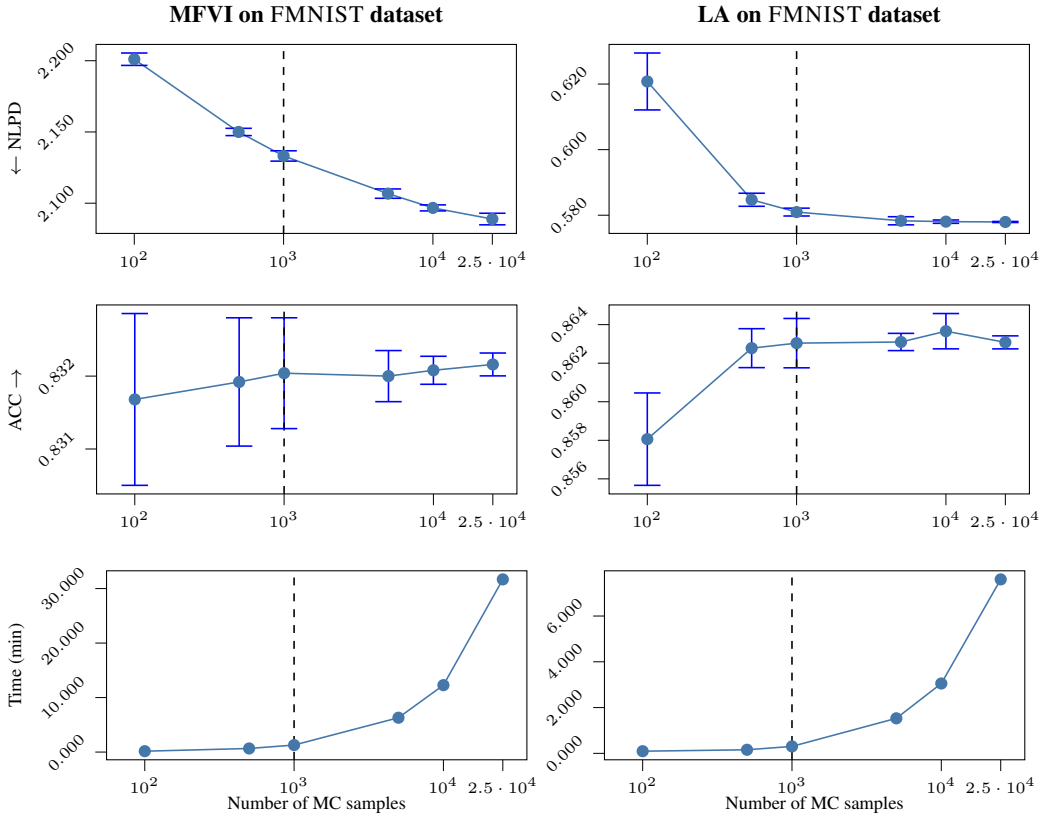
\begin{figure}[h!]
  \centering\scriptsize
  \setlength{\figurewidth}{.4\textwidth}
  \setlength{\figureheight}{.18\textwidth}
  \pgfplotsset{
    height=\figureheight,
    width=\figurewidth,
    tick align=inside, 
    x tick label style={font=\tiny}, 
    y tick label style={rotate=45, font=\tiny, /pgf/number format/.cd, zerofill, precision=3}, 
    y label style = {yshift=0em, font=\scriptsize}, 
    x label style = {yshift=0.5em}, scale only axis, 
	xtick = {100,  1000, 10000, 25000}, 
	xticklabels = {$10^2$,  $10^3$,  $10^4$, $2.5\cdot 10^4$}, 
    legend cell align={left},
  }
  \begin{subfigure}[t]{0.49\textwidth}
    \centering
    \pgfplotsset{ 
      ylabel = {$\leftarrow$ NLPD},
      title = {\small \bfseries MFVI on {\sc FMNIST} dataset}
    }
    % This file was created with tikzplotlib v0.10.1.post12.
\begin{tikzpicture}

\definecolor{darkgray176}{RGB}{176,176,176}

\begin{axis}[
xmode=log,
height=\figureheight,
tick align=outside,
tick pos=left,
width=\figurewidth,
x grid style={darkgray176},
xmin=-1145, xmax=36245,
xtick style={color=black},
y grid style={darkgray176},
ymin=2.07878569556425, ymax=2.21144006677601,
ytick style={color=black}
]
\path [draw=blue, semithick]
(axis cs:100,2.19671509405295)
--(axis cs:100,2.20541032263002);

\path [draw=blue, semithick]
(axis cs:500,2.14744901990628)
--(axis cs:500,2.15256303062295);

\path [draw=blue, semithick]
(axis cs:1000,2.12952971993516)
--(axis cs:1000,2.1367799077468);

\path [draw=blue, semithick]
(axis cs:5000,2.10344684223982)
--(axis cs:5000,2.11000793515879);

\path [draw=blue, semithick]
(axis cs:10000,2.09458519993912)
--(axis cs:10000,2.09882464530101);

\path [draw=blue, semithick]
(axis cs:25000,2.08481543971023)
--(axis cs:25000,2.09294545746044);

\addplot [semithick, blue, mark=-, mark size=5, mark options={solid}, only marks]
table {%
100 2.19671509405295
500 2.14744901990628
1000 2.12952971993516
5000 2.10344684223982
10000 2.09458519993912
25000 2.08481543971023
};
\addplot [semithick, blue, mark=-, mark size=5, mark options={solid}, only marks]
table {%
100 2.20541032263002
500 2.15256303062295
1000 2.1367799077468
5000 2.11000793515879
10000 2.09882464530101
25000 2.09294545746044
};
\addplot [semithick, black, dashed]
table {%
1000 2.07878569556425
1000 2.21144006677601
};
\addplot [semithick, scBlue, mark=*, mark size=2, mark options={solid}]
table {%
100 2.20106270834149
500 2.15000602526461
1000 2.13315481384098
5000 2.1067273886993
10000 2.09670492262006
25000 2.08888044858534
};
\end{axis}

\end{tikzpicture}
  \end{subfigure}
  \hfill
  \begin{subfigure}[t]{0.49\textwidth}
    \centering
    \pgfplotsset{ 
      title = {\small \bfseries LA on {\sc FMNIST} dataset}
    }
    % This file was created with tikzplotlib v0.10.1.post12.
\begin{tikzpicture}

\definecolor{darkgray176}{RGB}{176,176,176}

\begin{axis}[
xmode=log,
height=\figureheight,
tick align=outside,
tick pos=left,
width=\figurewidth,
x grid style={darkgray176},
xmin=-1145, xmax=36245,
xtick style={color=black},
y grid style={darkgray176},
ymin=0.574448471378983, ymax=0.632085919562256,
ytick style={color=black}
]
\path [draw=blue, semithick]
(axis cs:100,0.61210671418823)
--(axis cs:100,0.629466035553926);

\path [draw=blue, semithick]
(axis cs:500,0.582720985864686)
--(axis cs:500,0.586705051817768);

\path [draw=blue, semithick]
(axis cs:1000,0.579763015631655)
--(axis cs:1000,0.582152728961983);

\path [draw=blue, semithick]
(axis cs:5000,0.577068355387313)
--(axis cs:5000,0.579541602136242);

\path [draw=blue, semithick]
(axis cs:10000,0.577528467835639)
--(axis cs:10000,0.578552486658522);

\path [draw=blue, semithick]
(axis cs:25000,0.577772576976928)
--(axis cs:25000,0.578100509742161);

\addplot [semithick, blue, mark=-, mark size=5, mark options={solid}, only marks]
table {%
100 0.61210671418823
500 0.582720985864686
1000 0.579763015631655
5000 0.577068355387313
10000 0.577528467835639
25000 0.577772576976928
};
\addplot [semithick, blue, mark=-, mark size=5, mark options={solid}, only marks]
table {%
100 0.629466035553926
500 0.586705051817768
1000 0.582152728961983
5000 0.579541602136242
10000 0.578552486658522
25000 0.578100509742161
};
\addplot [semithick, black, dashed]
table {%
1000 0.574448471378983
1000 0.632085919562256
};
\addplot [semithick, scBlue, mark=*, mark size=2, mark options={solid}]
table {%
100 0.620786374871078
500 0.584713018841227
1000 0.580957872296819
5000 0.578304978761778
10000 0.57804047724708
25000 0.577936543359545
};
\end{axis}

\end{tikzpicture}
  \end{subfigure}
  \hfill
  \begin{subfigure}[t]{0.49\textwidth}
    \centering
    \pgfplotsset{ 
      ylabel = {ACC $\rightarrow$}
    }
    % This file was created with tikzplotlib v0.10.1.post12.
\begin{tikzpicture}

\definecolor{darkgray176}{RGB}{176,176,176}

\begin{axis}[
xmode=log,
height=\figureheight,
tick align=outside,
tick pos=left,
width=\figurewidth,
x grid style={darkgray176},
xmin=-1145, xmax=36245,
xtick style={color=black},
y grid style={darkgray176},
ymin=0.830691764260165, ymax=0.831988235739835,
ytick style={color=black}
]
\path [draw=blue, semithick]
(axis cs:100,0.830750694781968)
--(axis cs:100,0.831929305218032);

\path [draw=blue, semithick]
(axis cs:500,0.831019981091315)
--(axis cs:500,0.831900018908685);

\path [draw=blue, semithick]
(axis cs:1000,0.831139941899179)
--(axis cs:1000,0.831900058100821);

\path [draw=blue, semithick]
(axis cs:5000,0.831324692270564)
--(axis cs:5000,0.831675307729436);

\path [draw=blue, semithick]
(axis cs:10000,0.831443980002083)
--(axis cs:10000,0.831636019997917);

\path [draw=blue, semithick]
(axis cs:25000,0.8315016)
--(axis cs:25000,0.8316584);

\addplot [semithick, blue, mark=-, mark size=5, mark options={solid}, only marks]
table {%
100 0.830750694781968
500 0.831019981091315
1000 0.831139941899179
5000 0.831324692270564
10000 0.831443980002083
25000 0.8315016
};
\addplot [semithick, blue, mark=-, mark size=5, mark options={solid}, only marks]
table {%
100 0.831929305218032
500 0.831900018908685
1000 0.831900058100821
5000 0.831675307729436
10000 0.831636019997917
25000 0.8316584
};
\addplot [semithick, black, dashed]
table {%
1000 0.830691764260164
1000 0.831988235739835
};
\addplot [semithick, scBlue, mark=*, mark size=2, mark options={solid}]
table {%
100 0.83134
500 0.83146
1000 0.83152
5000 0.8315
10000 0.83154
25000 0.83158
};
\end{axis}

\end{tikzpicture}
  \end{subfigure}
  \hfill
  \begin{subfigure}[t]{0.49\textwidth}
    \centering
    \pgfplotsset{ 
    }
  % This file was created with tikzplotlib v0.10.1.post12.
\begin{tikzpicture}

\definecolor{darkgray176}{RGB}{176,176,176}

\begin{axis}[
xmode=log,
height=\figureheight,
tick align=outside,
tick pos=left,
width=\figurewidth,
x grid style={darkgray176},
xmin=-1145, xmax=36245,
xtick style={color=black},
y grid style={darkgray176},
ymin=0.855218385814038, ymax=0.865021571762631,
ytick style={color=black}
]
\path [draw=blue, semithick]
(axis cs:100,0.855663985175338)
--(axis cs:100,0.860456014824662);

\path [draw=blue, semithick]
(axis cs:500,0.861769886659825)
--(axis cs:500,0.863790113340175);

\path [draw=blue, semithick]
(axis cs:1000,0.861760133506963)
--(axis cs:1000,0.864319866493037);

\path [draw=blue, semithick]
(axis cs:5000,0.862653051233361)
--(axis cs:5000,0.863546948766639);

\path [draw=blue, semithick]
(axis cs:10000,0.862744027598669)
--(axis cs:10000,0.864575972401331);

\path [draw=blue, semithick]
(axis cs:25000,0.862742788849532)
--(axis cs:25000,0.863417211150468);

\addplot [semithick, blue, mark=-, mark size=5, mark options={solid}, only marks]
table {%
100 0.855663985175338
500 0.861769886659825
1000 0.861760133506963
5000 0.862653051233361
10000 0.862744027598669
25000 0.862742788849532
};
\addplot [semithick, blue, mark=-, mark size=5, mark options={solid}, only marks]
table {%
100 0.860456014824662
500 0.863790113340175
1000 0.864319866493037
5000 0.863546948766639
10000 0.864575972401331
25000 0.863417211150468
};
\addplot [semithick, black, dashed]
table {%
1000 0.855218385814038
1000 0.865021571762631
};
\addplot [semithick, scBlue, mark=*, mark size=2, mark options={solid}]
table {%
100 0.85806
500 0.86278
1000 0.86304
5000 0.8631
10000 0.86366
25000 0.86308
};
\end{axis}

\end{tikzpicture}
  \end{subfigure}
  \hfill
  \begin{subfigure}[t]{0.49\textwidth}
    \centering
    \pgfplotsset{ 
      xlabel={Number of MC samples}, 
      ylabel = {Time (min)},
    }
    % This file was created with tikzplotlib v0.10.1.post12.
\begin{tikzpicture}

\definecolor{darkgray176}{RGB}{176,176,176}

\begin{axis}[
xmode=log,
height=\figureheight,
tick align=outside,
tick pos=left,
width=\figurewidth,
x grid style={darkgray176},
xmin=-1145, xmax=36245,
xtick style={color=black},
y grid style={darkgray176},
ymin=-1.41747325758139, ymax=33.262763556838,
ytick style={color=black}
]
\addplot [semithick, scBlue, mark=*, mark size=2, mark options={solid}]
table {%
100 0.158901143074036
500 0.658502753575643
1000 1.28663577238719
5000 6.30121177434921
10000 12.279457394282
25000 31.6863891561826
};
\addplot [semithick, black, dashed]
table {%
1000 -1.41747325758139
1000 33.262763556838
};
\end{axis}

\end{tikzpicture}
  \end{subfigure}
  \hfill
  \begin{subfigure}[t]{0.49\textwidth}
    \centering
    \pgfplotsset{ 
      xlabel={Number of MC samples}, 
    }
  % This file was created with tikzplotlib v0.10.1.post12.
\begin{tikzpicture}

\definecolor{darkgray176}{RGB}{176,176,176}

\begin{axis}[
xmode=log,
height=\figureheight,
tick align=outside,
tick pos=left,
width=\figurewidth,
x grid style={darkgray176},
xmin=-1145, xmax=36245,
xtick style={color=black},
y grid style={darkgray176},
ymin=-0.281304899255435, ymax=7.97575006822745,
ytick style={color=black}
]
\addplot [semithick, scBlue, mark=*, mark size=2, mark options={solid}]
table {%
100 0.0940157810846965
500 0.156471455097198
1000 0.303456405798594
5000 1.53082801103592
10000 3.0523065884908
25000 7.60042938788732
};
\addplot [semithick, black, dashed]
table {%
1000 -0.281304899255435
1000 7.97575006822745
};
\end{axis}

\end{tikzpicture}
  \end{subfigure}
  \hfill
  \caption{Effects of the number of MC samples on performance for LA and MFVI on FMNIST. In the results reported in main paper, we set the number of MC samples to $1000$ (dashed line).}

  \label{fig:mc_classification}
\end{figure}

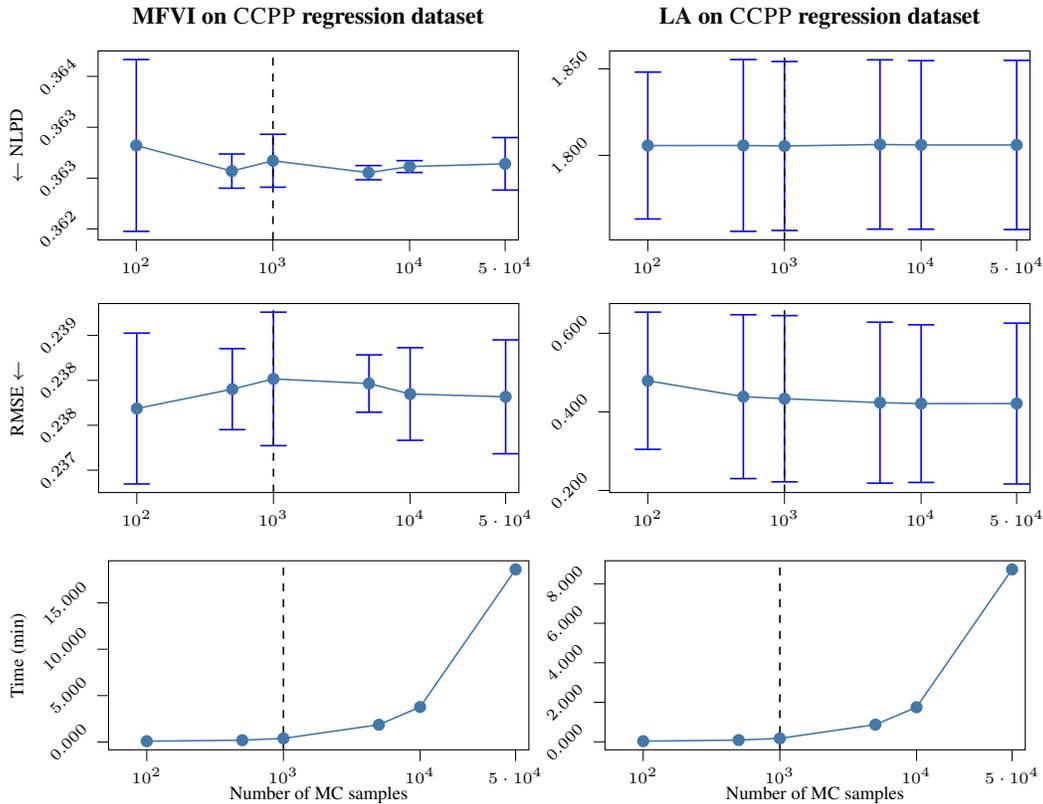
\begin{figure}[h!]
  \centering\scriptsize
  \setlength{\figurewidth}{.4\textwidth}
  \setlength{\figureheight}{.18\textwidth}
  \pgfplotsset{
    height=\figureheight,
    width=\figurewidth,
    tick align=inside, 
    x tick label style={font=\tiny}, 
    y tick label style={rotate=45, font=\tiny, /pgf/number format/.cd, zerofill, precision=3}, 
    y label style = {yshift=0em, font=\scriptsize}, 
    x label style = {yshift=0.5em}, scale only axis, 
	xtick = {100,  1000, 10000, 50000}, 
	xticklabels = {$10^2$,  $10^3$,  $10^4$, $5\cdot 10^4$}, 
    legend cell align={left},
  }
  \begin{subfigure}[t]{0.49\textwidth}
    \centering
    \pgfplotsset{ 
      ylabel = {$\leftarrow$ NLPD},
      title = {\small \bfseries MFVI on {\sc CCPP} regression dataset}
    }
    % This file was created with tikzplotlib v0.10.1.post12.
\begin{tikzpicture}

\definecolor{darkgray176}{RGB}{176,176,176}

\begin{axis}[
xmode=log,
height=\figureheight,
tick align=outside,
tick pos=left,
width=\figurewidth,
x grid style={darkgray176},
xmin=-2395, xmax=62495,
xtick style={color=black},
y grid style={darkgray176},
ymin=0.361892295895985, ymax=0.363749483208883,
ytick style={color=black}
]
\path [draw=blue, semithick]
(axis cs:100,0.361976713501117)
--(axis cs:100,0.363665065603751);

\path [draw=blue, semithick]
(axis cs:500,0.362401217181693)
--(axis cs:500,0.362737484813521);

\path [draw=blue, semithick]
(axis cs:1000,0.36241079263982)
--(axis cs:1000,0.362930903702013);

\path [draw=blue, semithick]
(axis cs:5000,0.362483978887202)
--(axis cs:5000,0.362623351991374);

\path [draw=blue, semithick]
(axis cs:10000,0.362555172569444)
--(axis cs:10000,0.362671955697844);

\path [draw=blue, semithick]
(axis cs:50000,0.362382564659753)
--(axis cs:50000,0.3628985209109);

\addplot [semithick, blue, mark=-, mark size=5, mark options={solid}, only marks]
table {%
100 0.361976713501117
500 0.362401217181693
1000 0.36241079263982
5000 0.362483978887202
10000 0.362555172569444
50000 0.362382564659753
};
\addplot [semithick, blue, mark=-, mark size=5, mark options={solid}, only marks]
table {%
100 0.363665065603751
500 0.362737484813521
1000 0.362930903702013
5000 0.362623351991374
10000 0.362671955697844
50000 0.3628985209109
};
\addplot [semithick, black, dashed]
table {%
1000 0.361892295895985
1000 0.363749483208883
};
\addplot [semithick, scBlue, mark=*, mark size=2, mark options={solid}]
table {%
100 0.362820889552434
500 0.362569350997607
1000 0.362670848170916
5000 0.362553665439288
10000 0.362613564133644
50000 0.362640542785327
};
\end{axis}

\end{tikzpicture}
  \end{subfigure}
  \hfill
   \begin{subfigure}[t]{0.49\textwidth}
    \centering
    \pgfplotsset{ 
      title = {\small \bfseries LA on {\sc CCPP} regression dataset}
    }
    % This file was created with tikzplotlib v0.10.1.post12.
\begin{tikzpicture}

\definecolor{darkgray176}{RGB}{176,176,176}

\begin{axis}[
xmode=log,
height=\figureheight,
tick align=outside,
tick pos=left,
width=\figurewidth,
x grid style={darkgray176},
xmin=-2395, xmax=62495,
xtick style={color=black},
y grid style={darkgray176},
ymin=1.75116037909348, ymax=1.86033684984685,
ytick style={color=black}
]
\path [draw=blue, semithick]
(axis cs:100,1.76328126894342)
--(axis cs:100,1.8481332716653);

\path [draw=blue, semithick]
(axis cs:500,1.75612294594591)
--(axis cs:500,1.85537428299442);

\path [draw=blue, semithick]
(axis cs:1000,1.7566549527022)
--(axis cs:1000,1.85424131836765);

\path [draw=blue, semithick]
(axis cs:5000,1.7574016856358)
--(axis cs:5000,1.85522272594714);

\path [draw=blue, semithick]
(axis cs:10000,1.75736128706316)
--(axis cs:10000,1.85475219350478);

\path [draw=blue, semithick]
(axis cs:50000,1.75717185991936)
--(axis cs:50000,1.85489027959175);

\addplot [semithick, blue, mark=-, mark size=5, mark options={solid}, only marks]
table {%
100 1.76328126894342
500 1.75612294594591
1000 1.7566549527022
5000 1.7574016856358
10000 1.75736128706316
50000 1.75717185991936
};
\addplot [semithick, blue, mark=-, mark size=5, mark options={solid}, only marks]
table {%
100 1.8481332716653
500 1.85537428299442
1000 1.85424131836765
5000 1.85522272594714
10000 1.85475219350478
50000 1.85489027959175
};
\addplot [semithick, black, dashed]
table {%
999.999999999999 1.75116037909348
999.999999999999 1.86033684984685
};
\addplot [semithick, scBlue, mark=*, mark size=2, mark options={solid}]
table {%
100 1.80570727030436
500 1.80574861447016
1000 1.80544813553492
5000 1.80631220579147
10000 1.80605674028397
50000 1.80603106975555
};
\end{axis}

\end{tikzpicture}
  \end{subfigure}
  \hfill
  \begin{subfigure}[t]{0.49\textwidth}
    \centering
    \pgfplotsset{ 
      ylabel = {RMSE $\leftarrow$}
    }
    % This file was created with tikzplotlib v0.10.1.post12.
\begin{tikzpicture}

\definecolor{darkgray176}{RGB}{176,176,176}

\begin{axis}[
xmode=log,
height=\figureheight,
tick align=outside,
tick pos=left,
width=\figurewidth,
x grid style={darkgray176},
xmin=-2395, xmax=62495,
xtick style={color=black},
y grid style={darkgray176},
ymin=0.237250950802396, ymax=0.239355331540666,
ytick style={color=black}
]
\path [draw=blue, semithick]
(axis cs:100,0.237346604472318)
--(axis cs:100,0.239026215070567);

\path [draw=blue, semithick]
(axis cs:500,0.237951429070029)
--(axis cs:500,0.238852904259172);

\path [draw=blue, semithick]
(axis cs:1000,0.237773235496209)
--(axis cs:1000,0.239259677870745);

\path [draw=blue, semithick]
(axis cs:5000,0.238145057270714)
--(axis cs:5000,0.238784034143102);

\path [draw=blue, semithick]
(axis cs:10000,0.237832775730855)
--(axis cs:10000,0.238863445700242);

\path [draw=blue, semithick]
(axis cs:50000,0.237683158776674)
--(axis cs:50000,0.238950887261317);

\addplot [semithick, blue, mark=-, mark size=5, mark options={solid}, only marks]
table {%
100 0.237346604472318
500 0.237951429070029
1000 0.237773235496209
5000 0.238145057270714
10000 0.237832775730855
50000 0.237683158776674
};
\addplot [semithick, blue, mark=-, mark size=5, mark options={solid}, only marks]
table {%
100 0.239026215070567
500 0.238852904259172
1000 0.239259677870745
5000 0.238784034143102
10000 0.238863445700242
50000 0.238950887261317
};
\addplot [semithick, black, dashed]
table {%
1000 0.237250950802396
1000 0.239355331540666
};
\addplot [semithick, scBlue, mark=*, mark size=2, mark options={solid}]
table {%
100 0.238186409771442
500 0.2384021666646
1000 0.238516456683477
5000 0.238464545706908
10000 0.238348110715548
50000 0.238317023018996
};
\end{axis}

\end{tikzpicture}
  \end{subfigure}
  \hfill
  \begin{subfigure}[t]{0.49\textwidth}
    \centering
    \pgfplotsset{ 
    }
  % This file was created with tikzplotlib v0.10.1.post12.
\begin{tikzpicture}

\definecolor{darkgray176}{RGB}{176,176,176}

\begin{axis}[
xmode=log,
height=\figureheight,
tick align=outside,
tick pos=left,
width=\figurewidth,
x grid style={darkgray176},
xmin=-2395, xmax=62495,
xtick style={color=black},
y grid style={darkgray176},
ymin=0.194630068688775, ymax=0.675679019416535,
ytick style={color=black}
]
\path [draw=blue, semithick]
(axis cs:100,0.304585732475482)
--(axis cs:100,0.653813158019818);

\path [draw=blue, semithick]
(axis cs:500,0.230138554498327)
--(axis cs:500,0.647211005365082);

\path [draw=blue, semithick]
(axis cs:1000,0.22190645657962)
--(axis cs:1000,0.645016678939408);

\path [draw=blue, semithick]
(axis cs:5000,0.218525812375079)
--(axis cs:5000,0.628286363137235);

\path [draw=blue, semithick]
(axis cs:10000,0.22034606533465)
--(axis cs:10000,0.621560401546784);

\path [draw=blue, semithick]
(axis cs:50000,0.216495930085492)
--(axis cs:50000,0.625877601176588);

\addplot [semithick, blue, mark=-, mark size=5, mark options={solid}, only marks]
table {%
100 0.304585732475482
500 0.230138554498327
1000 0.22190645657962
5000 0.218525812375079
10000 0.22034606533465
50000 0.216495930085492
};
\addplot [semithick, blue, mark=-, mark size=5, mark options={solid}, only marks]
table {%
100 0.653813158019818
500 0.647211005365082
1000 0.645016678939408
5000 0.628286363137235
10000 0.621560401546784
50000 0.625877601176588
};
\addplot [semithick, black, dashed]
table {%
1000 0.194630068688775
1000 0.675679019416535
};
\addplot [semithick, scBlue, mark=*, mark size=2, mark options={solid}]
table {%
100 0.47919944524765
500 0.438674779931704
1000 0.433461567759514
5000 0.423406087756157
10000 0.420953233440717
50000 0.42118676563104
};
\end{axis}

\end{tikzpicture}
  \end{subfigure}
  \hfill
  \begin{subfigure}[t]{0.49\textwidth}
    \centering
    \pgfplotsset{ 
      xlabel={Number of MC samples}, 
      ylabel = {Time (min)},
    }
    % This file was created with tikzplotlib v0.10.1.post12.
\begin{tikzpicture}

\definecolor{darkgray176}{RGB}{176,176,176}

\begin{axis}[
xmode=log,
height=\figureheight,
tick align=outside,
tick pos=left,
width=\figurewidth,
x grid style={darkgray176},
xmin=-2395, xmax=62495,
xtick style={color=black},
y grid style={darkgray176},
ymin=-0.84257117887338, ymax=19.5397548633814,
ytick style={color=black}
]
\addplot [semithick, scBlue, mark=*, mark size=2, mark options={solid}]
table {%
100 0.0838981866836548
500 0.190464516480764
1000 0.381777850786845
5000 1.85865616798401
10000 3.76989523967107
50000 18.6132854978244
};
\addplot [semithick, black, dashed]
table {%
1000 -0.84257117887338
1000 19.5397548633814
};
\end{axis}

\end{tikzpicture}
  \end{subfigure}
  \hfill
  \begin{subfigure}[t]{0.49\textwidth}
    \centering
    \pgfplotsset{ 
      xlabel={Number of MC samples}, 
    }
  % This file was created with tikzplotlib v0.10.1.post12.
\begin{tikzpicture}

\definecolor{darkgray176}{RGB}{176,176,176}

\begin{axis}[
xmode=log,
height=\figureheight,
tick align=outside,
tick pos=left,
width=\figurewidth,
x grid style={darkgray176},
xmin=-2395, xmax=62495,
xtick style={color=black},
y grid style={darkgray176},
ymin=-0.403762175838153, ymax=9.16388326307138,
ytick style={color=black}
]
\addplot [semithick, scBlue, mark=*, mark size=2, mark options={solid}]
table {%
100 0.0311307986577352
500 0.0869595249493917
1000 0.174017000198364
5000 0.868802587191264
10000 1.74966649611791
50000 8.72899028857549
};
\addplot [semithick, black, dashed]
table {%
1000 -0.403762175838153
1000 9.16388326307138
};
\end{axis}

\end{tikzpicture}
  \end{subfigure}
  \hfill
  \caption{Effects of the number of MC samples on performance for LA and MFVI on regression tasks. In the results reported in main paper, we set the number of MC samples to $1000$ (dashed line).}
  \label{fig:mc_regression}
\end{figure}

On regression tasks with small scale neural network (two layer MLP), we run experiments with the range of $[100, 500, 1000, 5000, 10000, 50000]$. On classification tasks with medium scale neural network (four layer MLP), we run experiments with the range of $[100, 500, 1000, 5000, 10000, 25000]$. On classification tasks with large scale neural network (ViT-Base), we run experiments with the range of $[10, 20, \ldots, 100]$. The results are reported in \cref{fig:mc_vit,fig:mc_classification,fig:mc_regression}. The number of samples we used to report results in the main paper is shown in the dashed line. 

In \cref{fig:mc_vit} and \cref{fig:mc_regression}, we observe that for regression on small scale networks and classification on large scale networks, the performance saturates when set the number of samples to $50$ samples (classification with ViT-Base) or $1000$ samples (regression with two layer MLP). In \cref{fig:mc_classification}, the performance saturates with $1000$ samples with LA. For MFVI the performance saturates with $5000$ samples, as the improvement gained on NLPD is marginal (from $2.13$ to $2.11$) from $1000$ samples to $5000$ samples, in experiment we set the number of samples as $1000$ for MFVI as well.

\subsection{Estimating Degree of Local Linearity} \label{app:experiment-linearity}
\newcommand{\expnumber}[2]{{#1}\mathrm{e}{#2}}
We performed an additional experiment to assess the degree of local linearity of a trained MLP with ReLU activation functions. In particular, for trained MLP $f(\cdot)$, we are estimating the expected absolute error
\begin{equation}
  \delta_{\text{Lin}} = \EE_{\bz \sim p(\bz)} \left[ \abs{ f(\bz (1 \pm \epsilon)) - f(\bz) (1 \pm \epsilon) } \right] ,
\end{equation}
where $\epsilon \geq 0$ and $\delta_{\text{Lin}}$ is zero for any $\epsilon$ if $f(\cdot)$ is linear around each $\bz$.

\begin{figure}[h]
  \centering

  \pgfplotsset{
    height=0.25\textwidth,
    width=0.9\linewidth,
    tick align=inside, 
    x tick label style={font=\tiny}, 
    y tick label style={font=\tiny, /pgf/number format/fixed, /pgf/number format/precision=5},
    scaled y ticks=false,
    try min ticks=5,
    y label style = {font=\scriptsize}, 
    scale only axis, 
    xlabel = {Value of $\epsilon$},
    ylabel = {Epxected Absolute Error: $\delta_{\text{Lin}}$}, 
  }
  % This file was created with tikzplotlib v0.10.1.post12.
\begin{tikzpicture}
  
  \begin{axis}[
  log basis x={10},
  xmode=log,
  ]
  \addplot [thick, black]
  table {%
  1e-06 1.82487863945548e-08
  1e-05 2.21547593623532e-08
  0.0001 6.99905485848307e-08
  0.001 6.58305095940673e-07
  0.01 6.58062724075337e-06
  0.1 6.58143488042341e-05
  1 0.00106569316618529
  };
  \addplot [thick, black]
  table {%
  1e-06 2.54620818942426e-08
  1e-05 3.51191399247217e-08
  0.0001 2.7528803777076e-07
  0.001 2.77530870935109e-06
  0.01 2.77350922128499e-05
  0.1 0.000277090139865695
  1 0.00342696154344427
  };
  \addplot [thick, black]
  table {%
  1e-06 1.83743146754215e-08
  1e-05 2.2297627646938e-08
  0.0001 1.03550656805881e-07
  0.001 1.02552240774664e-06
  0.01 1.0252702665826e-05
  0.1 0.000102376388436729
  1 0.00172305598563072
  };
  \addplot [thick, black]
  table {%
  1e-06 2.38427743895901e-08
  1e-05 2.70253995959796e-08
  0.0001 1.50294016363579e-07
  0.001 1.53483073546099e-06
  0.01 1.53388623296684e-05
  0.1 0.000153324729199714
  1 0.00237978287337972
  };
  \addplot [thick, black]
  table {%
  1e-06 2.20148203103136e-08
  1e-05 2.33228550400391e-08
  0.0001 5.39428201379018e-08
  0.001 4.42762915406541e-07
  0.01 4.43248843030904e-06
  0.1 4.43786795816908e-05
  1 0.00072526549152932
  };
  \addplot [thick, black]
  table {%
  1e-06 2.44618473403694e-08
  1e-05 3.14190052627305e-08
  0.0001 1.71712075899869e-07
  0.001 1.68595523799363e-06
  0.01 1.68275282737315e-05
  0.1 0.000168401893818764
  1 0.00234909390896264
  };
  \addplot [thick, black]
  table {%
  1e-06 2.42699152780046e-08
  1e-05 2.50260063011029e-08
  0.0001 1.18414401862712e-07
  0.001 1.12287620870897e-06
  0.01 1.1237607120715e-05
  0.1 0.000112476018074316
  1 0.000645667200962255
  };
  \addplot [thick, black]
  table {%
  1e-06 2.41269835242189e-08
  1e-05 3.67407503213725e-08
  0.0001 3.17684214033846e-07
  0.001 3.12394728011859e-06
  0.01 3.12274355585675e-05
  0.1 0.000312616221214864
  1 0.00380418243897272
  };
  \addplot [thick, black]
  table {%
  1e-06 3.12390252958294e-08
  1e-05 9.3396776056337e-08
  0.0001 9.52444927843084e-07
  0.001 9.49016025105137e-06
  0.01 9.4867266330588e-05
  0.1 0.000948787009751441
  1 0.00908596948906611
  };
  \addplot [thick, black]
  table {%
  1e-06 2.5547679216831e-08
  1e-05 2.99554438298651e-08
  0.0001 1.1854202244627e-07
  0.001 1.15868060228891e-06
  0.01 1.15261125229788e-05
  0.1 0.000115155108817754
  1 0.000806140922877374
  };
  \end{axis}
  
\end{tikzpicture}  
  \caption{Estimated divergence from a locally linear function as a function of $\epsilon$. Note that a value of zero means that the function behaves locally like a linear function.   \label{fig:linearisation_error} }
\end{figure}
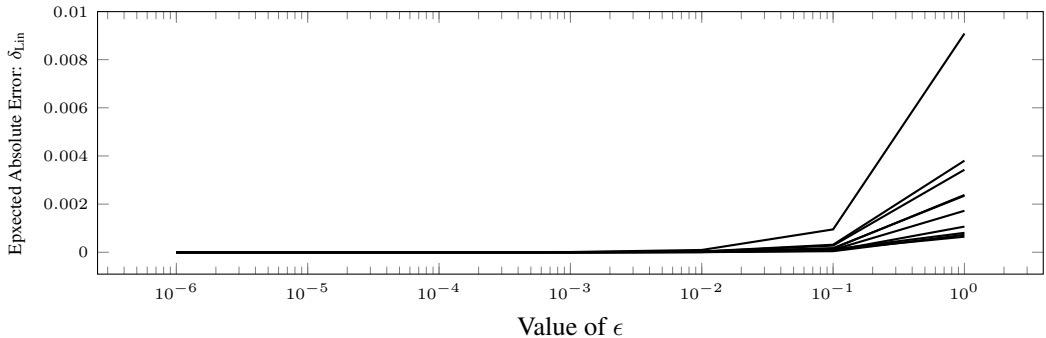

In our experiments we vary $\epsilon$ in the range of $\epsilon \in [\expnumber{1}{-6}, \expnumber{1}{-5}, \dots, 1]$ for a fully connected ReLU MLP with layers with sizes $[784, 128, 64, 10]$ trained on MNIST digits. After training, we removed the softmax operation on the last layer and measured the local linearisation error on the logits. We estimated the error on a random subset of $124$ validation data points and estimated the range of the inputs and the function outputs on the same subset.
The range of input values is $3.246$ and the range of the function outputs varies between $153.072$ and $291.168$. 
\cref{fig:linearisation_error} shows the results for each of the ten output dimensions scaled relative to their respective range.
We observe that the trained ReLU MLP obtains low expected absolute error and behaves locally linear to a certain degree. 

\subsection{Comparison of Value Covariance in Transformers}
One way to improve efficiency in transformer is dropping the correlation between values, \ie, drop the correlation $\cov{v_k}{v_l}$ given in \cref{eq:covariance_transformer}. We compare the performance of both approximations and the results are given in \cref{table: vit_cov_nlpd,table: vit_cov_acc,table: vit_cov_ece}. Both approximation results in almost the same results for NLPD, ACC and ECE.

\begin{table}[h!]
\setlength{\tabcolsep}{9.5pt}
\scriptsize
\centering
\caption{Negative Log Predictive Density (NLPD) for ViT with posterior approximation in the attention layers. We compare only considering variance for value $\bV$ and considering full covariance. For MFVI and LA, both approximations results in almost the same result.}
\begin{tabular}{l|cc|cc}
\toprule
{} & \multicolumn{2}{c|}{\textit{Mean Field Variational Inference}} & \multicolumn{2}{c}{\textit{Laplace Approximation}} \\[0.2em]
Dataset     & Full Covariance     & Only Variance      & Full Covariance     & Only Variance       \\[0.2em]
\midrule
\sc CIFAR-10    & $0.083 \pm 0.005$   & $0.083 \pm 0.005$  & $0.087 \pm 0.005$   & $0.087 \pm 0.005$   \\
\sc CIFAR-100   & $0.451 \pm 0.012$   & $0.450 \pm 0.012$  & $0.426 \pm 0.011$   & $0.426 \pm 0.011$   \\
\sc DTD         & $0.909 \pm 0.032$   & $0.919 \pm 0.032$  & $0.981 \pm 0.030$   & $0.981 \pm 0.030$   \\
\sc RESISC      & $0.272 \pm 0.011$   & $0.272 \pm 0.011$  & $0.297 \pm 0.011$   & $0.297 \pm 0.011$   \\
\sc ImageNet-R  & $1.080 \pm 0.046$   & $1.081 \pm 0.047$  & $1.192 \pm 0.042$   & $1.193 \pm 0.042$   \\
\bottomrule
\end{tabular}
\label{table: vit_cov_nlpd}
\end{table}

\begin{table}[h!]
\centering
\setlength{\tabcolsep}{9.5pt}
\scriptsize
\caption{Accuracy (ACC) for ViT with posterior approximation n the attention layers. We compare only considering variance for value $\bV$ and considering full covariance. For MFVI and LA, both approximations results in almost the same result.}
\begin{tabular}{l|cc|cc}
\toprule
{} & \multicolumn{2}{c|}{\textit{Mean Field Variational Inference}} & \multicolumn{2}{c}{\textit{Laplace Approximation}} \\[0.2em]
Dataset     & Full Covariance     & Only Variance      & Full Covariance     & Only Variance       \\[0.2em]
\midrule
\sc CIFAR-10    & $0.975 \pm 0.002$   & $0.975 \pm 0.002$  & $0.974 \pm 0.002$   & $0.974 \pm 0.002$   \\
\sc CIFAR-100   & $0.885 \pm 0.003$   & $0.885 \pm 0.003$  & $0.879 \pm 0.003$   & $0.879 \pm 0.003$   \\
\sc DTD         & $0.740 \pm 0.010$   & $0.743 \pm 0.010$  & $0.725 \pm 0.010$   & $0.725 \pm 0.010$   \\
\sc RESISC      & $0.917 \pm 0.003$   & $0.918 \pm 0.003$  & $0.906 \pm 0.004$   & $0.906 \pm 0.004$   \\
\sc ImageNet-R  & $0.736 \pm 0.012$   & $0.736 \pm 0.012$  & $0.725 \pm 0.012$   & $0.725 \pm 0.012$   \\
\bottomrule
\end{tabular}
\label{table: vit_cov_acc}
\end{table}

\begin{table}[h!]
\centering
\setlength{\tabcolsep}{9.5pt}
\scriptsize
\caption{Expected Calibration Error (ECE) for ViT with posterior approximation n the attention layers. We compare only considering variance for value $\bV$ and considering full covariance. For MFVI and LA, both approximations results in almost the same result.}
\begin{tabular}{l|cc|cc}
\toprule
{} & \multicolumn{2}{c|}{\textit{Mean Field Variational Inference}} & \multicolumn{2}{c}{\textit{Laplace Approximation}} \\[0.2em]
Dataset     & Full Covariance     & Only Variance      & Full Covariance     & Only Variance       \\[0.2em]
\midrule
\sc CIFAR-10    & $0.005$            & $0.004$            & $0.007$             & $0.007$             \\
\sc CIFAR-100   & $0.023$            & $0.024$            & $0.017$             & $0.017$             \\
\sc DTD         & $0.031$            & $0.034$            & $0.041$             & $0.041$             \\
\sc RESISC      & $0.008$            & $0.010$            & $0.013$             & $0.013$             \\
\sc ImageNet-R  & $0.037$            & $0.039$            & $0.089$             & $0.090$             \\
\bottomrule
\end{tabular}
\label{table: vit_cov_ece}
\end{table}

\subsection{Runtime Experiment}\label{app:runtime}
We compared the runtime of our method against sampling (using the `torch-laplace' library \cite{daxberger2021laplaceredux} for Laplace and the IVON \cite{shen2024ivon}) and the GLM implementation of the `torch-laplace' library for diagonal posterior covariances.
For our streamlined approach on ViT, we assessed two cases: (i) propagating $\cov{v_k}{v_l}$ covariance terms (\cf, \cref{eq:covariance_transformer}) through the transformer (+Cov), and (ii) ignoring $\cov{v_k}{v_l}$ covariance terms.
 We used a pre-trained ViT base model on CIFAR-10 and a pre-trained MLP on MNIST. For comparison we also list the runtime for a single forward pass. 
For this, we ran experiments on an NVIDIA H100 80GB GPU for $400$ data points, batchsize of one, and for each data point we repeated the measurement ten times. To account for code compilation overheads, we dropped the first run on each data point. We report the mean and standard deviation of the runtime (in milliseconds) over the remaining nine runs and all $400$ data points. The results are shown in \cref{tab:runtime}.
For ViT, we can see that our method without $\cov{v_k}{v_l}$ covariance terms has a comparable runtime to a single forward pass in the deterministic model. When additionally accounting for covariance terms, we obtain slight speed improvements over GLM but overall comparable performance. Note that our implementation is not optimised for speed and larger speedups may be obtained by optimising the code.
For MLP, we obtain slight speed improvements over LA GLM but overall comparable performance.
\begin{table}[h]
\centering
\setlength{\tabcolsep}{9.5pt}
\scriptsize
\caption{Wallclock times for ViT base on CIFAR-10 and MLP on MNIST in milliseconds. \label{tab:runtime} }
\begin{tabular}{l|l|r}
  \toprule
  Model & Methods & \multicolumn{1}{c}{\sc Avg. Runtime ($\pm$ std) $\downarrow$} \\
  \hline
  & MAP & \val{}{3.737}{0.093} \\ \cdashline{2-3} \rule{0pt}{2.25ex}
  & LA Sampling & \val{}{190.806}{0.137} \\
  & LA GLM & \val{}{17.191}{0.734}  \\
  \cellcolor{white} ViT & MFVI Sampling & \val{}{207.854}{0.307}  \\
\rowcolor{highlight} \cellcolor{white}
   & Ours (+ Cov) & \val{}{14.728}{0.144} \\ \rowcolor{highlight} \cellcolor{white}
   & Ours & \val{}{4.350}{0.079} \\ 
   \hline
   & MAP & \val{}{0.069}{0.001} \\ \cdashline{2-3} \rule{0pt}{2.25ex}
   & LA Sampling & \val{}{98.584}{3.737} \\
   & LA GLM & \val{}{1.656}{0.049}  \\
   \cellcolor{white} MLP & MFVI Sampling & \val{}{190.302}{0.466}  \\
 \rowcolor{highlight} \cellcolor{white}
    & Ours & \val{}{0.542}{0.073} \\ 
  \bottomrule
\end{tabular}
\end{table}


\begin{thebibliography}{71}
\providecommand{\natexlab}[1]{#1}
\providecommand{\url}[1]{\texttt{#1}}
\expandafter\ifx\csname urlstyle\endcsname\relax
  \providecommand{\doi}[1]{doi: #1}\else
  \providecommand{\doi}{doi: \begingroup \urlstyle{rm}\Url}\fi

\bibitem[Baumann et~al.(2024)Baumann, Li, Klasson, Mentu, Karthik, Akata,
  Solin, and Trapp]{baumann2024posthoc}
Anton Baumann, Rui Li, Marcus Klasson, Santeri Mentu, Shyamgopal Karthik,
  Zeynep Akata, Arno Solin, and Martin Trapp.
\newblock Post-hoc probabilistic vision-language models.
\newblock \emph{arXiv preprint arXiv:2412.06014}, 2024.

\bibitem[Begoli et~al.(2019)Begoli, Bhattacharya, and Kusnezov]{begoli2019need}
Edmon Begoli, Tanmoy Bhattacharya, and Dimitri Kusnezov.
\newblock The need for uncertainty quantification in machine-assisted medical
  decision making.
\newblock \emph{Nature Machine Intelligence}, 1\penalty0 (1):\penalty0 20--23,
  2019.

\bibitem[Blei et~al.(2017)Blei, Kucukelbir, and McAuliffe]{blei2017variational}
David~M Blei, Alp Kucukelbir, and Jon~D McAuliffe.
\newblock Variational inference: A review for statisticians.
\newblock \emph{Journal of the American Statistical Association}, 112\penalty0
  (518):\penalty0 859--877, 2017.

\bibitem[Blundell et~al.(2015)Blundell, Cornebise, Kavukcuoglu, and
  Wierstra]{blundell2015weight}
Charles Blundell, Julien Cornebise, Koray Kavukcuoglu, and Daan Wierstra.
\newblock Weight uncertainty in neural network.
\newblock In \emph{Proceedings of the 32th International Conference on Machine
  Learning (ICML)}, volume~37 of \emph{Proceedings of Machine Learning
  Research}, pp.\  1613--1622. PMLR, 2015.

\bibitem[Burroni et~al.(2024)Burroni, Domke, and Sheldon]{burroni2024sample}
Javier Burroni, Justin Domke, and Daniel Sheldon.
\newblock Sample average approximation for black-box variational inference.
\newblock In \emph{Proceedings of the 40th Conference on Uncertainty in
  Artificial Intelligence (UAI)}. {AUAI} Press, 2024.

\bibitem[Cheng et~al.(2017)Cheng, Han, and Lu]{cheng2017resisc}
Gong Cheng, Junwei Han, and Xiaoqiang Lu.
\newblock Remote sensing image scene classification: Benchmark and state of the
  art.
\newblock \emph{Proceedings of the IEEE}, 105\penalty0 (10):\penalty0
  1865--1883, 2017.

\bibitem[Cimpoi et~al.(2014)Cimpoi, Maji, Kokkinos, Mohamed, and
  Vedaldi]{cimpoi2014dtd}
Mircea Cimpoi, Subhransu Maji, Iasonas Kokkinos, Sammy Mohamed, and Andrea
  Vedaldi.
\newblock Describing textures in the wild.
\newblock In \emph{Proceedings of the IEEE Conference on Computer Vision and
  Pattern Recognition (CVPR)}, pp.\  3606--3613. {IEEE} Computer Society, 2014.

\bibitem[Coker et~al.(2022)Coker, Bruinsma, Burt, Pan, and
  Doshi-Velez]{coker2022wide}
Beau Coker, Wessel~P Bruinsma, David~R Burt, Weiwei Pan, and Finale
  Doshi-Velez.
\newblock Wide mean-field bayesian neural networks ignore the data.
\newblock In \emph{Proceedings of the twenty fifth International Conference on
  Artificial Intelligence and Statistics (AISTATS)}, volume 131 of
  \emph{Proceedings of Machine Learning Research}, pp.\  5276--5333. PMLR,
  2022.

\bibitem[Dangel et~al.(2025)Dangel, Mucsányi, Weber, and
  Eschenhagen]{dangel2025kroneckerfactored}
Felix Dangel, Bálint Mucsányi, Tobias Weber, and Runa Eschenhagen.
\newblock Kronecker-factored approximate curvature (kfac) from scratch.
\newblock \emph{arXiv}, 2025.
\newblock URL \url{https://github.com/f-dangel/kfac-tutorial}.

\bibitem[Daxberger et~al.(2021{\natexlab{a}})Daxberger, Kristiadi, Immer,
  Eschenhagen, Bauer, and Hennig]{daxberger2021laplaceredux}
Erik Daxberger, Agustinus Kristiadi, Alexander Immer, Runa Eschenhagen,
  Matthias Bauer, and Philipp Hennig.
\newblock Laplace redux -- effortless {B}ayesian deep learning.
\newblock In \emph{Advances in Neural Information Processing Systems 34
  (NeurIPS)}, volume~34, pp.\  20089--20103. Curran Associates, Inc.,
  2021{\natexlab{a}}.

\bibitem[Daxberger et~al.(2021{\natexlab{b}})Daxberger, Nalisnick, Allingham,
  Antor{\'a}n, and Hern{\'a}ndez-Lobato]{daxberger2021subnetwork}
Erik Daxberger, Eric Nalisnick, James~U Allingham, Javier Antor{\'a}n, and
  Jos{\'e}~Miguel Hern{\'a}ndez-Lobato.
\newblock {B}ayesian deep learning via subnetwork inference.
\newblock In \emph{Proceedings of the 38th International Conference on Machine
  Learning (ICML)}, volume 119 of \emph{Proceedings of Machine Learning
  Research}, pp.\  2510--2521. PMLR, 2021{\natexlab{b}}.

\bibitem[Deng et~al.(2009)Deng, Dong, Socher, Li, Li, and
  Fei-Fei]{deng2009imagenet}
Jia Deng, Wei Dong, Richard Socher, Li-Jia Li, Kai Li, and Li~Fei-Fei.
\newblock Imagenet: A large-scale hierarchical image database.
\newblock In \emph{Proceedings of the IEEE Conference on Computer Vision and
  Pattern Recognition (CVPR)}, pp.\  248--255. {IEEE} Computer Society, 2009.

\bibitem[Dhawan et~al.(2023)Dhawan, Huang, Bae, and
  Grosse]{dhawan2023efficient}
Nikita Dhawan, Sicong Huang, Juhan Bae, and Roger~Baker Grosse.
\newblock Efficient parametric approximations of neural network function space
  distance.
\newblock In \emph{Proceedings of the 40th International Conference on Machine
  Learning {(ICML)}}, Proceedings of Machine Learning Research, pp.\
  7795--7812. PMLR, 2023.

\bibitem[Dosovitskiy et~al.(2021)Dosovitskiy, Beyer, Kolesnikov, Weissenborn,
  Zhai, Unterthiner, Dehghani, Minderer, Heigold, Gelly, Uszkoreit, and
  Houlsby]{dosovitskiy2021vit}
Alexey Dosovitskiy, Lucas Beyer, Alexander Kolesnikov, Dirk Weissenborn,
  Xiaohua Zhai, Thomas Unterthiner, Mostafa Dehghani, Matthias Minderer, Georg
  Heigold, Sylvain Gelly, Jakob Uszkoreit, and Neil Houlsby.
\newblock An image is worth 16x16 words: Transformers for image recognition at
  scale.
\newblock In \emph{International Conference on Learning Representations
  (ICLR)}, 2021.

\bibitem[Eschenhagen et~al.(2021)Eschenhagen, Daxberger, Hennig, and
  Kristiadi]{eschenhagen2021deepensemble1}
Runa Eschenhagen, Erik Daxberger, Philipp Hennig, and Agustinus Kristiadi.
\newblock Mixtures of {L}apkace approximations for improved post-hoc
  uncertainty in deep learning.
\newblock In \emph{NeurIPS workshop on {B}ayesian deep learning}, 2021.

\bibitem[Foong et~al.(2020)Foong, Burt, Li, and
  Turner]{foong2020expressiveness}
Andrew Foong, David Burt, Yingzhen Li, and Richard Turner.
\newblock On the expressiveness of approximate inference in {B}ayesian neural
  networks.
\newblock In \emph{Advances in Neural Information Processing Systems 33
  (NeurIPS)}, pp.\  15897--15908. Curran Associates, Inc., 2020.

\bibitem[Fortuin et~al.(2021)Fortuin, Garriga-Alonso, Ober, Wenzel, R{\"a}tsch,
  Turner, van~der Wilk, and Aitchison]{fortuin2021bayesian}
Vincent Fortuin, Adri{\`a} Garriga-Alonso, Sebastian~W Ober, Florian Wenzel,
  Gunnar R{\"a}tsch, Richard~E Turner, Mark van~der Wilk, and Laurence
  Aitchison.
\newblock {B}ayesian neural network priors revisited.
\newblock In \emph{International Conference on Learning Representations
  (ICLR)}, 2021.

\bibitem[Gal \& Ghahramani(2016)Gal and Ghahramani]{gal2016dropout}
Yarin Gal and Zoubin Ghahramani.
\newblock Dropout as a bayesian approximation: Representing model uncertainty
  in deep learning.
\newblock In \emph{Proceedings of the 33th International Conference on Machine
  Learning (ICML)}, volume~48 of \emph{Proceedings of Machine Learning
  Research}, pp.\  1050--1059. PMLR, 2016.

\bibitem[Gal et~al.(2017)Gal, Islam, and Ghahramani]{gal2017deep}
Yarin Gal, Riashat Islam, and Zoubin Ghahramani.
\newblock Deep bayesian active learning with image data.
\newblock In \emph{Proceedings of the 34th International Conference on Machine
  Learning (ICML)}, Proceedings of Machine Learning Research, pp.\  1183--1192.
  PMLR, 2017.

\bibitem[Gelberg et~al.(2024)Gelberg, van~der Ouderaa, van~der Wilk, and
  Gal]{gelberg2024variational}
Yoav Gelberg, Tycho F.~A. van~der Ouderaa, Mark van~der Wilk, and Yarin Gal.
\newblock Variational inference failures under model symmetries: Permutation
  invariant posteriors for {B}ayesian neural networks.
\newblock In \emph{ICML 2024 Workshop on Geometry-grounded Representation
  Learning and Generative Modeling}, 2024.

\bibitem[Giordano et~al.(2024)Giordano, Ingram, and
  Broderick]{giordano2024black}
Ryan Giordano, Martin Ingram, and Tamara Broderick.
\newblock Black box variational inference with a deterministic objective:
  Faster, more accurate, and even more black box.
\newblock \emph{Journal of Machine Learning Research}, 25\penalty0
  (18):\penalty0 1--39, 2024.

\bibitem[Goulet et~al.(2021)Goulet, Nguyen, and Amiri]{goulet2021tractable}
James-A Goulet, Luong~Ha Nguyen, and Saeid Amiri.
\newblock Tractable approximate {G}aussian inference for {B}ayesian neural
  networks.
\newblock \emph{Journal of Machine Learning Research}, 22\penalty0
  (251):\penalty0 1--23, 2021.

\bibitem[Havasi et~al.(2021)Havasi, Jenatton, Fort, Liu, Snoek,
  Lakshminarayanan, Dai, and Tran]{havasi2021deepensemble2}
Marton Havasi, Rodolphe Jenatton, Stanislav Fort, Jeremiah~Zhe Liu, Jasper
  Snoek, Balaji Lakshminarayanan, Andrew~Mingbo Dai, and Dustin Tran.
\newblock Training independent subnetworks for robust prediction.
\newblock In \emph{International Conference on Learning Representations
  (ICLR)}, 2021.

\bibitem[He et~al.(2016)He, Zhang, Ren, and Sun]{he2016deep}
Kaiming He, Xiangyu Zhang, Shaoqing Ren, and Jian Sun.
\newblock Deep residual learning for image recognition.
\newblock In \emph{IEEE Conference on Computer Vision and Pattern Recognition
  (CVPR)}, pp.\  770--778. {IEEE} Computer Society, 2016.

\bibitem[Hendrycks et~al.(2021)Hendrycks, Zhao, Basart, Steinhardt, and
  Song]{hendrycks2021imagenetr}
Dan Hendrycks, Kevin Zhao, Sebastian Basart, Jacob Steinhardt, and Dawn Song.
\newblock The many faces of robustness: A critical analysis of
  out-of-distribution generalization.
\newblock In \emph{Proceedings of the IEEE/CVF International Conference on
  Computer Vision (ICCV)}, pp.\  8340--8349. {IEEE}, 2021.

\bibitem[Immer et~al.(2021{\natexlab{a}})Immer, Bauer, Fortuin, R{\"a}tsch, and
  Emtiyaz]{immer2021scalable}
Alexander Immer, Matthias Bauer, Vincent Fortuin, Gunnar R{\"a}tsch, and
  Khan~Mohammad Emtiyaz.
\newblock Scalable marginal likelihood estimation for model selection in deep
  learning.
\newblock In \emph{Proceedings of the 38th International Conference on Machine
  Learning (ICML)}, Proceedings of Machine Learning Research, pp.\  4563--4573.
  PMLR, 2021{\natexlab{a}}.

\bibitem[Immer et~al.(2021{\natexlab{b}})Immer, Korzepa, and
  Bauer]{immer2021laplaceglm}
Alexander Immer, Maciej Korzepa, and Matthias Bauer.
\newblock Improving predictions of {B}ayesian neural nets via local
  linearization.
\newblock In \emph{Proceedings of the twenty forth International Conference on
  Artificial Intelligence and Statistics (AISTATS)}, volume 130 of
  \emph{Proceedings of Machine Learning Research}, pp.\  703--711. PMLR,
  2021{\natexlab{b}}.

\bibitem[Kampen et~al.(2024)Kampen, Als, and Andersen]{kampen2024towards}
Peter~JT Kampen, Gustav~RS Als, and Michael~Riis Andersen.
\newblock Towards scalable {B}ayesian transformers: Investigating stochastic
  subset selection for nlp.
\newblock In \emph{Proceedings of the 40th Conference on Uncertainty in
  Artificial Intelligence (UAI)}. {AUAI} Press, 2024.

\bibitem[Kan(2008)]{kan2008moments}
Raymond Kan.
\newblock From moments of sum to moments of product.
\newblock \emph{Journal of Multivariate Analysis}, 99\penalty0 (3):\penalty0
  542--554, 2008.

\bibitem[Kelly et~al.(2023)Kelly, Longjohn, and Nottingham]{ucirepository}
Markelle Kelly, Rachel Longjohn, and Kolby Nottingham.
\newblock The {UCI} machine learning repository, 2023.
\newblock {URL}: https://archive.ics.uci.edu.

\bibitem[Kristiadi et~al.(2020)Kristiadi, Hein, and
  Hennig]{kristiadi2020beingbayesian}
Agustinus Kristiadi, Matthias Hein, and Philipp Hennig.
\newblock Being bayesian, even just a bit, fixes overconfidence in relu
  networks.
\newblock In \emph{Proceedings of the 37th International Conference on Machine
  Learning (ICML)}, volume 119 of \emph{Proceedings of Machine Learning
  Research}, pp.\  5436--5446. PMLR, 2020.

\bibitem[Kristiadi et~al.(2023)Kristiadi, Immer, Eschenhagen, and
  Fortuin]{kristiadi2023promises}
Agustinus Kristiadi, Alexander Immer, Runa Eschenhagen, and Vincent Fortuin.
\newblock Promises and pitfalls of the linearized {L}apkace in {B}ayesian
  optimization.
\newblock In \emph{Fifth Symposium on Advances in Approximate {B}ayesian
  Inference}, 2023.

\bibitem[Krizhevsky \& Hinton(2009)Krizhevsky and
  Hinton]{krizhevsky2009learning}
Alex Krizhevsky and Geoffrey Hinton.
\newblock Learning multiple layers of features from tiny images.
\newblock Technical report, Toronto, ON, Canada, 2009.

\bibitem[Lakshminarayanan et~al.(2017)Lakshminarayanan, Pritzel, and
  Blundell]{lakshminarayanan2017deepensemble}
Balaji Lakshminarayanan, Alexander Pritzel, and Charles Blundell.
\newblock Simple and scalable predictive uncertainty estimation using deep
  ensembles.
\newblock In \emph{Advances in Neural Information Processing Systems 30
  (NeurIPS)}, volume~30, pp.\  6402--6413. Curran Associates, Inc., 2017.

\bibitem[LeCun et~al.(1998)LeCun, Bottou, Bengio, and
  Haffner]{lecun1998gradient}
Yann LeCun, L{\'e}on Bottou, Yoshua Bengio, and Patrick Haffner.
\newblock Gradient-based learning applied to document recognition.
\newblock \emph{Proceedings of the IEEE}, 86\penalty0 (11):\penalty0
  2278--2324, 1998.

\bibitem[Lu et~al.(2021)Lu, Yao, Zhang, Zhu, Xu, Gao, Xu, Xiang, and
  Zhang]{lu2021soft}
Jiachen Lu, Jinghan Yao, Junge Zhang, Xiatian Zhu, Hang Xu, Weiguo Gao,
  Chunjing Xu, Tao Xiang, and Li~Zhang.
\newblock Soft: Softmax-free transformer with linear complexity.
\newblock In \emph{Advances in Neural Information Processing Systems 34
  (NeurIPS)}, pp.\  21297--21309. Curran Associates, Inc., 2021.

\bibitem[MacKay(1992{\natexlab{a}})]{mackay1992information}
David~JC MacKay.
\newblock Information-based objective functions for active data selection.
\newblock \emph{Neural Computation}, 4\penalty0 (4):\penalty0 590--604,
  1992{\natexlab{a}}.

\bibitem[MacKay(1992{\natexlab{b}})]{mackay1992prohit-1}
David~JC MacKay.
\newblock {B}ayesian interpolation.
\newblock \emph{Neural computation}, 4\penalty0 (3):\penalty0 415--447,
  1992{\natexlab{b}}.

\bibitem[MacKay(1996)]{mackay1996bayesian}
David~JC MacKay.
\newblock {B}ayesian methods for backpropagation networks.
\newblock In \emph{Models of Neural Networks {III}: {A}ssociation,
  Generalization, and Representation}, pp.\  211--254. Springer, 1996.

\bibitem[Maddox et~al.(2019)Maddox, Izmailov, Garipov, Vetrov, and
  Wilson]{maddox2019swag}
Wesley~J. Maddox, Pavel Izmailov, Timur Garipov, Dmitry~P. Vetrov, and
  Andrew~Gordon Wilson.
\newblock A simple baseline for {B}ayesian uncertainty in deep learning.
\newblock In \emph{Advances in Neural Information Processing Systems 32
  (NeurIPS)}, pp.\  13132--13143. Curran Associates, Inc., 2019.

\bibitem[Martens \& Grosse(2015)Martens and Grosse]{martens2015optimizing}
James Martens and Roger Grosse.
\newblock Optimizing neural networks with {K}ronecker-factored approximate
  curvature.
\newblock In \emph{Proceedings of the 32nd International Conference on Machine
  Learning (ICML)}, Proceedings of Machine Learning Research, pp.\  2408--2417.
  PMLR, 2015.

\bibitem[Meronen et~al.(2021)Meronen, Trapp, and Solin]{meronen2021periodic}
Lassi Meronen, Martin Trapp, and Arno Solin.
\newblock Periodic activation functions induce stationarity.
\newblock In \emph{Advances in Neural Information Processing Systems 34
  (NeurIPS)}, pp.\  1673--1685. Curran Associates, Inc., 2021.

\bibitem[Meronen et~al.(2024)Meronen, Trapp, Pilzer, Yang, and
  Solin]{meronen2024fixing}
Lassi Meronen, Martin Trapp, Andrea Pilzer, Le~Yang, and Arno Solin.
\newblock Fixing overconfidence in dynamic neural networks.
\newblock In \emph{Proceedings of the IEEE/CVF Winter Conference on
  Applications of Computer Vision (WACV)}, pp.\  2680--2690, 2024.

\bibitem[Michelmore et~al.(2020)Michelmore, Wicker, Laurenti, Cardelli, Gal,
  and Kwiatkowska]{michelmore2020uncertainty}
Rhiannon Michelmore, Matthew Wicker, Luca Laurenti, Luca Cardelli, Yarin Gal,
  and Marta Kwiatkowska.
\newblock Uncertainty quantification with statistical guarantees in end-to-end
  autonomous driving control.
\newblock In \emph{IEEE International Conference on Robotics and Automation
  (ICRA)}, pp.\  7344--7350. {IEEE}, 2020.

\bibitem[Nadarajah \& Pog{\'a}ny(2016)Nadarajah and
  Pog{\'a}ny]{nadarajah2016gaussian-product}
Saralees Nadarajah and Tibor~K Pog{\'a}ny.
\newblock On the distribution of the product of correlated normal random
  variables.
\newblock \emph{Comptes Rendus. Math{\'e}matique}, 354\penalty0 (2):\penalty0
  201--204, 2016.

\bibitem[Nalisnick(2018)]{nalisnick2018priors}
Eric~Thomas Nalisnick.
\newblock \emph{On priors for {B}ayesian neural networks}.
\newblock University of California, Irvine, 2018.

\bibitem[Netzer et~al.(2011)Netzer, Wang, Coates, Bissacco, Wu, Ng,
  et~al.]{netzer2011reading}
Yuval Netzer, Tao Wang, Adam Coates, Alessandro Bissacco, Baolin Wu, Andrew~Y
  Ng, et~al.
\newblock Reading digits in natural images with unsupervised feature learning.
\newblock In \emph{NIPS workshop on deep learning and unsupervised feature
  learning}, 2011.

\bibitem[Papamarkou et~al.(2024)Papamarkou, Skoularidou, Palla, Aitchison,
  Arbel, Dunson, Filippone, Fortuin, Hennig, Hern{\'{a}}ndez{-}Lobato, Hubin,
  Immer, Karaletsos, Khan, Kristiadi, Li, Mandt, Nemeth, Osborne, Rudner,
  R{\"{u}}gamer, Teh, Welling, Wilson, and Zhang]{papamarkou2024position}
Theodore Papamarkou, Maria Skoularidou, Konstantina Palla, Laurence Aitchison,
  Julyan Arbel, David~B. Dunson, Maurizio Filippone, Vincent Fortuin, Philipp
  Hennig, Jos{\'{e}}~Miguel Hern{\'{a}}ndez{-}Lobato, Aliaksandr Hubin,
  Alexander Immer, Theofanis Karaletsos, Mohammad~Emtiyaz Khan, Agustinus
  Kristiadi, Yingzhen Li, Stephan Mandt, Christopher Nemeth, Michael~A.
  Osborne, Tim G.~J. Rudner, David R{\"{u}}gamer, Yee~Whye Teh, Max Welling,
  Andrew~Gordon Wilson, and Ruqi Zhang.
\newblock Position: {B}ayesian deep learning is needed in the age of
  large-scale ai.
\newblock In \emph{Proceedings of the 41st International Conference on Machine
  Learning (ICML)}, volume 235 of \emph{Proceedings of Machine Learning
  Research}. PMLR, 2024.

\bibitem[Petersen et~al.(2024)Petersen, Mishra, Kuehne, Borgelt, Deussen, and
  Yurochkin]{peterse2024stable}
Felix Petersen, Aashwin~Ananda Mishra, Hilde Kuehne, Christian Borgelt, Oliver
  Deussen, and Mikhail Yurochkin.
\newblock Uncertainty quantification via stable distribution propagation.
\newblock In \emph{International Conference on Learning Representations
  (ICLR)}, 2024.

\bibitem[Psaros et~al.(2023)Psaros, Meng, Zou, Guo, and
  Karniadakis]{psaros2023uncertainty}
Apostolos~F Psaros, Xuhui Meng, Zongren Zou, Ling Guo, and George~Em
  Karniadakis.
\newblock Uncertainty quantification in scientific machine learning: Methods,
  metrics, and comparisons.
\newblock \emph{Journal of Computational Physics}, 477:\penalty0 111902, 2023.

\bibitem[Radford et~al.(2019)Radford, Wu, Child, Luan, Amodei, and
  Sutskever]{radford2019language}
Alec Radford, Jeff Wu, Rewon Child, David Luan, Dario Amodei, and Ilya
  Sutskever.
\newblock Language models are unsupervised multitask learners.
\newblock \emph{OpenAI blog}, 2019.

\bibitem[Ritter et~al.(2018)Ritter, Botev, and Barber]{ritter2018scalable}
Hippolyt Ritter, Aleksandar Botev, and David Barber.
\newblock A scalable {L}aplace approximation for neural networks.
\newblock In \emph{International Conference on Learning Representations
  (ICLR)}, 2018.

\bibitem[S{\"a}rkk{\"a} \& Svensson(2023)S{\"a}rkk{\"a} and
  Svensson]{sarkka2023bayesian}
Simo S{\"a}rkk{\"a} and Lennart Svensson.
\newblock \emph{{B}ayesian Filtering and Smoothing}.
\newblock Cambridge University Press, 2023.

\bibitem[Scannell et~al.(2024)Scannell, Mereu, Chang, Tamir, Pajarinen, and
  Solin]{scannell2024function}
Aidan Scannell, Riccardo Mereu, Paul~Edmund Chang, Ella Tamir, Joni Pajarinen,
  and Arno Solin.
\newblock Function-space parameterization of neural networks for sequential
  learning.
\newblock In \emph{International Conference on Learning Representations
  (ICLR)}, 2024.

\bibitem[Shen et~al.(2024)Shen, Daheim, Cong, Nickl, Marconi, Raoul, Yokota,
  Gurevych, Cremers, Khan, and Möllenhoff]{shen2024ivon}
Yuesong Shen, Nico Daheim, Bai Cong, Peter Nickl, Gian~Maria Marconi, Bazan
  Clement Emile~Marcel Raoul, Rio Yokota, Iryna Gurevych, Daniel Cremers,
  Mohammad~Emtiyaz Khan, and Thomas Möllenhoff.
\newblock Variational learning is effective for large deep networks.
\newblock In \emph{Proceedings of the 41st International Conference on Machine
  Learning (ICML)}, volume 235 of \emph{Proceedings of Machine Learning
  Research}. PMLR, 2024.

\bibitem[Smith et~al.(2023)Smith, Kirsch, Farquhar, Gal, Foster, and
  Rainforth]{smith2023prediction}
Freddie~Bickford Smith, Andreas Kirsch, Sebastian Farquhar, Yarin Gal, Adam
  Foster, and Tom Rainforth.
\newblock Prediction-oriented bayesian active learning.
\newblock In \emph{Proceedings of the twenty sixth International Conference on
  Artificial Intelligence and Statistics (AISTATS)}, Proceedings of Machine
  Learning Research, pp.\  7331--7348. PMLR, 2023.

\bibitem[Tran et~al.(2022)Tran, Rossi, Milios, and Filippone]{tran2022all}
Ba-Hien Tran, Simone Rossi, Dimitrios Milios, and Maurizio Filippone.
\newblock All you need is a good functional prior for {B}ayesian deep learning.
\newblock \emph{Journal of Machine Learning Research}, 23\penalty0
  (74):\penalty0 1--56, 2022.

\bibitem[Vaswani et~al.(2017)Vaswani, Shazeer, Parmar, Uszkoreit, Jones, Gomez,
  Kaiser, and Polosukhin]{vaswani2017attention}
Ashish Vaswani, Noam Shazeer, Niki Parmar, Jakob Uszkoreit, Llion Jones,
  Aidan~N Gomez, \L~ukasz Kaiser, and Illia Polosukhin.
\newblock Attention is all you need.
\newblock In \emph{Advances in Neural Information Processing Systems 30
  (NeurIPS)}. Curran Associates, Inc., 2017.

\bibitem[Vono et~al.(2022)Vono, Dobigeon, and Chainais]{vono2022high}
Maxime Vono, Nicolas Dobigeon, and Pierre Chainais.
\newblock High-dimensional {G}aussian sampling: a review and a unifying
  approach based on a stochastic proximal point algorithm.
\newblock \emph{SIAM Review}, 64\penalty0 (1):\penalty0 3--56, 2022.

\bibitem[Wang et~al.(2019{\natexlab{a}})Wang, Pruksachatkun, Nangia, Singh,
  Michael, Hill, Levy, and Bowman]{wang2019superglue}
Alex Wang, Yada Pruksachatkun, Nikita Nangia, Amanpreet Singh, Julian Michael,
  Felix Hill, Omer Levy, and Samuel Bowman.
\newblock Superglue: {A} stickier benchmark for general-purpose language
  understanding systems.
\newblock In \emph{Advances in Neural Information Processing Systems 32
  (NeurIPS)}, pp.\  3266--3280. Curran Associates, Inc., 2019{\natexlab{a}}.

\bibitem[Wang et~al.(2019{\natexlab{b}})Wang, Singh, Michael, Hill, Levy, and
  Bowman]{wang2018glue}
Alex Wang, Amanpreet Singh, Julian Michael, Felix Hill, Omer Levy, and Samuel~R
  Bowman.
\newblock Glue: A multi-task benchmark and analysis platform for natural
  language understanding.
\newblock In \emph{International Conference on Learning Representations
  (ICLR)}, 2019{\natexlab{b}}.

\bibitem[Wenzel et~al.(2020)Wenzel, Roth, Veeling, Swiatkowski, Tran, Mandt,
  Snoek, Salimans, Jenatton, and Nowozin]{wenzel2020good}
Florian Wenzel, Kevin Roth, Bastiaan Veeling, Jakub Swiatkowski, Linh Tran,
  Stephan Mandt, Jasper Snoek, Tim Salimans, Rodolphe Jenatton, and Sebastian
  Nowozin.
\newblock How good is the {B}ayes posterior in deep neural networks really?
\newblock In \emph{Proceedings of the 37th International Conference on Machine
  Learning (ICML)}, volume 119 of \emph{Proceedings of Machine Learning
  Research}, pp.\  10248--10259. PMLR, 2020.

\bibitem[Wilson(2020)]{wilson2020case}
Andrew~Gordon Wilson.
\newblock The case for {B}ayesian deep learning.
\newblock \emph{arXiv preprint arXiv:2001.10995}, 2020.

\bibitem[Wilson \& Izmailov(2020)Wilson and Izmailov]{wilson2020bayesian}
Andrew~Gordon Wilson and Pavel Izmailov.
\newblock {B}ayesian deep learning and a probabilistic perspective of
  generalization.
\newblock In \emph{Advances in Neural Information Processing Systems 33
  (NeurIPS)}, pp.\  4697--4708. Curran Associates, Inc., 2020.

\bibitem[Wolf et~al.(2019)Wolf, Debut, Sanh, Chaumond, Delangue, Moi, Cistac,
  Rault, Louf, Funtowicz, and Brew]{wolf2019huggingface}
Thomas Wolf, Lysandre Debut, Victor Sanh, Julien Chaumond, Clement Delangue,
  Anthony Moi, Pierric Cistac, Tim Rault, R{\'{e}}mi Louf, Morgan Funtowicz,
  and Jamie Brew.
\newblock Huggingface's transformers: State-of-the-art natural language
  processing.
\newblock \emph{arXiv preprint arXiv:1910.03771}, 2019.

\bibitem[Wolinski \& Arbel(2022)Wolinski and Arbel]{wolinski2022gaussian}
Pierre Wolinski and Julyan Arbel.
\newblock Gaussian pre-activations in neural networks: Myth or reality?
\newblock \emph{arXiv preprint arXiv:2205.12379}, 2022.

\bibitem[Wu et~al.(2019)Wu, Nowozin, Meeds, Turner, Hern{\'{a}}ndez{-}Lobato,
  and Gaunt]{wu2018deterministic}
Anqi Wu, Sebastian Nowozin, Edward Meeds, Richard~E. Turner, Jos{\'{e}}~Miguel
  Hern{\'{a}}ndez{-}Lobato, and Alexander~L. Gaunt.
\newblock Deterministic variational inference for robust {B}ayesian neural
  networks.
\newblock In \emph{International Conference on Learning Representations
  (ICLR)}, 2019.

\bibitem[Xiao et~al.(2017)Xiao, Rasul, and Vollgraf]{xiao2017fashion}
Han Xiao, Kashif Rasul, and Roland Vollgraf.
\newblock Fashion-mnist: a novel image dataset for benchmarking machine
  learning algorithms.
\newblock \emph{arXiv preprint arXiv:1708.07747}, 2017.

\bibitem[Yang et~al.(2024)Yang, Robeyns, Wang, and
  Aitchison]{yang2024bayeslora}
Adam~X Yang, Maxime Robeyns, Xi~Wang, and Laurence Aitchison.
\newblock {B}ayesian low-rank adaptation for large language models.
\newblock In \emph{International Conference on Learning Representations
  (ICLR)}, 2024.

\bibitem[Yang et~al.(2023)Yang, Shi, Wei, Liu, Zhao, Ke, Shi, Li, Hu, Gao, Xu,
  Rubin, and Roth]{Yang2023medmnist}
Jiancheng Yang, Rui Shi, Donglai Wei, Zeju Liu, Lin Zhao, Bilian Ke, Ziyang
  Shi, Yunzhu Li, Xiaoyang Hu, Yang Gao, Ye~Xu, Daniel~L. Rubin, and Holger~R.
  Roth.
\newblock Medmnist v2: A large-scale lightweight benchmark for 2d and 3d
  biomedical image classification.
\newblock \emph{Scientific Data}, 10\penalty0 (1):\penalty0 1--14, 2023.
\newblock \doi{10.1038/s41597-023-02552-1}.

\bibitem[Zhang et~al.(2018)Zhang, B{\"u}tepage, Kjellstr{\"o}m, and
  Mandt]{zhang2018advances}
Cheng Zhang, Judith B{\"u}tepage, Hedvig Kjellstr{\"o}m, and Stephan Mandt.
\newblock Advances in variational inference.
\newblock \emph{IEEE Transactions on Pattern Analysis and Machine
  Intelligence}, 41\penalty0 (8):\penalty0 2008--2026, 2018.

\end{thebibliography}
\end{document}